\documentclass[twocolumn]{svjour3b}        % twocolumn
\smartqed  % flush right qed marks, e.g. at end of proof

\usepackage{graphicx}
\usepackage{times}
\usepackage{epsfig}
\usepackage{graphicx}
\usepackage{amsmath}
\usepackage{amssymb}

\usepackage{nicefrac}
\usepackage[linesnumbered,ruled]{algorithm2e}
\usepackage{algorithmic}
\usepackage{float}
\usepackage{multirow}
\usepackage{bbm}

\usepackage{wrapfig}
\usepackage[linesnumbered,ruled]{algorithm2e}

\usepackage{subfig}
\usepackage{tabularx}
\usepackage{afterpage}
\usepackage{floatflt}
\usepackage{array}
\usepackage{arydshln}
\usepackage{verbatim}
\usepackage{makecell}
\usepackage[english]{babel}
\usepackage [autostyle, english = american]{csquotes}
\MakeOuterQuote{"}
\usepackage{microtype}
\usepackage{url}
\usepackage[htt]{hyphenat}
\usepackage{xcolor}

\newcommand{\inner}[1]{\left\langle#1\right\rangle}

\def\A{\mathcal{A}}

\def\R{\mathbb{R}}

\def\U{\mathcal{U}}

\newcommand{\norm}[1]{\left\|#1\right\|}

\def\V{\mathcal{V}}

\def\sgn{\mathop{\rm sgn}\nolimits}

\def\ones{\mathop{\rm e}\nolimits}

\def\argmax{\mathop{\rm arg\,max}\limits}%    a math operator.
\def\argmin{\mathop{\rm arg\,min}\limits}%    a math operator.
\def\minop{\mathop{\rm min}\limits}
\def\maxop{\mathop{\rm max}\limits}

\def\max{\mathop{\rm max}\nolimits}
\def\ones{\mathbf{1}}

\newcolumntype{C}[1]{>{\centering\arraybackslash}p{#1}}
\newcolumntype{L}[1]{>{\raggedright\arraybackslash}p{#1}}
\newcolumntype{R}[1]{>{\raggedleft\arraybackslash}p{#1}}
\begin{document}

\title{Scaling up the randomized gradient-free adversarial attack reveals overestimation of
robustness using established attacks}

\author{Francesco Croce$^*$ \and Jonas Rauber$^*$ \and Matthias Hein
}

\authorrunning{F. Croce \and J. Rauber \and M. Hein} % if too long for running head

\institute{$^*$ indicates shared first author.\\
Department of Computer Science, University of T{\"u}bingen
}

\maketitle

\begin{abstract}
Modern neural networks are highly non-robust against adversarial manipulation. A significant amount of work has been invested in techniques to compute lower bounds on robustness through formal guarantees and to build provably robust models. However, it is still difficult to get guarantees for larger networks or robustness against larger perturbations. Thus attack strategies are needed to provide tight upper bounds on the actual robustness.
%just evaluated using existing attack strategies.
We significantly improve the randomized gradient-free attack for ReLU networks \cite{CroHei18}, in particular by scaling it up to large networks. We show that our attack achieves similar or significantly smaller robust accuracy than state-of-the-art attacks like PGD or the one of Carlini and Wagner, thus revealing an overestimation of the robustness by these state-of-the-art methods.
Our attack is not based on a gradient descent scheme and in this sense gradient-free, which makes it less sensitive to the choice of hyperparameters as no careful selection of the stepsize is required. 

\keywords{Adversarial attacks}
\end{abstract}

\section{Introduction}
Recent work has shown that state-of-the-art neural networks are non-robust \cite{SzeEtAl2014,GooShlSze2015}, in the sense that a small adversarial change of a (even with high confidence) correctly classified input leads to a wrong decision again potentially with high confidence. While \cite{SzeEtAl2014,GooShlSze2015} have brought up this problem in object recognition tasks, the problem itself has been discussed for some time in the area of email spam classification \cite{DalEtAl2004,LowMee2005}. However, since machine learning is nowadays used as a component for automated decision making in safety critical systems e.g. autonomous driving or medical diagnosis systems, fixing this problem should have high priority as it potentially can lead to fatal failures beyond the eminent security issue \cite{LiuEtAl2016}.\\

While a lot of research has been done on attacks and defenses \cite{PapEtAl2016a,LiuEtAl2016,KurGooBen2016a,YuanEtAL2017} it has been shown that all existing defense strategies can be broken again \cite{CarWag2017,AthEtAl2018}, with two exceptions. The first one are methods which provide provable guarantees on the robustness of a network
\cite{KatzEtAl2017,TjeTed2017,HeiAnd2017,RagSteLia2018,WonKol2018,MirGehVec2018,WengEtAl2018,SchEtAl19,CroEtAl2018} and which have proposed new ways of training \cite{WonKol2018,MirGehVec2018} or of regularizing neural networks \cite{HeiAnd2017,CroEtAl2018} to make them more robust. While this area has made huge progress it is still difficult to provide such guarantees for medium-sized networks \cite{WonKol2018,MirGehVec2018}. Then the only way to evaluate robustness for large networks is still to use successful attacks which thus provide, for every clean input, an upper bound on the norm of the minimal perturbation necessary to change the class. In fact, this is an approach to the problem of estimating robustness symmetric to formal certificates, which are lower bounds on the actual robustness. The first attack scheme based on L-BFGS has been proposed in \cite{SzeEtAl2014}, afterwards research has produced a variety of adversarial attacks of growing effectiveness \cite{GooShlSze2015,HuaEtAl2016,MooFawFro2016}. However, it has been recognized that simple attacks often fail when they face a defense created against the specific attacks but which can be easily broken again using other more powerful techniques \cite{CarWag2017,AthEtAl2018}.
Apart from these white-box attacks (model is known at attack time), also several black-box attacks have been proposed \cite{LiuEtAl2016,NarKas16,BreRauBet18}.\\
The second exception is adversarial training with a relatively powerful attack \cite{MadEtAl2018} based on projected gradient descent (PGD). This defense technique could not be broken even using the state-of-the-art attack of Carlini and Wagner \cite{CarWag2016,CarWag2017,AthEtAl2018}.\\

In this paper we extend the white-box attack scheme proposed in \cite{CroHei18}, originally designed to attack fully-connected neural networks using ReLU type activation function.
It is well known that these networks result in continuous piecewise affine functions \cite{AroEtAl2018}, that is the domain is decomposed into linear regions given as polytopes on which the classifier is affine. The principle of the attack of \cite{CroHei18} is then to solve the minimal adversarial perturbation problem on each linear region as it boils down to a convex
optimization problem. In \cite{CroHei18} they report that the attack outperforms the DeepFool attack \cite{MooFawFro2016} and the state-of-the-art Carlini-Wagner attack (CW) \cite{CarWag2016} by up to $9\%$ relative improvement in the norm of the smallest perturbation $\delta$ needed to change the classifier decision. However, the attack has been limited to
small fully-connected neural networks with up to 10000 neurons. In this paper we show that this attack can also be applied to convolutional, residual and dense networks with piecewise affine activation functions as well as max and average pooling layers and scales to networks consisting of more than 2.5 million neurons and achieving state-of-the-art performance.\\
The main contributions of this paper are 1) an upscaling of the attack to large networks so that it can be applied to standard networks for CIFAR-10 and 2) supporting now the most common types of layers, e.g. convolutional and residual ones. The key for the upscaling is a very fast solver for the dual of the quadratic program which has to be solved for finding minimal $l_2$ perturbations. The employed accelerated gradient descent scheme achieves quickly medium accuracy, which is enough for our purposes. Moreover, we use the fact that the solver just needs matrix-vector products of the constraint matrix and thus the explicit computation of the constraint matrix is not needed. This leads to a small memory footprint so that we can use this solver directly on the GPU. Finally, compared to \cite{CroHei18} we have designed a more efficient sampling scheme of the next region to be checked. All these speed-ups together allow us now to attack networks as long as they basically fit into GPU memory. In this paper the largest network has over 2.8 million neurons, which is 280 times more than in \cite{CroHei18}. We show that in most of the cases our attack  performs at least as good as the best attack among PGD \cite{MadEtAl2018}, DeepFool \cite{MooFawFro2016} and CW \cite{CarWag2016}. In particular our algorithm works well across architectures, datasets and training schemes, being always the most successful or very close to the best of the competitors. Notably, we show especially for models trained with adversarial training \cite{MadEtAl2018} against the $l_\infty$-norm and provably robust models \cite{CroEtAl2018,WonKol2018} that the other attacks overestimate, partially by large margin, the robustness wrt to the $l_2$-norm.\\
We thus recommend our attack if a reliable estimation of the real robustness of a network is needed, as our attack not only performs well on average but does not, unlike the established state-of-the-art attacks (PGD, DeepFool, CW), lead to gross overestimation of the robustness of the network in some cases.

\section{Piecewise affine formulation of ReLU networks} \label{sec:PAF}
It has been noted in \cite{AroEtAl2018,CroHei18} that ReLU networks, that is networks that use only the ReLU activation function, result in continuous piecewise affine classifiers in the form $f:\R^d \rightarrow \R^K$, where $d$ is the input dimension and $K$ the number of classes. This implies that there exists a \textit{finite} set of polytopes $Q_r$, with $r=1,\ldots,R$, such that on each polytope the classifier is an affine function, that is there exists $W \in\R^{K \times d}$ and $b \in \R^K$ such that $f(x)=Wx+b$ holds for $x \in Q_r$.
Note that, although we here focus on ReLU, the same property hold for any piecewise affine activation function, like Leaky ReLU.
In the following we generalize the construction from \cite{CroHei18} done for fully-connected networks to the case of other layer types, so that it extends to convolutional networks (CNNs), ResNets \cite{HeZhaRen2015} and DenseNets \cite{HuaEtAl16a}.\\

We can write ReLU networks with $L$ hidden layers as the composition of $L+1$ functions, each standing for one of the layers (including the output layer), $f^{(1)},\ldots, f^{(L+1)}$. Note that we consider the application of the activation function as a stand-alone layer (called ReLU layer). Denoting with $n_j$, for $j=1,\ldots,L+1$, the number of units in layer $j$ (in particular $n_{L+1}=K$ and we assume $n_0=d$), we can define, for $j=1,\ldots,L+1$, \begin{equation} f^{(j)}: \R^{n_0}\times \R^{n_1} \times \ldots \times \R^{n_{j-1}} \longrightarrow \R^{n_j} \end{equation} and the output of the $j$-th layer is obtained as \begin{equation} x^{(j)}=f^{(j)}(x,x^{(1)},\ldots,x^{(j-1)}), \label{eq:layers_output} \end{equation} where $x$ is the input of the network.
Then, the final output of the classifier $f$ is given by \begin{equation} f(x) \equiv x^{(L+1)} = f^{(L+1)}(x,x^{(1)},\ldots,x^{(L)}). \end{equation} If we make explicit the relation between each $x^{(j)}$ and the input $x$ we can recover the formulation of classifier $f$ as a function from $\R^d$ to $\R^K$. While this definition of a classifier differs from the usual recursive formulation, it allows to handle also connections between non-consecutive layers (as it happens for residual networks \cite{HeZhaRen2015}). Finally, the class $c$ which is assigned to $x$ is given by \[c=\argmax_{r=1,\ldots,K}f_r(x) = \argmax_{r=1,\ldots,K} x^{(L+1)}_r. \]
\\
Every layer has one of the following types: dense, convolutional, skip connection, ReLU (or leaky ReLU), avg-pooling, max-pooling, batch normalization. We now show how it is possible to rewrite each of them, $f^{(j)}$, $j=1,\ldots,L+1$, as an affine function \begin{equation} \begin{split} &\A^{(j)}: \R^{n_0}\times \R^{n_1} \times \ldots \times \R^{n_{j-1}} \longrightarrow \R^{n_j},\\ &\A^{(j)}(t)=A^{(j)}t + a^{(j)}, \end{split} \end{equation} with $A^{(j)}\in \R^{n_j \times N_j}, a^{(j)}\in\R^{n_j}, N_j=\sum_{i=0}^{j-1}n_i$, in the linear region $Q(x)$ corresponding to the input $x$ (see below for the definition). {%\color{blue}
For simplicity in the following we call $y = (x, x^{(1)}, \ldots, x^{(j-1)})$ the fixed input of the layer $f^{(j)}$ we are considering (see Equation \eqref{eq:layers_output}).}\\
First, let us notice that dense, convolutional, skip connection and batch normalization (at inference time) layers are already affine operations, which means $\A^{(j)}\equiv f^{(j)}$.\\
Second, ReLU layers apply the function $\sigma(t)=\max\{0,t\}$ componentwise to the output of the previous layer. Thus, they can be, noticing that $y$ is defined so that the last $n_j$ components correspond to $x^{(j-1)}$, replaced by linear functions explicitly represented by the matrices $\Sigma \in \R^{n_j\times N_j}$ defined as %\[\Sigma = \left(\begin{array}{c c c c c c c} h(x^{(j-1)}_1) & 0 & \ldots & 0 & 0 & \ldots & 0\\ 0 & h(x^{(j-1)}_2) & \ldots & 0 &  0 & \ldots & 0\\ \vdots & \vdots & \ddots & \vdots & \vdots & \ddots & \vdots \\ 0 & 0 & \ldots & h(x^{(j-1)}_{n_{j}}) & 0 & \ldots & 0 \end{array}\right), \]
\[\Sigma = \left(\begin{array}{c c c c c c c}0 & \ldots & 0& h(x^{(j-1)}_1) & 0 & \ldots & 0\\ 0 & \ldots & 0 & 0 & h(x^{(j-1)}_2) & \ldots & 0\\ \vdots & \ddots & \vdots& \vdots & \vdots & \ddots & \vdots \\ 0 & \ldots & 0& 0 & 0 & \ldots & h(x^{(j-1)}_{n_{j}}) \end{array}\right), \]
with \[h:\R\longrightarrow \R, \quad h(t)=\left\lbrace \begin{array}{l l}0 & \textrm{if } t<0 \\ 1 &\textrm{else}\end{array}\right. . \] Then, the desired affine function is $\A^{(j)}(t)=\Sigma t$. {%\color{blue}
Since $\Sigma$ depends on the input of the layer, $A^{(j)}$ and $\A^{(j)}$ are not shared by all the input points.}\\
Third, average pooling computes the mean over certain subsets of the input vector. For example, the average of the first four entries of $y$ is obtained, introducing  
\[ a=\left(\frac{1}{4}, \frac{1}{4}, \frac{1}{4} ,\frac{1}{4}, 0, \ldots, 0 \right) \in \R^{N_j},\] as $\inner{a,y}$. Then, since we have $n_j$ pools of $p$ elements, it is sufficient to create $n_j$ vectors similar to $a$, with entries equal to $\nicefrac{1}{p}$ in the positions of the elements we want to average and zero else. We use then these vectors as rows of the matrix $A^{(j)}$, getting $\A^{(j)}(t)=A^{(j)}t$ . We notice that $\A^{(j)}$ does not depend on the input $y$ (as avg-pooling is already an affine function).\\
Finally, the construction of $A^{(j)}$ for max-pooling layers is analogue (as these layers return the maximum instead of the mean). The main difference is that in this case $A^{(j)}$ may change as $y$ does. In fact, going back to our example, if we want to extract the maximum among the first four entries of $y$ and assume that it is realized by the second component, we can set $a=(0,1,0,0,\ldots,0)\in \R^{N_j}$. If the position of the maximum changes also the vector $a$ changes. Again, $\inner{a,y}$ returns the value we are interested in. If $p_1,\ldots, p_{n_j}$ are the positions of the maxima for each of the $n_j$ pools, we can then build $A^{(j)}\in\R^{n_j\times N_j}$ as \[A^{(j)}_{rs}=\left\lbrace \begin{array}{l l}1 & \textrm{if } s=p_r \\ 0 &\textrm{else}\end{array}\right., \quad s=1,\ldots, N_j, \quad r=1,\ldots,n_j, \] so that $\A^{(j)}(t)=A^{(j)}t$.
Please notice that, similarly to the case of ReLU layers, avg- and max-pooling layers usually involve only the output of the immediately preceding layer.\\
\\
Once we have computed the affine functions $\A^{(j)}$ for every $j=1,\ldots,L+1$ we can explicitly derive recursively the affine functions $\V^{(j)}:\R^d \longrightarrow \R^{n_j}$, represented by the matrices $V^{(j)}\in\R^{n_j\times d}$ and vectors $v^{(j)}\in\R^{n_j}$, satisfying the conditions \begin{equation} f^{(j)}(x,x^{(1)},\ldots,x^{(j-1)}) \equiv \V^{(j)}(x) =V^{(j)}x+v^{(j)}. \label{eq:lin_input}\end{equation}
Let us start with $j=1$, that is the first layer. Then $V^{(1)}$ and the $v^{(1)}$ are the linear function and the bias which define $\A^{(1)}$, namely $A^{(1)}$ and $a^{(1)}$.\\
Assuming now that $\V^{(l)}$ are available for $l=1,\ldots, j-1$, we get $\V^{(j)}$ combining Equations \eqref{eq:layers_output} and \eqref{eq:lin_input} and the definition of $\A^{(j)}$, so that \begin{equation} \V^{(j)}(x)= \A^{(j)}(x,\V^{(1)}(x),\ldots,\V^{(j-1)}(x)),\end{equation} which is affine as a composition of linear and affine functions is affine again.\\
\\
It still remains to compute the polytope $Q(x)$ containing $x$ on which all the previous affine approximations hold exactly. First, note that $\A^{(j)}$ is independent of its input $y$ (and thus from the input $x$ of the network as well) if $j$ is either a dense, convolutional, residual, avg-pooling or batch normalization layer, meaning that $\A^{(j)}$ is equivalent to $f^{(j)}$ on the whole input space. Thus these layers do not contribute to the definition of $Q(x)$. We are left to define where the linear reformulations of ReLU and max-pooling layers hold.
As we noticed above, these kinds of layers only take into account the output of the immediately previous layer. Therefore, while considering layer $j$, we are allowed to act like the only input of $f^{(j)}$ was $x^{(j-1)}$.\\
Let $f^{(j)}$ be a ReLU layer and notice that the matrix $\Sigma$ computed for $x^{(j-1)}$ is the same for any vector whose components have the same sign as those of $x^{(j-1)}$. Defining $\delta$ elementwise as \[\delta_r=\sgn (x^{(j-1)}_r), \quad r=1,\ldots,n_{j-1},\] %\textcolor{blue}{
with the convention $\sgn(0)=1$\footnote{The case $x^{(j-1)}_r=0$ implies that the region on which the affine approximation holds has dimension smaller than that of the input space. Setting $\sgn(0)=1$ we consider a polytope which contains as a face the hyperplane defined by the condition $x^{(j-1)}_r=0$.},%}
we then get the set \[\begin{split} S^{(j)}(x^{(j-1)})= \{z\in\R^{n_{j-1}}\; | \;&\sgn (z_r)=\delta_r,\\ & r=1,\ldots,n_{j-1} \}\end{split}\] containing the points of $\R^{n_{j-1}}$ which lead to the same matrix $\Sigma$ as $x^{(j-1)}$.%We can assume that $x^{(j-1)}_r\neq 0$ for every $r$ since the subset of the input space of each layer on which that does not hold has measure zero.
We note that the condition $\sgn (z_r)=\delta_r$ is equivalent to {%\color{blue}
$z_r\delta_r\geq 0$} and that we are interested in the intersection of $S^{(j)}$ and the domain of layer $j$. With \eqref{eq:lin_input}, we define the polytope on which $f^{(j)}$ is affine,
\begin{align}\label{eq:const}
 \begin{split} Q^{(j)}(x)=\{ z\in\R^d\; \big|\; &\delta_r \V^{(j-1)}(z)\geq 0, \; r=1,\ldots,n_{j-1} \}\\
                         =\{ z\in\R^d\; \big|\; &\delta_r( V^{(j-1)}z + v^{(j-1)})\geq 0,\\ &r=1,\ldots,n_{j-1} \}. \end{split}
 \end{align}
The set $Q^{(j)}(x)$ defines the region of the input space containing $x$ and where $\A^{(j)}(x)$ and thus $f^{(j)}$ is an affine function.\\
If $f^{(j)}$ is instead a max-pooling layer, we can see that $A^{(j)}$ is preserved as long as the maximum within each pool is realized at the same position. We can denote the $n_j$ pools as the sets $P^1,\ldots,P^{n_j}$, whose elements are the indices of the components of the input (of the max-pooling layer) involved in the pool. Moreover we define for every $r=1,\ldots,n_j$ \[p^r_{max}=\argmax_{i\in P^r} x^{(j-1)}_{i}, \] that is the index of the component of $x^{(j-1)}$ attaining the maximum for each pool $P^r$.
Then, for $r=1,\ldots,n_{j}$, \[S^{(j)}_r(x^{(j-1)}) =\{z\in\R^{n_{j-1}} \;|\; z_{p^r_{max}}\geq z_i,\; \forall i \in P^r \} \] is the set of the vectors in $\R^{n_{j-1}}$ preserving the position of the maximum computed at $x^{(j-1)}$ for pool $P^r$. Similar to what has been done for ReLU layers, we define
\begin{align}\label{eq:const_max}
\begin{split} Q^{(j)}_r(x) = &\{z\in \R^d \;|\; \V^{(j-1)}_{p^r_{max}}(z)\geq \V^{(j-1)}_i(z),\; \forall i \in P^r \} \\ =&\left\lbrace z\in \R^d \;|\; \left(V^{(j-1)}_{p^r_{max}} -V^{(j-1)}_i\right)z\right. \\  & \qquad \left. + v^{(j-1)}_{p^r_{max}}- v^{(j-1)}_i \geq 0,\; \forall i \in P^r \right\rbrace,
\end{split}
\end{align}
so that, finally, $Q^{(j)}(x)=\bigcap_{r=1}^{n_j} Q^{(j)}_r (x)$ is the subset of the input space containing $x$ on which $f^{(j)}$ is an affine function.\\
Note that $Q^{(j)}(x)=\R^d$ if $j$ is neither a ReLU nor a max-pooling layer. The polytope $Q(x)$ on which $f^{(L+1)}$ (and all layers below) is affine is given by \[Q(x)=\bigcap_{j=1}^{L+1} Q^{(j)}(x).\] In the following we refer to $Q(x)$ as the \emph{linear region} of $x$. Note also that the intersection of $Q(x)$ with any other polytope is still a polytope (e.g. this is necessary when the input domain of a classifier is a subset of $\R^d$). Note that the explicit storage of the matrices $V^{(j)}$ is not possible for 
large networks and high input dimension as one needs $O(Nd)$ memory. In Section \ref{sec:strategy} it will turn out that our attack algorithm only requires matrix-vector products $V^{(j)}x$
which can be done without computing $V^{(j)}$ explicitly and thus we can do the whole attack on the GPU as long as the network itself fits into GPU memory.

\section{Minimal adversarial perturbation inside a linear region}\label{sec:adversarial}
Classifiers based on neural networks have been shown to be vulnerable to \emph{adversarial samples}, that is they misclassify inputs which are almost indistinguishable from an original, correctly recognized test image \cite{SzeEtAl2014,GooShlSze2015}. The minimal adversarial perturbation $\delta$ wrt an $l_p$-norm is defined as the solution of the following optimization problem
\begin{equation}\label{eq:advopt} \begin{split} \minop_{\delta \in \mathbb{R}^d} \; \norm{\delta}_p \quad \textrm{s.th.}  \quad    &\maxop_{l\neq c} \; f_l(x+\delta) \geq f_c(x+\delta),\\ & x+\delta \in C, \end{split}\end{equation}
with $C$ being a set of constraints the input of $f$ has to satisfy (in the following we assume that $C$ is a polytope), e.g. images scaled to be in $[0,1]^d$, $x\in\R^d$ is the original point and $c$ the class assigned to $x$ by $f$ (we assume $x$ is correctly classified by $f$). The $l_p$-norm of $\delta$ measures the difference between original and adversarial inputs (changing $p$ leads to adversarial samples with different properties). In practice, one often uses $p=2$ or $p=\infty$. We concentrate for simplicity in this paper on $p=2$, even though the framework allows to handle any $p$-norm given that a fast solver is available for the following linearized problem \eqref{eq:advopt_lin_targeted}. Note that \eqref{eq:advopt} represents an untargeted attack, that is we just want that the decision changes but we do not want to achieve that $x+\delta$ is classified as a particular class.\\
\\
The optimization problem \eqref{eq:advopt} is in general non-convex and NP-hard \cite{KatzEtAl2017}. However, as shown in \cite{CroHei18}, one can solve it efficiently inside every linear region of the classifier, that is if we add to \eqref{eq:advopt} the constraint $x+\delta \in Q(y)$, where $Q(y)$ is the linear region which contains the point $y \in \R^d$. In fact, recalling Section \ref{sec:PAF}, we introduce for $l\neq c$ the vectors $\delta_l$ as the solutions of the $K-1$ convex problems (note that we assume that $C$ is a polytope) \begin{equation} \begin{split} \minop_{\delta \in \mathbb{R}^d} \; \norm{\delta}_p \quad \textrm{s.th.} \quad & \inner{V_l^{(L+1)}-V_c^{(L+1)},x+\delta}\\ &+ v^{(L+1)}_l-v^{(L+1)}_c \geq 0, \\ & x+\delta \in C \cap Q(y). \end{split} \label{eq:advopt_lin} \end{equation}  Then, the solution of Problem \eqref{eq:advopt} restricted to the linear region $Q(y)$ is $\argmin_{\{\delta_l: l\neq c\}}\norm{\delta_l}_p$. 
While we are mainly interested in untargeted attacks, we would like to highlight that targeted attacks against any of the classes $s\neq c$ are easily possible by solving instead the following problem: 
\begin{align}\label{eq:advopt_lin_targeted} \begin{split} \minop_{\delta \in \mathbb{R}^d} \; \norm{\delta}_p \quad \textrm{s.th.} \quad  & \inner{V_s^{(L+1)}-V_r^{(L+1)},x+\delta} \\ &+ v^{(L+1)}_s-v^{(L+1)}_r \geq 0, \; \forall r \neq s,\\ & x+\delta \in C \cap Q(y). \end{split} \end{align} 
Please note that if one would solve \eqref{eq:advopt_lin} for all possible linear regions and take the smallest perturbation, then this the exact solution of \eqref{eq:advopt}. However, due to the extremely large number of linear regions this is infeasible in practice. Thus we use a randomized scheme for selecting the next linear region which is described in Section \ref{sec:strategy} together with a description of the particular solver for the resulting quadratic program in \eqref{eq:advopt_lin} for the choice of $p=2$.
%In the next section we introduce the overall scheme and focus on solving \eqref{eq:advopt_lin} for $p=2$ for which we develop a specific QP solver adapted to this problem.

\section{Generation of adversarial samples through randomized local search} \label{sec:strategy}
In the following we present an improved selection scheme of the linear regions compared to the one in \cite{CroHei18}. The observation motivating our scheme is that the decision surface dividing areas of the input space assigned to different classes extends continuously across neighboring linear regions. If a point, say $y$, lying on the decision boundary is available, it is highly likely to find in its vicinity other points, again on the decision boundary between two classes, closer than $y$ to the target image $x$. However, as pointed out in \cite{CroHei18} it is very difficult to determine neighboring regions as a large number of the constraints defining the polytope are active at the solution of \eqref{eq:advopt_lin}. In this case the neighboring region is not unique and checking all of them is infeasible and inefficient.\\
Thus we sample random points (more details below) in a small ball centered around the currently best point $y$, that is realizing the smallest adversarial perturbation found so far, and then solve \eqref{eq:advopt_lin} in the corresponding linear region until we find a better adversarial sample.\\
Moreover, we save the activation patterns of the linear regions we have explored. Before checking a point and its corresponding linear region we compare the activation pattern to the ones of the points which we have already visited. If the activation patterns agree it means that the two points belong to the same region and then we can skip checking it again.\\
\\
Algorithm \ref{alg:algorithm-label} shows our overall attack for a general $l_p$-norm trying to solve the optimization problem \eqref{eq:advopt} for the minimal adversarial perturbation. In the experiments we use either $N=400$ or $N=500$, that is we check $400$ resp. $500$ linear regions.
Please note that Algorithm \ref{alg:algorithm-label} requires to be fed with a feasible point $\delta_{WS}$ of \eqref{eq:advopt}. There are several possibilities e.g. an adversarial sample of a fast attack like DeepFool as has been used in \cite{CroHei18}. In this paper we prefer
to be independent of another attack. Thus we are using the following scheme to choose $M$ starting points. At $x$ we rank the classes $\{1,\ldots,K\}$ according to the components of corresponding classifier output $f(x)$ in descending order {%\color{blue}
$\rho$, where $\rho_1$ is the class which is assigned to $x$. We choose the $M$ classes $\rho_2,\ldots,\rho_{M+1}$ in the ranking and compute the point $z_j$ in the training set correctly classified by $f$ in class $\rho_j$} which is closest to $x$ for $j=2,\ldots, M+1$. In order to be speed up the attack we do for each $z_j$ a binary search on
$[x,z_j]$ and identify the point $u_j$ which is closest to $x$ but is classified differently from $x$ and use $\delta^{(j)}_{WS}=u_j-x$, $j=2,\ldots,M+1$, as starting perturbations for Algorithm \ref{alg:algorithm-label}.\\ %In the experiments we always use $M=5$.
\begin{algorithm}[t]
	\SetAlgoNoLine
	\caption{Our attack}
	\label{alg:algorithm-label}
	\SetKwInOut{Input}{Input}
	\SetKwInOut{Output}{Output}
	\Input{$x$ original image, $\delta_{WS}$ starting perturbation, $\gamma,N,p$}
	\Output{$\delta$ adversarial perturbation}
	$\delta \gets \delta_{WS}$, $u\gets \norm{\delta}_p$\\
	\For{$j=1,\ldots,N$}{
		%$\epsilon\gets$ random point in $B(0,1)$\\ $r\gets$ uniformly sampled in $[0,1]$\\
		%$\epsilon\gets \frac{\epsilon}{\norm{\epsilon}_p} r^\gamma$\\
		%$y\gets x + \delta + \epsilon$\\
		$y \gets$ sampled according to \eqref{eq:next_region}\\
		\If{region containing $y$ has not been checked already}{
			computation of $Q(y)$\\
			$\delta_{temp}\gets$ solution of Problem \eqref{eq:advopt_lin} on $Q(y)$\\
			\lIf{$\norm{\delta_{temp}}_p<u$}{$\delta \gets \delta_{temp}$, $u\gets\norm{\delta}_p$}
	}}
\end{algorithm}

At each of the $N$ iterations we sample a point around the current best (smallest $l_p$-norm) feasible point $y:=x+\delta$ of \eqref{eq:advopt}. The following sampling scheme is biased towards $x$, where $q \in [\frac{1}{2},1]$ is a parameter controlling the bias towards $x$ ($q=\frac{1}{2}$ no bias,
$q=1$ maximal bias) and $\gamma >0$ is a parameter controlling how localized our search is (the larger $\gamma$, the more localized). We provide an analysis of the influence of these parameters in Section \ref{sec:parameters}. We sample i) uniformly a point $y^{\perp}$ from the intersection of the unit sphere $\mathcal{S}^{d}$ centered in $y$ and the hyperplane containing $y$ with normal vector $\delta$, and ii) an angle $\theta \in [-\pi,\pi]$ given by \begin{equation} \begin{split} &X_1 \;\textrm{r.v.}:\; \mathbb{P}(X_1=1)=q, \; \mathbb{P}(X_1=-1)=1-q,\\ &X_2\sim \U[0,\pi],\\ &\theta = X_1 X_2,  \end{split} \label{eq:psampl}
\end{equation} where $\U[0,\pi]$ is the uniform distribution on the interval $[0,\pi]$. We define $\delta^{\perp}=y^{\perp}-y$. Note that by construction $\norm{\delta^\perp}_2=1$. Finally, \begin{equation} \begin{split} &\delta_{\textrm{new}}=\cos(\theta)\delta^\perp - \sin(\theta)\frac{\delta}{\norm{\delta}_2},\\ & r_\textrm{new}=\norm{\delta}_2 X_3^\gamma\quad \textrm{with} \; X_3\sim \U[0,1]\end{split} \label{eq:sampl_2} \end{equation} give direction and step size to produce the next point $y_\textrm{new}$ whose linear region will be checked, defined as \begin{equation} y_\textrm{new}= y + r_\textrm{new} \delta_\textrm{new}. \label{eq:next_region}
\end{equation} Note that the larger $\gamma$ the more biased $y_\textrm{new}$ will be towards $y$. On the other hand our sampling scheme makes a difference between the half-sphere centered at $y$ with pole at $x=y-\delta$ versus the half-sphere with pole at $y+\delta$. If $q=\frac{1}{2}$ samples from both half-sphere are equally probable, whereas if $q=1$ one
samples just from the half-sphere pointing towards $x$. At first sight it might look strange that we do not choose $q=1$, as points sampled from the half-sphere pointing away from $x$
have larger distance from $x$ than $y$. However, experiments on a small subset of points show that a value of $q=0.8$ leads to best results even though the difference to $q=1$ is not large and thus we fix it to $q=0.8$ for all experiments. Moreover, we use $\gamma=6$ or $\gamma=9$ for all experiments, noting anyway that the attack is not very sensitive to this value as $\gamma$ in the range between $3$ and $9$ lead to very similar results. {%\color{blue}
In Section \ref{sec:parameters} we provide a detailed analysis of the influence of these two parameters on the performance of our scheme.}\\
%In this way $y_\textrm{new}$ is biased to be more aligned with the direction joining $x$ and $y$, while $p>\frac{1}{2}$ allows to get a larger fraction of sampled points lying in the half space, identified by the hyperplane through $y$ and orthogonal to $\delta$, which contains the target image $x$. In practice $p=0.8$ is our choice for the experiments.\\
\\
If $C$ is a polytope e.g. $C=[0,1]^d$, then the optimization problems \eqref{eq:advopt_lin} and \eqref{eq:advopt_lin_targeted} are equivalent to linear programs (LP) for $p=\infty$ and $p=1$ and equivalent to a quadratic program (QP) for $p=2$. The main cost of the attack is to solve the optimization problem. Next we describe an
efficient scalable way of solving \eqref{eq:advopt_lin} for $p=2$, avoiding the explicit calculation of the linear regions. %and will comment on potential extensions to $p=\infty$.

\subsection{A scalable and efficient solver for the quadratic program}\label{sec:QP_solver}
Let us suppose $x+\delta$ is our current best found solution, then we would like that the solution of \eqref{eq:advopt_lin} produces a new $\delta'$ which satisfies
$\norm{\delta'}_2 < \norm{\delta}_2$. This implies that as soon as we have a certificate that the optimal value of \eqref{eq:advopt_lin} is larger than $\norm{\delta}_2$ then
we can stop the solver as checking this region will not yield an improvement. Thus we work with the dual of  \eqref{eq:advopt_lin} as the dual objective is always a lower bound
on the primal objective. As soon as we have found dual parameters realizing a larger dual objective than $\norm{\delta}_2$ we can stop.\\
\\
In the following we describe first how we solve the generic resulting dual QP using accelerated gradient descent together with coordinate descent in a subset of the variables. 
Then we describe how this algorithm for solving the QP can be efficiently implemented on the GPU without having to ever to compute the constraint matrix.
Note that in \cite{CroHei18} we used the commercial package Gurobi for solving the QP on the CPU. Now we present an own implementation fully running on the GPU which is roughly three orders of magnitude faster than then our old implementation on the CPU and which allows us to deal with fully-connected, convolutional and residual layers. 

%\textbf{Solving the dual problem:}
\paragraph{Solving the dual problem.}
As we are mainly interested in applications in computer vision  we specialize to the case $C=[0,1]^d$ in \eqref{eq:advopt_lin}, which can then be formulated as
\begin{align}\label{eq:primal} \minop_{z \in \R^d} \; \norm{z-x}^2_2 \quad \textrm{s.th.} \quad  Az\leq b, \quad  z \in [0,1]^d. \end{align}
Note that the formulation is different from \eqref{eq:advopt_lin} but can be transformed into each other using $\delta=z-x$. The chosen formulation of the optimization problem
in \eqref{eq:primal} is better adapted to the componentwise constraints imposed by $C$.
The primal problem is strongly convex and thus has a unique solution.
We derive the dual problem as
\begin{align}\label{eq:dual} \begin{split} \maxop_{\alpha,\beta \in \R^d, \mu \in \R^m} \; q(\mu,\alpha,\beta) \quad \textrm{s.th.} \quad & \alpha \geq 0,\; \beta \geq 0, \;   \mu \geq 0, \end{split} \end{align}
where 
\begin{align*} q(\mu,\alpha,\beta)=&-\frac{1}{2}\norm{A^T \mu +\alpha-\beta}^2_2 + \inner{A^T\mu +\alpha-\beta,y}\\
&-\inner{\alpha,\ones}-\inner{\mu,b} \end{align*}
and the inequalities in \eqref{eq:dual} are componentwise. The correspondence between the primal variable $z$ and the dual optimal variables $\alpha,\beta,\mu$ is
given by
\begin{align}\label{eq:primal2}
 z = y -A^T \mu -\alpha + \beta. 
\end{align}
Note however that even for dual feasible $\alpha,\beta$ and $\mu$, the primal variable $z$ need not to be feasible.
The KKT conditions are 
\[ \alpha_i (x_i-1)=0, \quad \beta_i x_i =0, \quad \mu_i ( (Ax)_i - b_i)=0.\]
This implies $\alpha_i\beta_i=0$. Solving for $\alpha,\beta$ yields
\begin{align}\label{eq:alphabetaoptimal}
\alpha = \max\{0, y-A^T\mu-\ones\}, \beta=\max\{0,A^T\mu-y\}.
\end{align}
Thus for fixed $\mu$ we can directly find the optimal values of $\alpha$ and $\beta$. The dual problem is also a quadratic program
%with matrix
%\[ \begin{pmatrix} \Id & 0 & 0\\ 0 & \Id & 0 \\ 0 & 0 & AA^T \end{pmatrix},\]
but it is not necessarily strongly convex as $AA^T$ does not need to be positive definite. However, the 
gradient 
\begin{align*}
\nabla_\mu q    &%= -A (A^T \mu+\alpha-\beta) + Ay -b
                = -AA^T \mu + A (y-\alpha+\beta)-b\\
\nabla_\alpha q &= A^T \mu+\alpha-\beta + y -\ones\\
\nabla_\beta q  &= -(A^T \mu+\alpha-\beta) - y
\end{align*}
is Lipschitz continuous and the Lipschitz constant $L$ can be upper bounded as,
\begin{align} L \leq \max\{ \norm{A^T A}^2, \norm{A^T A},1\}.\end{align}% = \begin{cases} 1 & \textrm{ if } \norm{A}\leq 1,\\ \norm{A}^2 & \textrm{ else}.\end{cases}\] 
We estimate $\norm{A^T A}^2$ via the power method with $20$ iterations, which is enough to get already
a quite accurate estimate. We solve the QP itself with accelerated projected gradient descent \cite{Nes1983,BecTeb2009,ChaPoc2011} in $\mu$ by setting $\alpha$
and $\beta$ to their optimal values for given $\mu$ as in \eqref{eq:alphabetaoptimal} which can be seen as a mixture of a coordinate descent in $\alpha,\beta$
and accelerated projected gradient descent in $\mu$. Note that in all steps we never need the matrix $A$ explicitly, but just matrix vector products $A^T\mu$
or $Az$ if we want to compute feasibility of the current primal variable $z$. Even for the computation of $\norm{A^T A}$ we use the power method which also only
requires matrix vector products. The only caveat is a good pre-conditioning of the problem, which can be achieved by normalizing the rows $a_i$, $i=1,\ldots, N$ 
of $A$ to have unit norm (with corresponding rescaling of $b$). One can compute them via matrix vector products $a_i=A^T e_i$, but this would require too many
of them. We discuss how this can be resolved in the next section and how the whole QP solver can be ported to the GPU.

\subsection{Solving the QP efficiently on the GPU without explicit computation of the constraint matrix $A$}
As discussed at the end of the previous section, the QP solver via accelerated gradient descent does not require the explicit computation of $A$ as long as there
is a way to compute matrix vector products $A^T\mu$ and $Az$ efficiently. While in \cite{CroHei18} the matrix $A$ has been explicitly computed on the CPU, this is no longer 
feasible for larger networks as the memory consumption is $O(Nd)$, where $d$ is the input dimension and $N$ the total number of neurons. Even if one uses sparse
matrix formats e.g. in the case of convolutional layers, this does not help to reduce the required memory significantly if the network is deep. Moreover, also the computation of the hyperplanes requires a computational cost equivalent to $d$ forward passes of the network.\\
Thus a major improvement of this paper compared to \cite{CroHei18} is the transfer of all computations from the CPU to the GPU which is only possible if the matrix $A$ is not 
explicitly computed as the GPU memory would not suffice for this. The major insight to do this is that accelerated gradient descent only requires matrix-vector products
of the form $A^T \mu$ and $Az$. Note that $A$ contains basically the concatenated matrices $V^{(j)}$ from \eqref{eq:const} and \eqref{eq:const_max}. However, we note that
according to \eqref{eq:lin_input} it holds
\[ f^{(j)}(x,x^{(1)},\ldots,x^{(j-1)})=\V^{(j)}(x) = V^{(j)}x + v^{(j)},\]
and thus $V^{(j)}$ is nothing else than the Jacobian $Jf^{(j)}$ of $f^{(j)}$ with respect to $x$ and 
\[ v^{(j)}=f^{(j)}(x,x^{(1)},\ldots,x^{(j-1)})-V^{(j)}x.\] 
Suppose for simplicity that 
\[ f^{(j)}(x)=g_j(g_{j-1}(\ldots(g_1(x))\ldots)).\]
Then the Jacobian $Jf^{(j)}$ of $f^{(j)}$ at $x$ is given by the chain rule as
\begin{align*}
  V^{(j)} &= Jf^{(j)}\big|_x\\
          & = Jg_j\big|_{g_{j-1}(x)}Jg_{j-1}\big|_{g_{j-2}(x)}\cdots Jg_1\big|_x.
\end{align*}
Note that $V^{(j)}u$ can be evaluated as
\begin{align*}
 V^{(j)}u &= Jf^{(j)}\big|_x u \\
          %&= Jg_j(g_{j-1}(x))\Big( J_{g-1}(g_{j-2}(x))\Big(\ldots \Big(Jg_1(x)u\Big)\ldots\Big)\Big).
          &= Jg_j\big|_{g_{j-1}(x)}\Big( J_{g-1}\big|_{g_{j-2}(x)}\Big(\cdots \Big(Jg_1\big|_x u\Big)\cdots\Big)\Big).
\end{align*}
In the same way we can compute $w^T V^{(j)}$ as
\begin{align*}
 w^T V^{(j)} &= w^T Jf^{(j)}\big|_x \\
             %&= \Big(\ldots \Big(w^T Jg_j(g_{j-1}(x))\Big)Jg_{j-1}(g_{j-2}(x))\Big)\ldots\Big)Jg_1(x)
             &= \Big(\cdots \Big(w^T Jg_j\big|_{g_{j-1}(x)}\Big)Jg_{j-1}\big|_{g_{j-2}(x)}\cdots\Big)Jg_1\big|_x.
\end{align*}
Thus calculating $V^{(j)}u$ requires a single forward pass through the network and $w^T V^{(j)}$ requires a forward pass for computing the values $g_{j}(x)$ and then a backward
pass through the network. More general, the computation of the Jacobian-vector products can be done via automatic differentiation (forward-mode resp. backward-mode automatic differentiation). Finally, to calculate the above expressions efficiently we still need a fast way to compute $Jg_k\big|_y v$ and $z^T Jg_k\big|_y$
for primitive functions $g_k$ e.g. if $g_k$ is a convolution, then
$Jg_k\big|_y v$ can be computed as well as a convolution and $z^T Jg_k\big|_y $ as the
transposed convolution. Fortunately, modern implementations of automatic
differentiation already come with a large collection of primitive functions and
corresponding rules for $Jg_k(y)v$ and $z^T Jg_k(y)$. Thus, we can directly and efficiently compute
them on the GPU without computing the Jacobians itself. Thus our QP solver does not require much more memory than the network itself which allows it to scale to large networks.\\

Note that for pre-conditioning of $A$ it would make sense to rescale the rows of $A$ to have unit norm (one has to rescale correspondingly also the vector $b$). While every row vector $a_i$ of $A$ can be obtained as $a_i=e_i^T A$ and thus
also just via matrix-vector products, doing this for every row is prohibitively expensive. Thus we use the fact that the norms of the row vectors corresponding to the same hidden layer
have quite similar norms (typically we see increasing norms as one moves from lower to upper layers). Thus we just sample a small number of rows (in our case $10$) of each layer, compute their norms, take the mean of them and use the inverse of that as a rescaling factor for that layer. While this coarse pre-conditioning scheme is worse than if one rescales
every row individually, it is significantly better than not doing any rescaling at all. There is one exception: we upscale the constraint of the decision boundary, as we have found that this leads to faster feasibility of this constraint which is the most important one of all the constraints.\\
Moreover, we do not need an accurate solution of \eqref{eq:advopt_lin} and thus we have found that in practice $500$ iterations of the accelerated gradient descent scheme suffice to
get a reasonable solution. As the primal variable $z$ in \eqref{eq:primal2} obtained from the dual variables need not be feasible, we explicitly check if the output $z$ is an
adversarial sample. If not then we check via a small line search  $x+\alpha(z-x)$, where $\alpha \geq 1$, if it is an adversarial sample as long as $\alpha \norm{z-x}_2 < \norm{\delta}$, where $\norm{\delta}$ is the norm of the perturbation of the currently best adversarial sample $x+\delta$. Finally, this leads to a scheme which is more than three orders of magnitude faster than that in \cite{CroHei18}.

\section{Experiments}\label{sec:exp}
\begin{table*}
	\centering
	\begin{tabular}{C{20mm}| C{12mm} C{12mm} C{12mm} C{12mm} C{12mm} C{12mm}}
		\multicolumn{7}{c}{\textbf{average difference to the best $l_2$ robust accuracy}}\\[4pt]
		model & PGD-1 & PGD-10k & CW-10k & CW-100k & DF & ours\\
		\hline
		\hline
		%&\multicolumn{6}{c}{average distance from the best}\\
		%\hline
		MNIST & 0.2367  &  0.1011  &  0.1701  &  0.1681  &  0.3135  &  \textbf{0.0051}\\
		GTS &0.0361 &   0.0237   & 0.0177  &  0.0172   & 0.0643        & \textbf{0}\\
		CIFAR-10 &0.0693&0.0515&0.0045&\textbf{0.0037}&0.0812&0.0060\\
		%CIFAR-10 &0.0689&0.0540&0.0041&\textbf{0.0033}&0.0808&0.0056
		%0.1123 &   0.0879 &   0.0037  &  \textbf{0.0029}  &  0.0804  &  0.0052\\
	\end{tabular}

\vspace{12pt}
\begin{tabular}{C{20mm}| C{12mm} C{12mm} C{12mm} C{12mm} C{12mm} C{12mm}}
	\multicolumn{7}{c}{\textbf{maximum difference to the best $l_2$ robust accuracy}}\\[4pt]
	model & PGD-1 & PGD-10k & CW-10k & CW-100k & DF & ours\\
	\hline
	\hline
	MNIST & 0.7800  &  0.5040  &  0.6200  &  0.6120   & 0.9000 &   \textbf{0.0500}\\
	GTS & 0.1600 &   0.1260  &  0.0440 &   0.0420  &  0.1140     &    \textbf{0} \\
	CIFAR-10 & 0.2280&0.2040&0.0180&0.0180&0.1220&\textbf{0.0140}\\
	%CIFAR-10 & 0.2280&0.2080&0.0180&0.0180&0.1220&\textbf{0.0140}\\
	%0.3300 &   0.2720 &   0.0180  &  0.0180 &   0.1220 &   \textbf{0.0140} \\
\end{tabular}
\caption{Performances of different attacks. For each dataset, attack and threshold $\epsilon$, we compute the differences between the robust accuracies estimated by an attack and the best one among those of all the attacks. We here report, given dataset and attack, the mean (top) and the maximum (bottom) of these differences across the thresholds. We can see that our attack has the smallest average distance from the best on two of three datasets and always achieves the best maximal distance. Notably, on GTS both mean and maximum for our attack are 0, which means that it gets the lowest robust accuracy for every model and $\epsilon$.}
\label{tab:perf_in_dataset}
\end{table*}

In this section we show that our attack often outperforms the state-of-the-art methods to compute upper bounds on the robust accuracy of a model, which is defined for a given $\epsilon>0$, as the minimal accuracy that the classifier can achieve if each test sample is allowed to be perturbed within a $p$-norm ball of radius $\epsilon$ in order to achieve a misclassification. The smaller the found robust accuracy the stronger is the attack and the less robust is the network. We focus here on the $l_2$-attack. The code for our attack is publicly available\footnote{\path{https://github.com/jonasrauber/linear-region-attack}}. \\ %while we present a particular case which involves $l_\infty$-robustness at the end of the Section.
We show that current state-of-the-art attacks sometimes overestimate the robustness of classifiers. In fact, with our attack we are often able to achieve smaller robust accuracy than our competitors, and even when we do not we never overestimate the robust test accuracy more than 5.0\% compared to the minimal one found by the competitors. In contrast, all the other attacks have cases where they achieve a robust accuracy at least 50.4\% larger than that provided by our method (see Table \ref{tab:perf_in_dataset}). Thus if one just evaluates robustness using the competing attacks, one would consider models robust which are in fact quite non-robust. Our technique does not show a similar weakness in any setting, pointing out how our algorithm is, on one side, able to recover in general small adversarial perturbations and, on the other side, less susceptible to changes in the characteristics of the network. Interestingly, we notice that $l_2$- gradient-based methods suffer especially when attacking models trained with $l_\infty$-adversarial training.\\
We consider three datasets: MNIST, German Traffic Sign (GTS) \cite{GTSB2012} and CIFAR-10 \cite{CIFAR10} (all images are scaled in $[0,1]^d$). On each of them three models are trained, the plain model (\textit{plain}), one with $l_2$-adversarial training (called $l_2$-\textit{at}) and one with  $l_\infty$-adversarial training ($l_\infty$-\textit{at}) {%\color{blue}
(we use the adversarial training scheme of \cite{MadEtAl2018} that is based on the Projected Gradient Descent attack)}. More details about architectures and training are provided below.\\
We compare our attack against: Projected Gradient Descent on the loss function (PGD) \cite{MadEtAl2018}, Carlini-Wagner $l_2$-attack (CW) \cite{CarWag2016} and DeepFool (DF) \cite{MooFawFro2016}. We use two versions of PGD: PGD-1 uses a single starting point, while PGD-10k exploits 10000 restarts, randomly sampled in the $l_2$-ball of radius $\epsilon$ around the original image. This large number of restarts is motivated by a recent paper which could break a certain defense only when using 10000 restarts of PGD \cite{MosEtAl18}. For both PGD versions we set $k=40$ iterations and, if $\epsilon$ is the threshold at which we want to evaluate robust accuracy, we use a step size of $\nicefrac{\epsilon}{4}$. Similarly, we evaluate CW in the implementation of \cite{Cleverhans2017} with 40 binary search steps and either 10000 (CW-10k) or 100000 iterations (CW-100k). We use the DF implementation as in \cite{foolbox}.
\\

Since the objective of PGD is only to find out if there exists an adversarial sample with norm less than the threshold $\epsilon$, it provides directly the robust accuracy at $\epsilon$ (and must be rerun for each threshold $\epsilon$). On the contrary, CW, DeepFool and our attack try to find the minimal adversarial perturbation as in \eqref{eq:advopt}. After running these attacks, we compute the robust accuracy for a given threshold as the fraction of points whose adversarial examples are farther, in $l_2$-distance, than $\epsilon$. Note that we only check correctly classified points for all methods. The obtained values of robust accuracy achieved for all attacks and different thresholds are reported in Tables \ref{tab:l2_mnist} (MNIST), \ref{tab:l2_gts} (GTS) and \ref{tab:l2_cifar10} (CIFAR-10).\\

In order to thoroughly evaluate the effectiveness of an attack, it is necessary to assess average and worst case performance. In this way one can see whether it overfits to some particular model, dataset or training scheme. For every dataset we compute for all thresholds $\epsilon$ the difference between the robust accuracy provided by every attack and the minimal robust accuracy across all the attacks for the fixed threshold. Thus the worse the performance a method achieves, the larger the difference is.
In Table \ref{tab:perf_in_dataset} we report for each attack the mean and the maximal distance from the best accuracy, across the three models and five thresholds $\epsilon$, for each of the three datasets. It can be directly seen from this table that our attack has at the same time the best \textit{worst case performance} for all three datasets and the best \textit{average performance} in two of three datasets with only a tiny difference in the case it is worse.\\
In  particular, we can see how on MNIST the second best attack (PGD-10k) is on average $10.11\%$ worse than the minimal robust accuracy, compared to $0.51\%$ for our attack, while in the worst case it returns a robust accuracy 50.4\% larger than the minimal one versus 5.0\% for our method. On GTS, both average and maximal difference are 0.0\% for our attack, meaning that it always achieves the minimal robust accuracy among all competing methods. Although on CIFAR-10 we cannot match the average result of CW attack, our attack has nevertheless the best \textit{worst case performance}, highlighting the quality of our approach.

\subsection{Main experiments: details}\label{sec:main_exp}
\begin{table}
	\centering
	\begin{tabular}{C{6mm}|C{4mm} | C{6mm} C{6mm} C{6mm} C{6mm} C{6mm} C{6mm}}
		\multicolumn{8}{c}{\textbf{$l_2$ robust accuracy on MNIST}}\\[4pt]
		model & $\epsilon$ & PGD-1 & PGD-10k & CW-10k & CW-100k & DF & ours\\
		\hline
		\hline
		\multirow{6}{*}{\textit{plain}}& 0.0 & \multicolumn{6}{c}{0.984}\\
		%\hdashline[0.5pt/1pt]
		\cdashline{2-8}
		& 0.5 & 0.928 &\textbf{0.926} & \textbf{0.926} &\textbf{0.926} & 0.936 &\textbf{0.926} \\
		& 1.0 & 0.508 & \textbf{0.472}  & 0.474 &0.474 & 0.586 &0.474	\\
		&1.5	& 0.168 & 0.106	& 0.088 & 0.088	& 0.198 & \textbf{0.078}\\
		&2.0	& 0.106 & 0.028	& 0.006 &0.006	& 0.018 & \textbf{0.002} \\
		&2.5	& 0.078 & 0.014	& \textbf{0.000} &\textbf{0.000}	& \textbf{0.000} & \textbf{0.000}\\
		\hline
		\multirow{6}{*}{$l_2$-\textit{at}} & 0.0 & \multicolumn{6}{c}{0.986}\\
		\cdashline{2-8}
		& 1.0 & 0.930 &0.930 & \textbf{0.926} &\textbf{0.926} &	0.938 & \textbf{0.926}\\
		&1.5	& 0.838& \textbf{0.834}	& 0.846 &0.848 	&0.872 & \textbf{0.834}\\
		&2.0	& 0.698 & \textbf{0.672}& 0.706 &0.706 &0.790 & 0.680\\
		&2.5	& 0.468 & \textbf{0.366} &0.466 &0.464 	&0.674 & 0.416\\
		&3.0	& 0.192	& \textbf{0.096} &0.170 &0.172	&0.542 & 0.112\\
		\hline
		\multirow{6}{*}{$l_\infty$-\textit{at}} & 0.0 & \multicolumn{6}{c}{0.984}\\
		\cdashline{2-8}
		& 1.0 & 0.924 & 0.878 & 0.888 &0.888 &0.948& \textbf{0.736}\\
		&1.5	& 0.886 &0.748	& 0.774 & 0.776 &0.932 & \textbf{0.258}\\
		&2.0	& 0.812 &0.536	& 0.652 & 0.644	&0.918 &\textbf{0.032}\\
		&2.5	& 0.758& 0.248	& 0.552 & 0.538	&0.904 & \textbf{0.004}\\
		&3.0	& 0.658 &0.064	& 0.480 &0.468 &0.848& \textbf{0.000}\\
		\hline
	\end{tabular}
	\caption{Robustness of MNIST models. We report upper bounds on the robust accuracy, that is the fraction of points in the test set which are still correctly classified when any perturbation of $l_2$-norm smaller than or equal to $\epsilon$ is allowed in order to achieve a misclassification (a smaller robust accuracy means a stronger attack). The statistics are computed on the first 500 points of the MNIST test set.}
	\label{tab:l2_mnist}
\end{table}
             
\paragraph{MNIST.} For MNIST we use the same architecture as in \cite{MadEtAl2018}, consisting in 2 convolutional layers of 16 and 32 filters, each followed by max-pooling, and 2 dense layers. In particular, the plain and $l_\infty$- trained models are the \textit{natural} and \textit{secret} models of "MNIST Adversarial Examples Challenge"\footnote{\path{https://github.com/MadryLab/mnist_challenge}}, based on
%\begin{center}
%https://github.com/MadryLab/mnist\_challenge
%\end{center} 
\cite{MadEtAl2018}. For $l_2$-\textit{at} we adapted the code
of \cite{MadEtAl2018} to perform adversarial training using the PGD attack wrt the $l_2$-norm with $\epsilon=2$ and 40 iterations. The clean accuracy of the models can be found in Table \ref{tab:l2_mnist} in the row corresponding to $\epsilon=0$. Moreover, we use $M=5$ different starting points for our attack (corresponding to five classes) and $N=500$ as the maximum number of linear regions checked, or equivalently iterations in Algorithm \ref{alg:algorithm-label}, for each starting point. Moreover we set the parameter $\gamma$ in Algorithm \ref{alg:algorithm-label} to $6$.\\
In Table \ref{tab:l2_mnist} we report the robust accuracy, computed on 500 points of the test set, for the three models when the $l_2$-norm of the perturbations is bounded by $\epsilon$. We see that in most of the cases our attack achieves the best performance. In particular, on the $l_\infty$-trained model all the other gradient-based methods suggest that the classifier is highly robust, while our attack shows that this is not the case, as it turns out to be just slightly less vulnerable to adversarial examples than the \textit{plain} model (e.g. at $\epsilon=2.0$ the best of other attacks reduces accuracy only to 53.6\% while our technique brings it down to 3.2\%).\\
Notably, in \cite{SchEtAl19} the same $l_\infty$-\textit{at} model was tested and, taking the pointwise best output among those of 11 attacks of various nature, the authors could decrease robust accuracy no more than 35\% with $\epsilon=1.5$. On the other hand we see that our attack alone, without even testing all the possible 9 target classes, yields an upper bound on robust accuracy for the same $\epsilon$ of 25.8\%, which is almost 10\% less than the current state-of-the-art \cite{BreRauBet18}.

\begin{table}[t]
	\centering
	\begin{tabular}{C{6mm}|C{4mm} | C{6mm} C{6mm} C{6mm} C{6mm} C{6mm} C{6mm}}
		\multicolumn{8}{c}{\textbf{$l_2$ robust accuracy on GTS}}\\[4pt]
		model & $\epsilon$ & PGD-1 & PGD-10k & CW-10k &CW-100k & DF & ours \\
		\hline
		\hline
		\multirow{6}{*}{\textit{plain}} & 0.0 & \multicolumn{6}{c}{0.946}\\
		\cdashline{2-8}
		& 0.1 & 0.746 &0.746 & 0.754 &0.754 & 0.788 &\textbf{0.740} \\
		& 0.2 & 0.568 & 0.562 & 0.566 &0.566 & 0.628 & \textbf{0.550}\\
		& 0.4	& 0.360 & 0.348 & 0.334 &0.334	& 0.408 &\textbf{0.316}\\
		& 0.6	& 0.298 & 0.274 & 0.214 &0.214	& 0.292 & \textbf{0.178}\\
		& 0.8	& 0.268 &0.234 & 0.124 &0.124	& 0.210 &\textbf{0.108}\\
		\hline
		\multirow{6}{*}{$l_2$-\textit{at}} & 0.0 & \multicolumn{6}{c}{0.908}\\
		\cdashline{2-8}
		& 0.1 & \textbf{0.818} & \textbf{0.818} & 0.826 & 0.820 & 0.826 & \textbf{0.818}\\
		&0.2	& 0.708 & \textbf{0.704} & 0.706 &0.710	& 0.728& \textbf{0.704}\\
		&0.4 & 0.488 & 0.472 & 0.496 &0.496 &0.538& \textbf{0.468}\\
		&0.6	& 0.328 & \textbf{0.320} & 0.322 & 0.322 &0.378& \textbf{0.320}\\
		&0.8	& 0.222 & 0.218 & 0.224 &0.222 &0.284 & \textbf{0.212}\\
		\hline
		\multirow{6}{*}{$l_\infty$-\textit{at}}& 0.0 & \multicolumn{6}{c}{0.904}\\
		\cdashline{2-8}
		& 0.25 & 0.690 & \textbf{0.686} & 0.692 &0.692 & 0.718 & \textbf{0.686}\\
		& 0.5& 0.460 & 0.446 & 0.468 &0.466 & 0.500 &\textbf{0.444} \\
		& 0.75&	0.302 &0.288 & 0.300 & 0.300 & 0.338 & \textbf{0.280}\\
		& 1.0	& 0.212& 0.200 &  0.214 &0.214& 0.246 & \textbf{0.194}\\
		& 1.25	& 0.164 & 0.130 & 0.116 &0.114 & 0.172 & \textbf{0.072}\\
		\hline
	\end{tabular}
	%\label{tab:l2_gts}
	\caption{Robustness of GTS models. We report upper bounds on the robust accuracy, that is the fraction of points in the test set which are still correctly classified when any perturbation of $l_2$-norm smaller than or equal to $\epsilon$ is allowed in order to achieve a misclassification (a smaller robust accuracy means a stronger attack). The statistics are computed on the first 500 points of the GTS test set.}
	\label{tab:l2_gts}
\end{table}

\paragraph{GTS.} In this case the models are CNNs with 2 convolutional layers (16 and 32 feature maps) with stride 2, which replaces max-pooling for downsizing, and 2 dense layers. Adversarial training is based on 40 iterations of PGD attack, with $\epsilon=0.5$ for $l_2$-\textit{at} and $\epsilon=\nicefrac{4}{255}$ for $l_\infty$-\textit{at}. Since GTS has 43 classes, we run our algorithm with $M=15$ starting points, 500 linear regions each and $\gamma=9$.\\
Table \ref{tab:l2_gts} shows how the upper bounds on robust accuracy, computed on the first 500 images of the test set, obtained through our technique are always smaller than those by the competitors, apart from 4 cases out of 15 where the PGD results can only match ours. We notice that, although in some cases the difference is not extremely large, in 3 of 15 settings our attack reduces the robust accuracy at least by 2\% compared to the best result of the other methods, with a maximum of 4.2\% for $\epsilon=1.25$ for the $l_\infty$-trained model (robust accuracy of 11.4\% by CW-100k vs 7.2\% for our attack).
%{\color{red}[removed] It is interesting to notice that, similarly to MNIST, the other attacks seem to have problems to perform well on the models adversarially trained wrt $l_\infty$-norm.}

\begin{table}[t]
	\centering
	\begin{tabular}{C{6mm}|C{4mm} | C{6mm} C{6mm} C{6mm} C{6mm} C{6mm} C{6mm}}
		\multicolumn{8}{c}{\textbf{$l_2$ robust accuracy on CIFAR-10}}\\[4pt]
		model & $\epsilon$ & PGD-1 & PGD-10k & CW-10k & CW-100k & DF & ours \\
		\hline
		\hline
		\multirow{6}{*}{\textit{plain}}& 0.0 & \multicolumn{6}{c}{0.892}\\
		\cdashline{2-8}
		&0.1 & 0.686 & \textbf{0.676} & 0.694 & 0.694 & 0.722& 0.690 \\
		& 0.15 & 0.546 & \textbf{0.536}  & 0.554 & 0.552 & 0.626 & 0.550\\
		& 0.2	& 0.440 & \textbf{0.422} & 0.434 &0.432 & 0.512 &0.434 \\
		& 0.3	& 0.256& 0.234 &\textbf{0.216} & \textbf{0.216} & 0.338 & 0.220\\
		& 0.4	& 0.182 &0.146 &0.094 & \textbf{0.092} & 0.208 & 0.098\\
		\hline
		\multirow{6}{*}{$l_2$-\textit{at}}& 0.0 & \multicolumn{6}{c}{0.812}\\
		\cdashline{2-8}
		& 0.25 & 0.658 & \textbf{0.656}& 0.660 & 0.660 & 0.670& \textbf{0.656} \\
		& 0.5& 0.496 & 0.488	& 0.482 & 0.482 & 0.538&\textbf{0.478}\\
		& 0.75& 0.382 & 0.362& \textbf{0.324} & \textbf{0.324} & 0.422 & \textbf{0.324}\\
		& 1.0	& 0.358 &0.322& 0.212 & \textbf{0.204} & 0.300& 0.216\\
		& 1.25	& 0.336 & 0.302& \textbf{0.114} &\textbf{0.114} & 0.224 & 0.124\\
		\hline
		\multirow{6}{*}{$l_\infty$-\textit{at}}& 0.0 & \multicolumn{6}{c}{0.794}\\
		\cdashline{2-8}
		& 0.25 & 0.646 & \textbf{0.644} & 0.646 & 0.646& 0.670 & \textbf{0.644}\\
		& 0.5& 0.488 &\textbf{0.484}& \textbf{0.484} &\textbf{0.484} & 0.530 & 0.488\\
		& 0.75& 0.390&0.368 & \textbf{0.332} &\textbf{0.332} &0.414& 0.334\\
		& 1.0& 0.352& 0.332 & \textbf{0.226} & 0.228 & 0.326& 0.228\\
		& 1.25& 0.348 &0.324  & \textbf{0.120} & \textbf{0.120} & 0.242& 0.130\\
		\hline
	\end{tabular}
	\caption{Robustness of CIFAR-10 models. We report upper bounds on the robust accuracy, that is the fraction of points in the test set which are still correctly classified when any perturbation of $l_2$-norm smaller than or equal to $\epsilon$ is allowed in order to achieve a misclassification (a smaller robust accuracy means a stronger attack). The statistics are computed on the first 500 points of the CIFAR-10 test set.}
	\label{tab:l2_cifar10}
\end{table}

\paragraph{CIFAR-10.} Since CIFAR-10 represents a more difficult classification task, we use for it a deeper and wider architecture, made of 8 convolutional layers (with number of filters increasing from 96 to 384) and 2 dense layers, which contains overall more than 375000 units. We perform adversarial training again with the PGD attack, with 10 iterations, $\epsilon=\nicefrac{80}{255}$ and $\epsilon=\nicefrac{4}{255}$ for $l_2$- and $l_\infty$-robust training respectively. We here run our attack with 400 iterations and 3 starting points, fixing $\gamma=9$.\\
The statistics over the first 500 points of the test set are summarized in Table \ref{tab:l2_cifar10}. Although with this dataset we see that the best performances are achieved by different methods in many situations, we can nevertheless notice that our attack clearly outperforms PGD and DF and is at most 1.4\% off from the best robust accuracy. CW attack performs here very well but it still has a slightly worse performance in the \textit{worst case} setting, as we can see in Table \ref{tab:perf_in_dataset}. \\
Moreover, these CIFAR-10 networks are less robust than those trained on MNIST and GTS, so that the task of crafting small adversarial examples is easier than previously. This implies that even weak attackers can succeed in finding good, maybe almost optimal, adversarial perturbations.

\begin{table}[t]
	\centering
	\begin{tabular}{C{9mm}|C{4mm} | C{6mm} C{6mm} C{6mm} C{6mm} C{6mm} C{6mm}}
		\multicolumn{8}{c}{\textbf{$l_2$ robust accuracy on MNIST}}\\[4pt]
		model &$\epsilon$ & PGD-1k & PGD-10k & CW-10k & CW-100k & DF & ours\\
		\hline
		\hline
		
		\multirow{6}{*}{\shortstack{$l_\infty$-\\MMR-\textit{at}}} & 0.0 & \multicolumn{6}{c}{0.988}\\
		\cdashline{2-8}
		& 1.0 & 0.828 & 0.816 & 0.854 & 0.854 & 0.868 & \textbf{0.704}\\
		&1.5	& 0.488 & 0.428	& 0.642 & 0.642 & 0.682 & \textbf{0.250}\\
		&2.0	& 0.310 & 0.270	& 0.414 &0.412 &0.642  &\textbf{0.048}\\
		&2.5	& 0.222& 0.180	& 0.196 &0.194 &0.238 & \textbf{0.004}\\
		&3.0	& 0.136 &0.116	& 0.074 &0.070 &0.084 & \textbf{0.000}\\
		\hline
		
		\multirow{6}{*}{\shortstack{$l_\infty$-KW}} & 0.0 & \multicolumn{6}{c}{0.982}\\
		\cdashline{2-8}
		&1.0 & 0.924 & 0.910 & 0.924 & 0.924 & 0.926 & \textbf{0.854}\\ &1.5 & 0.674 & 0.600 & 0.834 & 0.834 & 0.898 & \textbf{0.478}\\ &2.0 & 0.226 & 0.176 & 0.664 & 0.662 & 0.844 & \textbf{0.148}\\ &2.5 & 0.030 & 0.020 & 0.454 & 0.454 & 0.784 & \textbf{0.018}\\ &3.0 & 0.002 & \textbf{0.000} & 0.264 & 0.264 & 0.644 & 0.002\\
		\hline
		
		\multirow{6}{*}{\shortstack{$l_2$-\\MMR-\textit{at}}} & 0.0 & \multicolumn{6}{c}{0.986}\\
		\cdashline{2-8}
		& 1.0 & 0.848 & 0.848 & 0.850 & 0.850 & 0.868  &\textbf{0.842}\\
		& 1.5 & 0.608 & 0.606 & 0.622 & 0.622 & 0.682  &\textbf{0.576}\\
		& 2.0 & 0.286 & 0.270 &0.312 & 0.312 &0.462  & \textbf{0.238} \\
		& 2.5 &0.050 & 0.048 &0.090 & 0.090 & 0.238  &\textbf{0.044}\\
		& 3.0 & 0.016 & 0.012 & 0.032 & 0.030 & 0.084 & \textbf{0.010} \\
		%& time & & 32838s & 5128s & & 419s & 1917s \\
		\hline
		
		\multirow{6}{*}{$l_2$-KW} & 0.0 & \multicolumn{6}{c}{0.988}\\
		\cdashline{2-8}
		& 1.0 & 0.916 & 0.916 &0.914 & 0.914  & 0.928& \textbf{0.912} \\
		& 1.5 & 0.722& 0.716&0.740 & 0.740  &0.826 & \textbf{0.692}\\
		& 2.0 & 0.392& 0.366& 0.438& 0.438  & 0.690& \textbf{0.298} \\
		& 2.5 &0.214 & 0.202& 0.166& 0.166  &0.478 & \textbf{0.078}\\
		& 3.0 &0.172 & 0.152& 0.046& 0.046  &0.292 & \textbf{0.012}\\
		%& time & & 33156s& 6037s & & 426s & 1914s \\
		\hline
	\end{tabular}
\caption{Provably robust MNIST models. We report upper bounds on the robust accuracy, that is the fraction of points in the test set which are still correctly classified when any perturbation of $l_2$-norm smaller than or equal to $\epsilon$ is allowed in order to achieve a misclassification (a smaller robust accuracy means a stronger attack). The statistics are computed on the first 500 points of the MNIST test set.}
\label{tab:l2_mnist_rob}
\end{table}

\begin{table}[t]
	\centering
	\begin{tabular}{C{9mm}|C{4mm} | C{6mm} C{6mm} C{6mm} C{6mm} C{6mm} C{6mm}}
		\multicolumn{8}{c}{\textbf{$l_2$ robust accuracy on CIFAR-10}}\\[4pt]
		model & $\epsilon$ & PGD-1k & PGD-10k & CW-10k & CW-100k & DF & ours\\
		\hline
		\hline
		
		\multirow{6}{*}{\shortstack{$l_\infty$-\\MMR-\textit{at}}} & 0.0 & \multicolumn{6}{c}{0.638}\\
		\cdashline{2-8}
		& 0.25 & 0.504 & 0.504 & 0.490 & 0.490 & 0.498 &\textbf{0.484}\\
		& 0.5 & 0.332 & 0.330 & 0.340 &0.340 & 0.348 & \textbf{0.314}\\
		& 0.75 & 0.180 &0.174 & 0.176 & 0.174 & 0.210 &\textbf{0.154}  \\
		& 1.0 & 0.066 &0.064 & 0.070& 0.070 & 0.096 & \textbf{0.056}\\
		& 1.25 & 0.036 & 0.034&  0.032& 0.032 & 0.050 &\textbf{0.028} \\
		%& time & 16303s & & 1761s & &  & 12544s\\
		\hline
		
		\multirow{6}{*}{\shortstack{$l_\infty$-KW}} & 0.0 & \multicolumn{6}{c}{0.532}\\
		\cdashline{2-8}
		&0.25 & 0.390 & 0.390 & 0.376 & 0.376 & 0.374 & \textbf{0.364}\\ &0.5 & 0.238 & 0.236 & 0.218 & 0.218 & 0.236 & \textbf{0.216}\\ &0.75 & 0.132 & 0.130 & 0.128 & 0.128 & 0.146 & \textbf{0.104}\\ &1.0 & 0.060 & 0.060 & 0.064 & 0.064 & 0.082 & \textbf{0.036}\\ &1.25 & 0.018 & 0.018 & 0.032 & 0.032 & 0.036 & \textbf{0.014}\\
		\hline
		
		\multirow{6}{*}{\shortstack{$l_2$-\\MMR-\textit{at}}} & 0.0 & \multicolumn{6}{c}{0.618}\\
		\cdashline{2-8}
		& 0.25 & 0.418 & 0.418 & 0.404 & 0.404 & 0.412 & \textbf{0.398}\\
		& 0.5 & 0.270 &0.266 & 0.264 & 0.262&0.284 & \textbf{0.252}\\
		& 0.75 & 0.146 &0.144 & 0.146 & 0.146& 0.174 &\textbf{0.128} \\
		& 1.0 & 0.076 &0.076 & 0.094 &0.094 & 0.104  & \textbf{0.064}\\
		& 1.25 & 0.032 & 0.032& 0.050 & 0.050&  0.054 & \textbf{0.024} \\
		%& time & 18518s & & 2012s & &  & 4051s \\
		\hline
		
		\multirow{6}{*}{$l_2$-KW} & 0.0 & \multicolumn{6}{c}{0.614}\\
		\cdashline{2-8}
		& 0.25 & 0.492 & 0.492 & \textbf{0.478} & \textbf{0.478} & 0.480 & \textbf{0.478} \\
		& 0.5 & 0.384 &0.384 & 0.374& 0.374 & 0.376 & \textbf{0.360} \\
		& 0.75 & 0.266 &0.266 &0.262 & 0.262 & 0.284  & \textbf{0.246} \\
		& 1.0 &0.172 & 0.172 & 0.176 & 0.176 &0.190  & \textbf{0.152}\\
		& 1.25 & 0.094 &0.092 & 0.108 & 0.108 &0.122 & \textbf{0.082} \\
		%& time & 17934s & & 2083s  & &  & 12835s \\
		\hline
	\end{tabular}
\caption{Provably robust CIFAR-10 models. We report upper bounds on the robust accuracy, that is the fraction of points in the test set which are still correctly classified when any perturbation of $l_2$-norm smaller than or equal to $\epsilon$ is allowed in order to achieve a misclassification (a smaller robust accuracy means a stronger attack). The statistics are computed on the first 500 points of the CIFAR-10 test set.}
\label{tab:l2_cifar10_rob}
\end{table}

\subsection{Testing provably robust models}
In this section we test classifiers trained to be provably robust, that is it is possible to compute for a large fraction of the test points if there exists or not an adversarial perturbation with norm smaller than a fixed threshold. This means that non-trivial \textit{lower} bounds on the robust accuracy are provided. For what concerns \textit{upper} bounds, we have mostly to rely, especially for the $l_2$ case, on the adversarial examples provided by the attacks. Then, using powerful attacks allows also to correctly assess the tightness of the lower bounds or equivalently the effectiveness of the verification methods.\\
We consider the models presented in \cite{CroEtAl2018}, that is CNNs with 2 convolutional layers of 16 and 32 filters and a hidden fully-connected layer of 100 units. These are trained with the techniques of either \cite{CroEtAl2018} (called MMR) or \cite{WonKol2018,WonEtAl18} (KW) to be robust wrt the $l_2$-norm at $\epsilon_\textrm{train}=0.3$ for MNIST and $\epsilon_\textrm{train}=0.1$ for CIFAR-10, wrt the $l_\infty$-norm at $\epsilon_{\textrm{train}}=0.1$ for MNIST and $\epsilon_\textrm{train}=\nicefrac{2}{255}$. We decide to test the $l_2$ robustness of all the models with thresholds $\epsilon$ larger than those used for $l_2$ robust training since at those levels the uncertainty on robust accuracy is limited as tight bounds on it are available (see \cite{CroEtAl2018}).
We run our attack for 500 regions and 5 starting points. In Table \ref{tab:l2_mnist_rob} (MNIST) and Table \ref{tab:l2_cifar10_rob} (CIFAR-10) we report similarly to the previous section the upper bounds on the robust accuracy, computed with 500 test points, provided the different attacks (we here use PGD-1k with 1000 restarts instead of the weaker version with a single restart).\\ For both datasets we see that our attack outperforms, often significantly, the competitors, with the only exception being the largest value of $\epsilon$ on the model trained with KW technique wrt $l_\infty$-norm on MNIST. Moreover, note that similar to  Table \ref{tab:l2_mnist}, the largest differences (over 22\% between the upper bounds on robust accuracies of PGD-100k and our attack) are reached for the classifier trained on MNIST with adversarial training from \cite{MadEtAl2018} wrt $l_\infty$.

\begin{table}[t!]
	\centering
	\begin{tabular}{C{9mm}|C{8mm} | C{11mm} C{11mm} C{11mm}}
		\multicolumn{5}{c}{\textbf{$l_2$ robust accuracy of large networks on CIFAR-10}}\\[4pt]
		model & $\epsilon$ & PGD-1k & DF & ours\\
		\hline
		\hline
		
		\multirow{6}{*}{\textit{plain}} & 0.0 & \multicolumn{3}{c}{0.96}\\
		\cdashline{2-5}
		& 0.05 & \textbf{0.78}  & 0.83 & \textbf{0.78}\\ &0.075 & 0.61 & 0.75 & \textbf{0.60}\\ &0.1 & \textbf{0.43} & 0.63 & 0.44\\ &0.15 & \textbf{0.18}  & 0.42 & \textbf{0.18}\\ &0.2 & 0.08  & 0.26 & \textbf{0.07} \\
		\hline
		
		\multirow{6}{*}{$l_\infty$-\textit{at}} & 0.0 & \multicolumn{3}{c}{0.85}\\
		\cdashline{2-5}
		& 0.25 & 0.72 & 0.75 & \textbf{0.71}\\ &0.5 & \textbf{0.53} & 0.62 & 0.54\\ &0.75 & \textbf{0.36} & 0.51 & 0.37\\ &1.0 & 0.23 & 0.44 & \textbf{0.22}\\ &1.25 & 0.15  & 0.37 & \textbf{0.12} \\
		\hline
		
	\end{tabular}
	\caption{Large models. We report here the robust accuracy, that is an upper bound on the fraction of points in the test set which are correctly classified when any perturbation of $l_2$-norm smaller than or equal to $\epsilon$ is allowed (a smaller robust accuracy means a stronger attack). The statistics are computed on the first 100 points of the CIFAR-10 test set.}
	\label{tab:l2_madrymodel}
\end{table}

\begin{figure*}[t]
	%\centering\includegraphics[width=0.65\columnwidth]{pl_psampl_1_2}
	%\includegraphics[width=0.65\columnwidth]{pl_psampl_2_2}
	%\includegraphics[width=0.65\columnwidth]{pl_psampl_3_2}
	\centering\includegraphics[width=0.65\columnwidth, clip, trim=20mm 0mm 16mm 1mm]{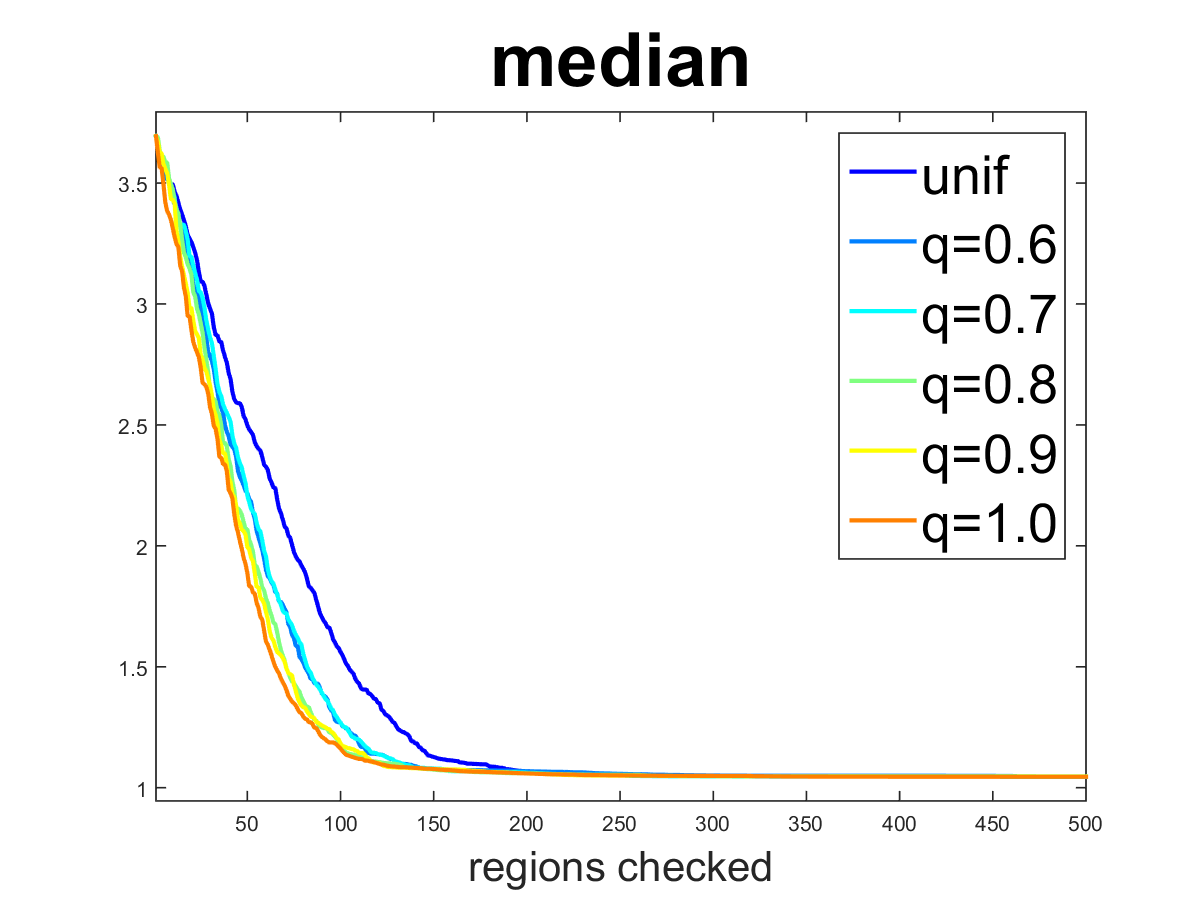}
	\includegraphics[width=0.65\columnwidth, clip, trim=20mm 0mm 16mm 1mm]{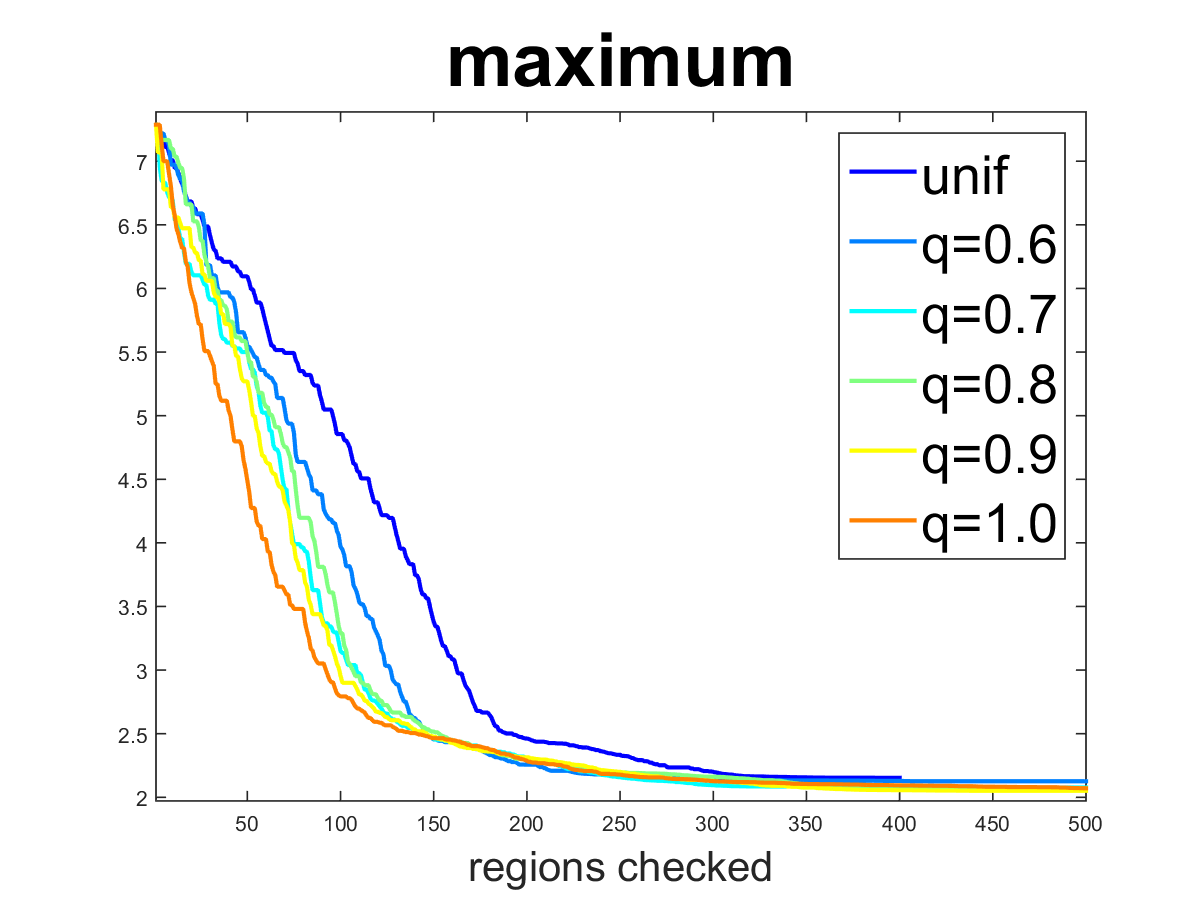}
	\includegraphics[width=0.65\columnwidth, clip, trim=20mm 0mm 16mm 1mm]{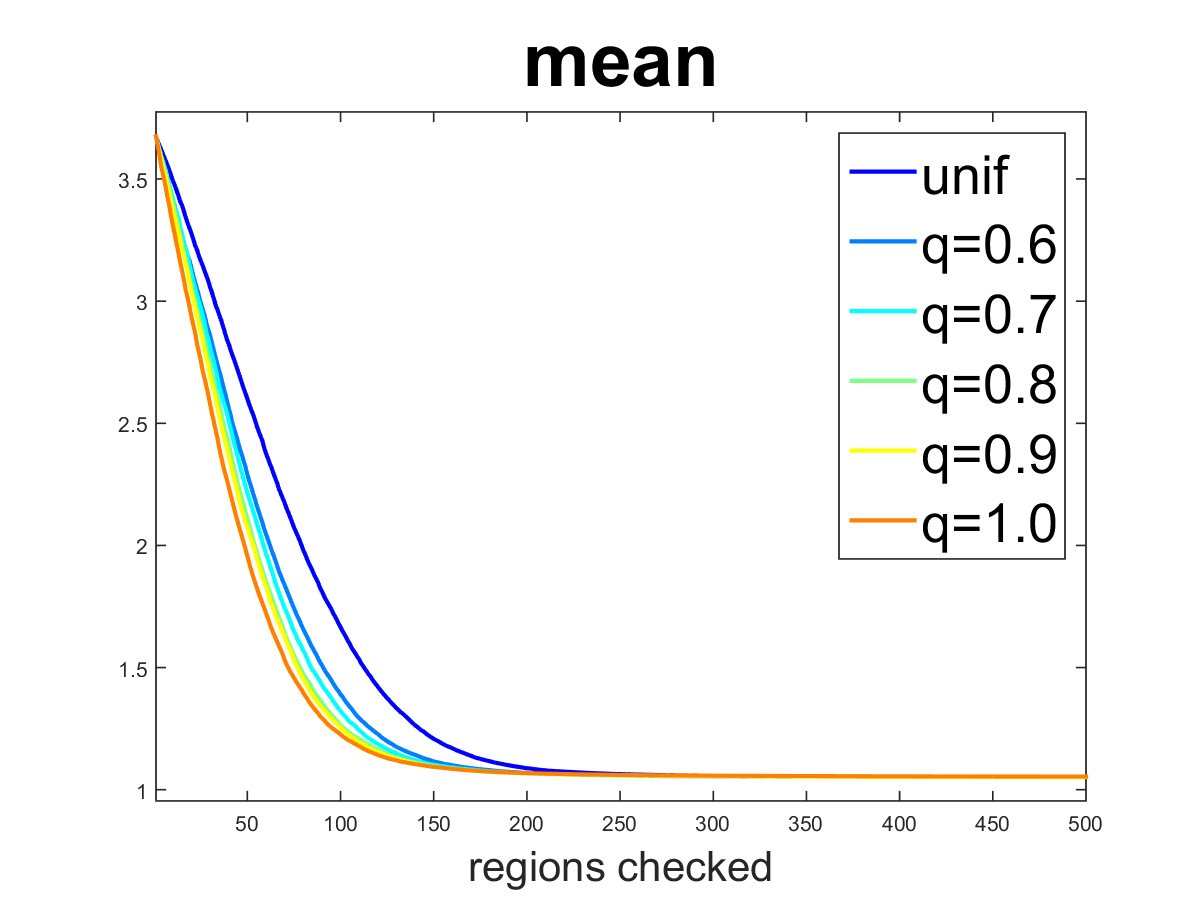}
	\caption{Progression of our attack for different sampling schemes on MNIST. We show median (left), maximum (center) and mean (right) of the norms of the adversarial perturbations found by our attack as a function of the explored linear regions. We repeat the experiments for different values of $q$ (see Equation \eqref{eq:psampl}), represented in different colors, and with the uniform sampling scheme ($q=0.5$) from \cite{CroHei18} as a comparison.}\label{fig:psampl}
\end{figure*}

\begin{figure*}[t]
	\centering\includegraphics[width=0.65\columnwidth, clip, trim=20mm 0mm 16mm 1mm]{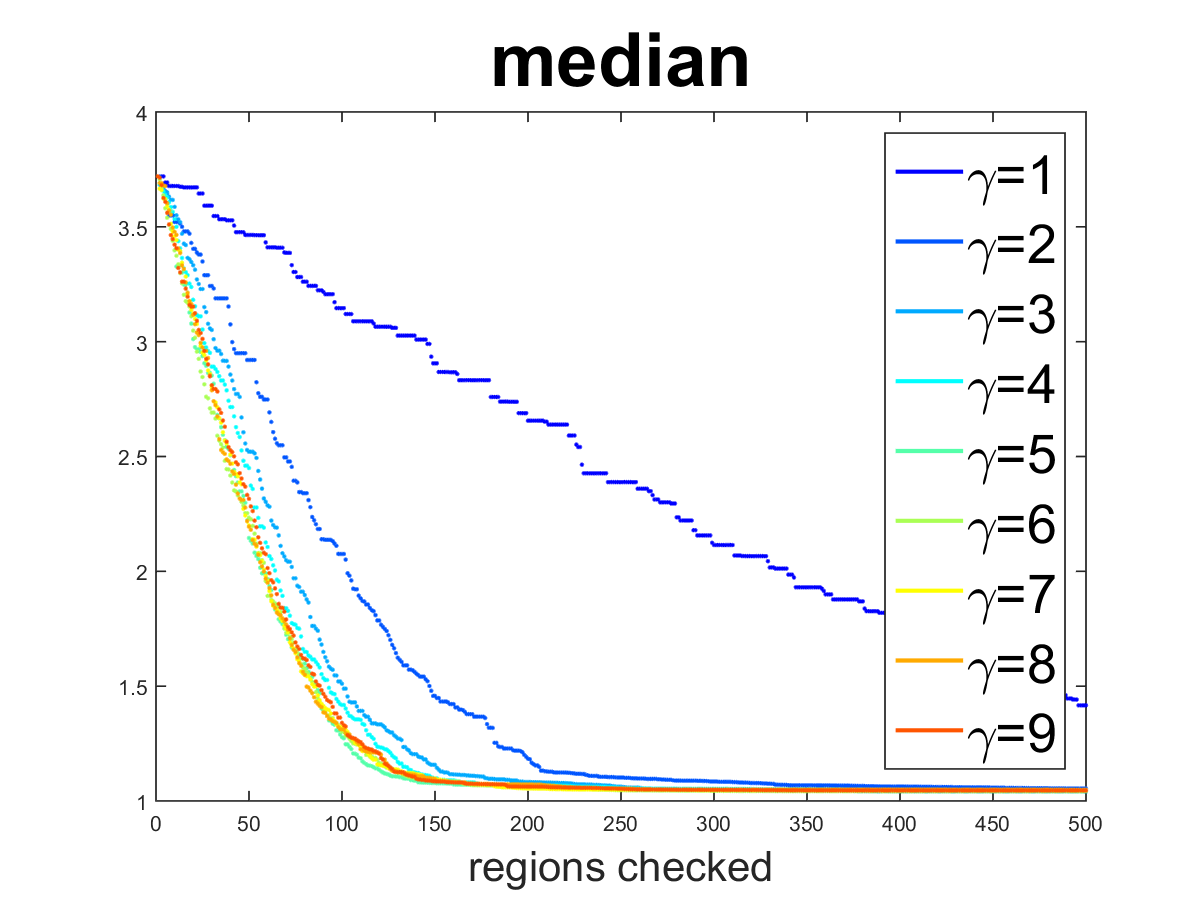}
	\includegraphics[width=0.65\columnwidth, clip, trim=20mm 0mm 16mm 1mm]{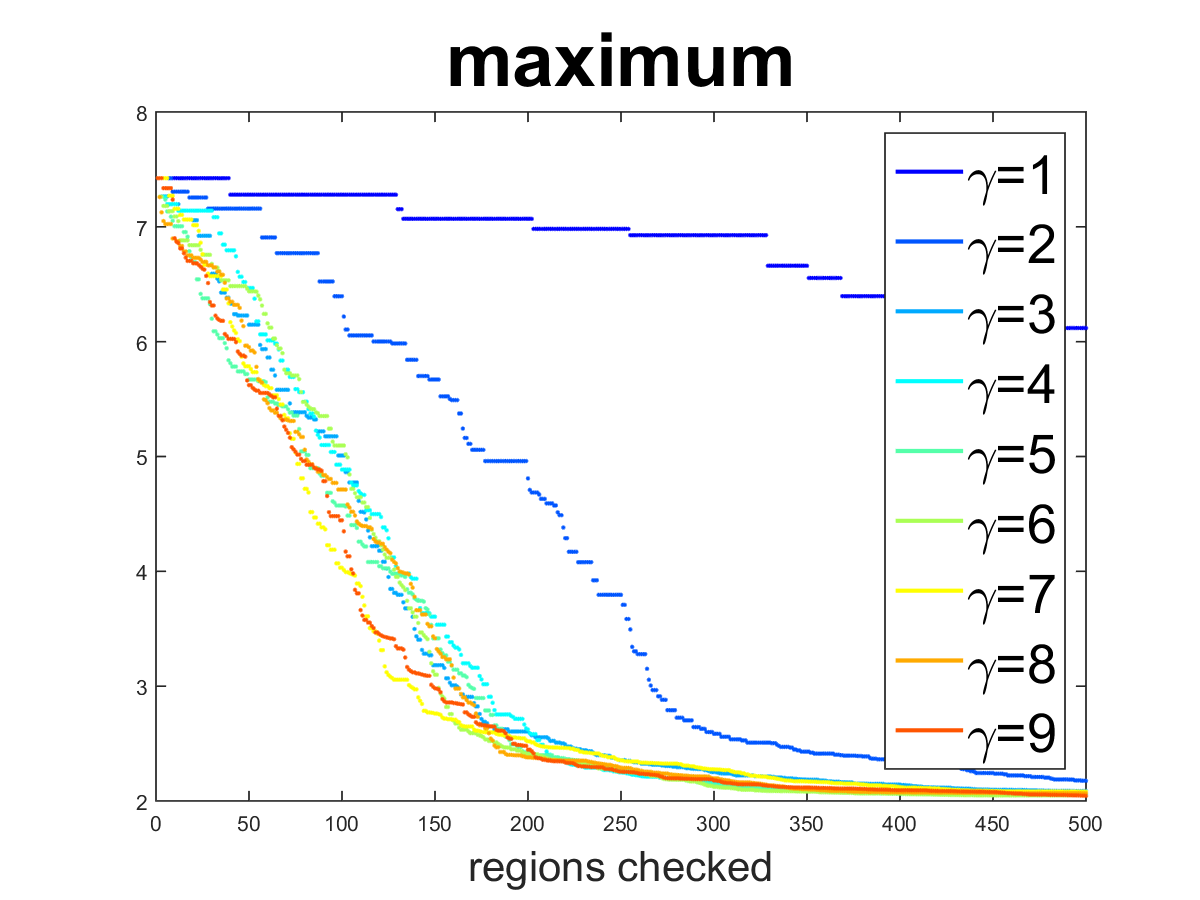}
	\includegraphics[width=0.65\columnwidth, clip, trim=20mm 0mm 16mm 1mm]{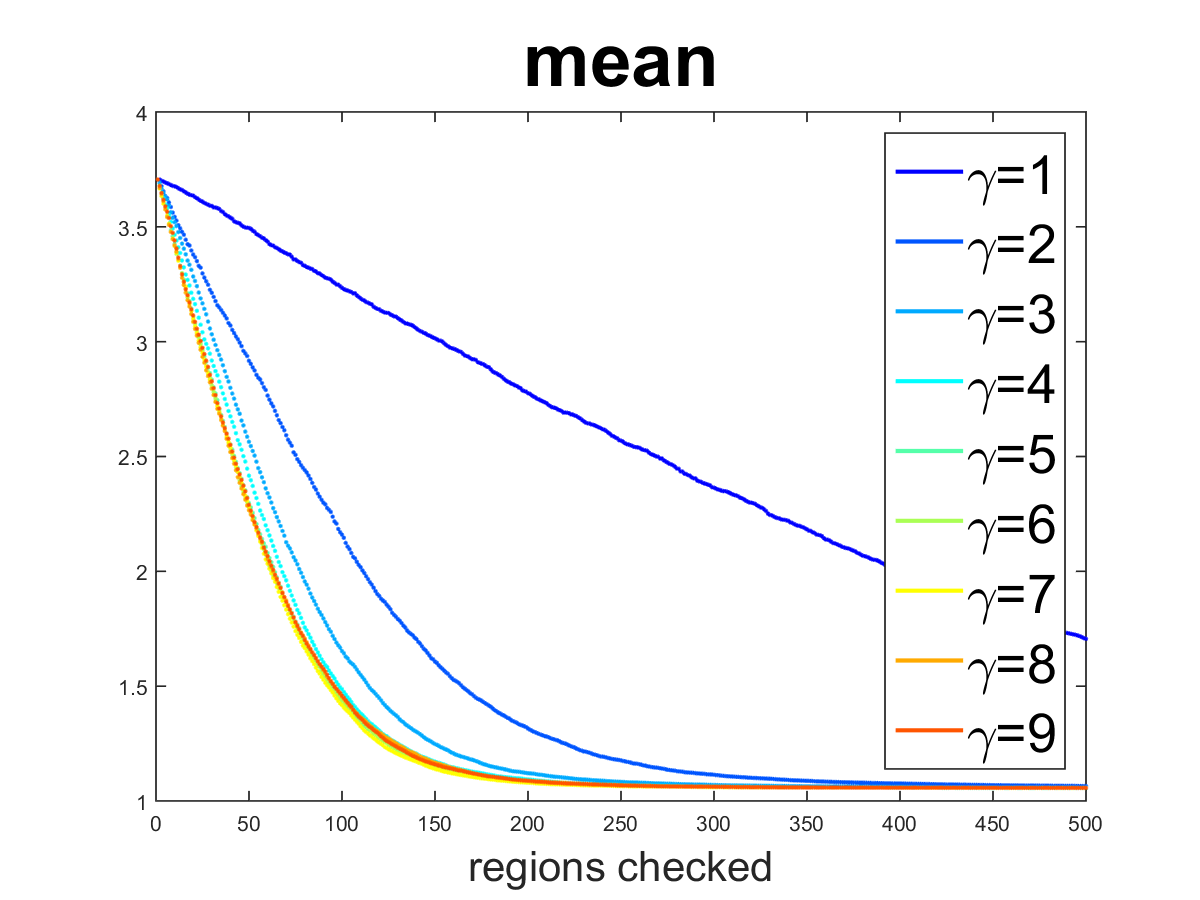}
	\caption{Progression of our attack for different values of the parameter $\gamma$ on MNIST. We show median (left), maximum (center) and mean (right) of the norms of the adversarial perturbations found by our attack as a function of the explored linear regions. We repeat the experiments for $\gamma=1, \ldots, 9$ (see Equation \eqref{eq:sampl_2}) represented in different colors.}\label{fig:stats_gammas}
\end{figure*}

\begin{figure*}[t]
	\centering\includegraphics[width=0.65\columnwidth, clip, trim=20mm 0mm 16mm 1mm]{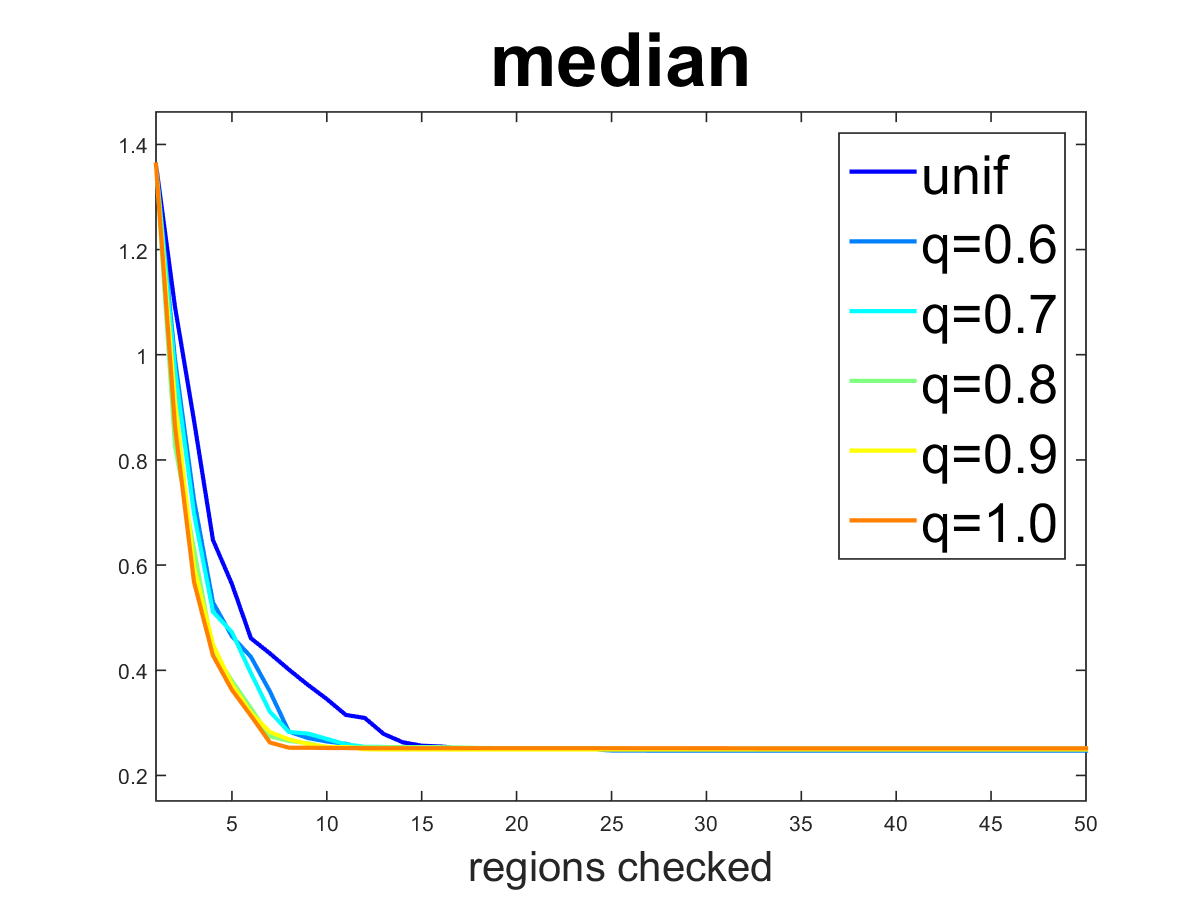}
	\includegraphics[width=0.65\columnwidth, clip, trim=20mm 0mm 16mm 1mm]{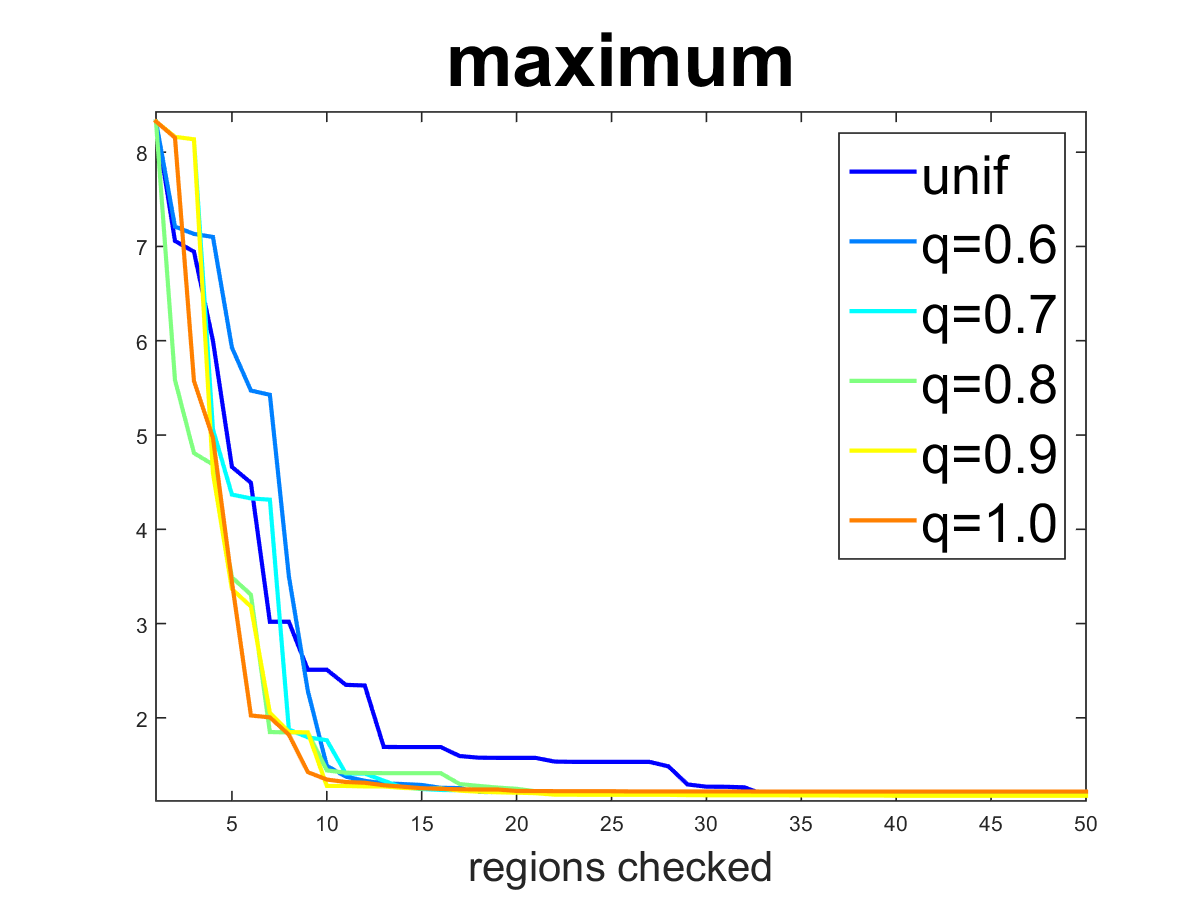}
	\includegraphics[width=0.65\columnwidth, clip, trim=20mm 0mm 16mm 1mm]{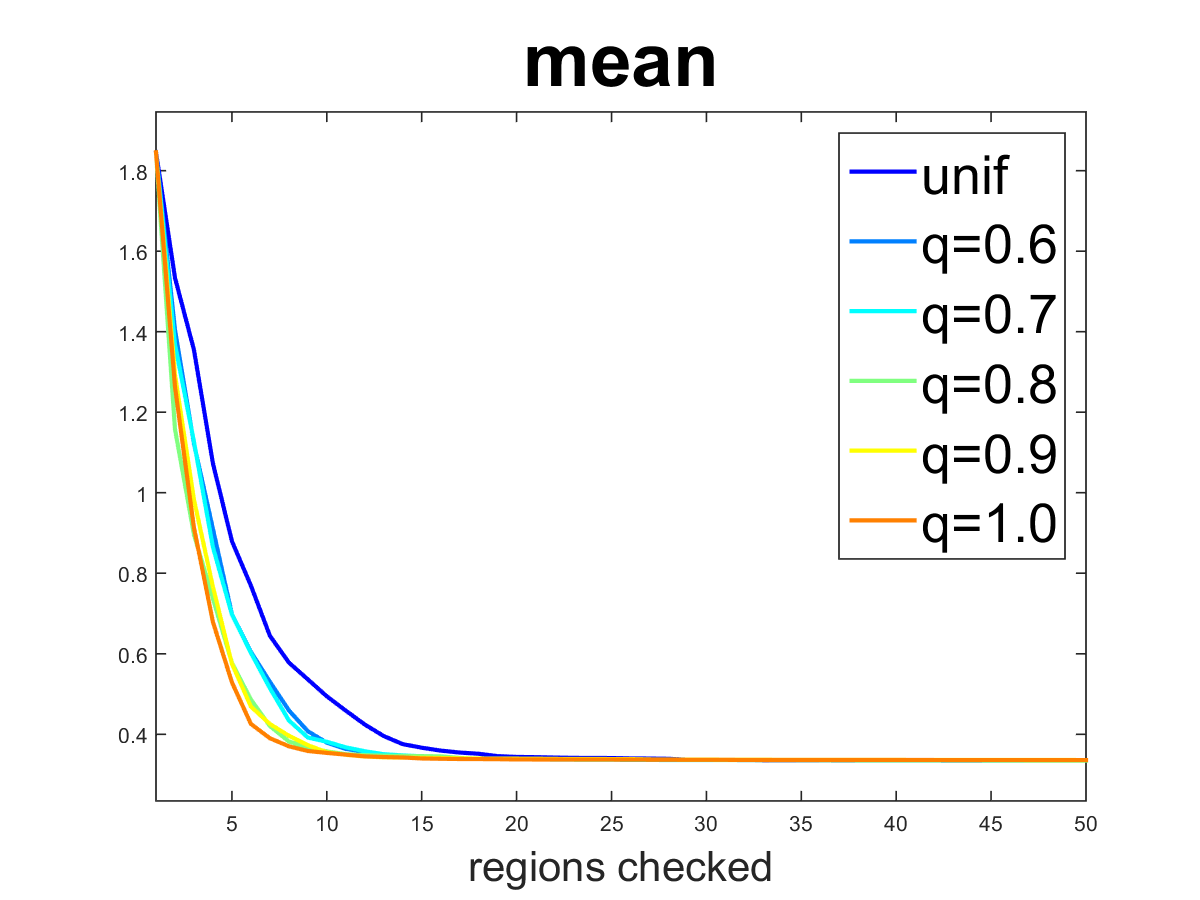}
	\caption{Progression of our attack for different sampling schemes on GTS. We show median (left), maximum (center) and mean (right) of the norms of the adversarial perturbations found by our attack as a function of the explored linear regions. We repeat the experiments for different values of $q$ (see Equation \eqref{eq:psampl}), represented in different colors, and with the uniform sampling scheme ($q=0.5$) from \cite{CroHei18} as a comparison.}\label{fig:psampl_gts}
\end{figure*}

\begin{figure*}[t]
	\centering\includegraphics[width=0.65\columnwidth, clip, trim=20mm 0mm 16mm 1mm]{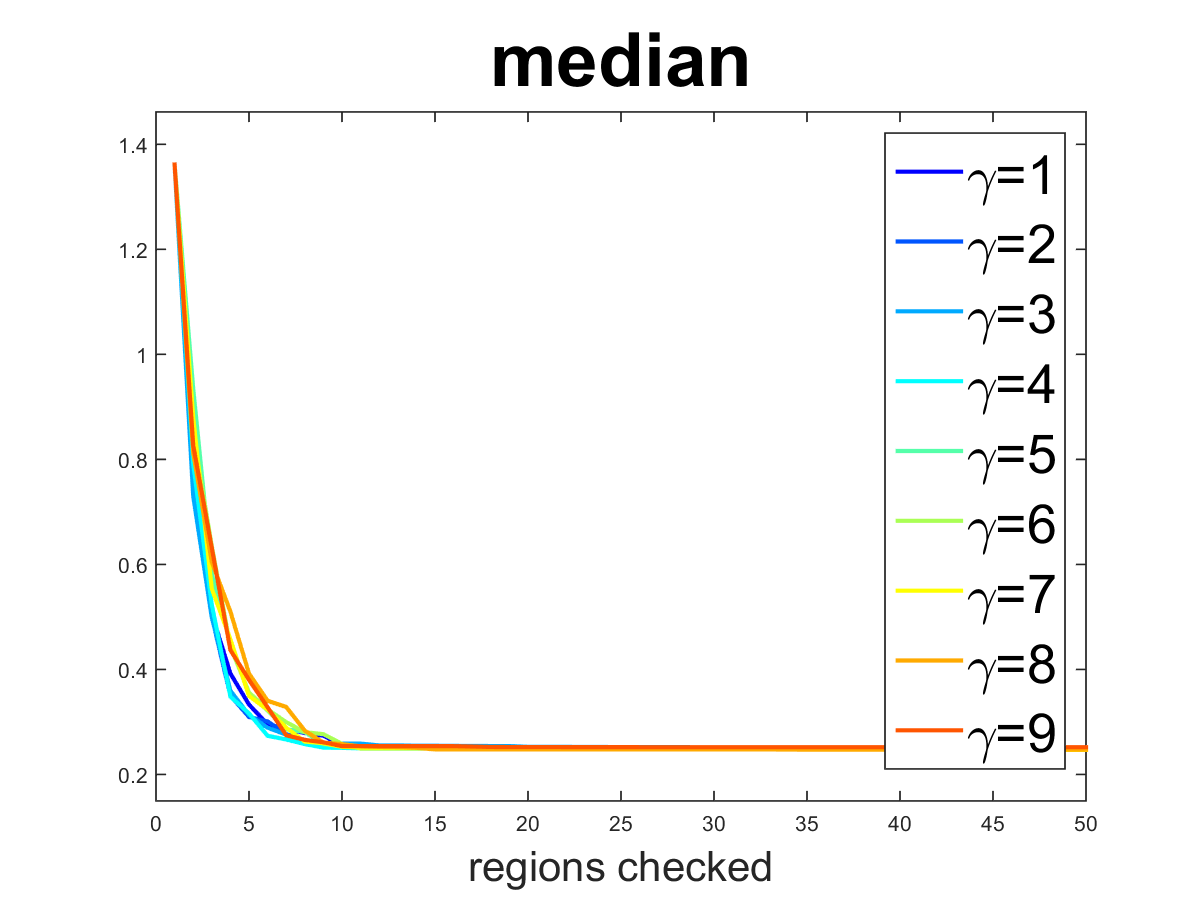}
	\includegraphics[width=0.65\columnwidth, clip, trim=20mm 0mm 16mm 1mm]{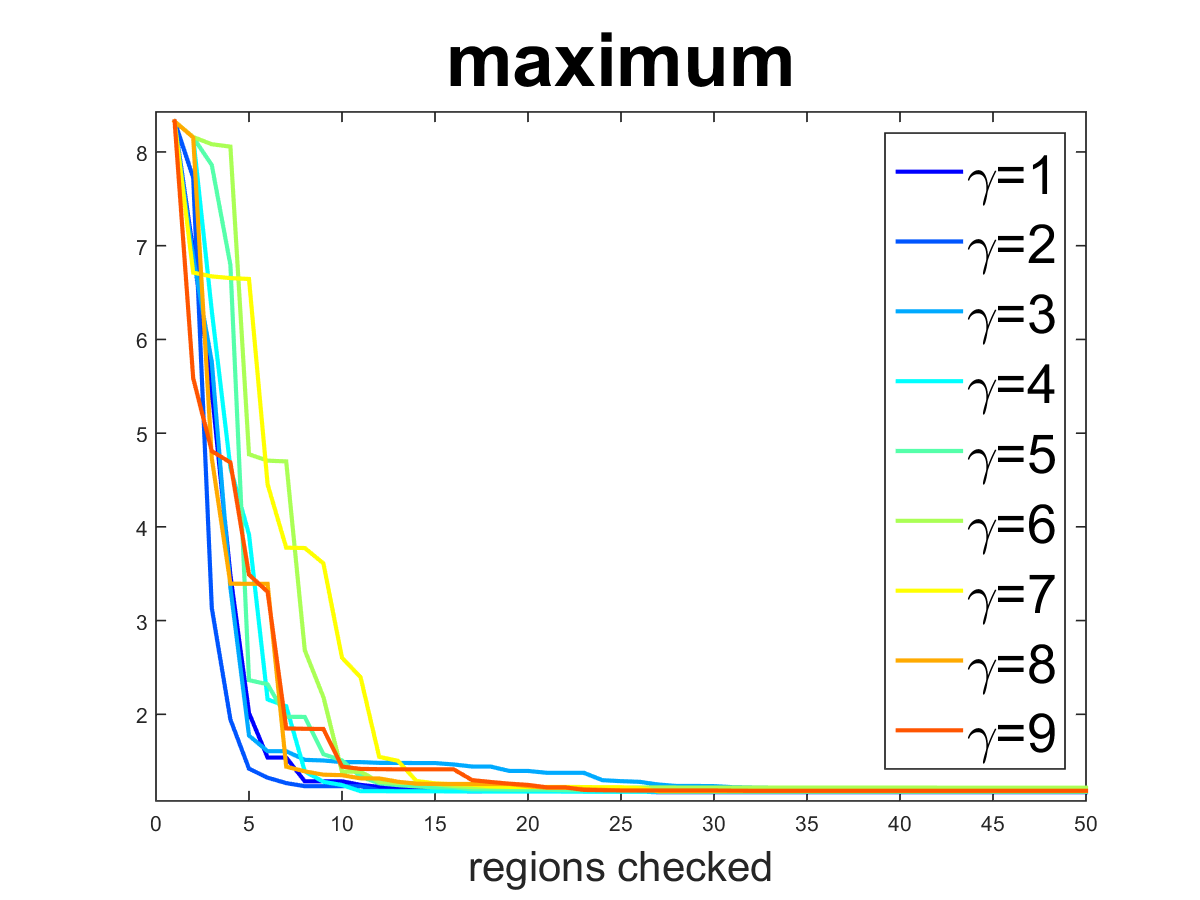}
	\includegraphics[width=0.65\columnwidth, clip, trim=20mm 0mm 16mm 1mm]{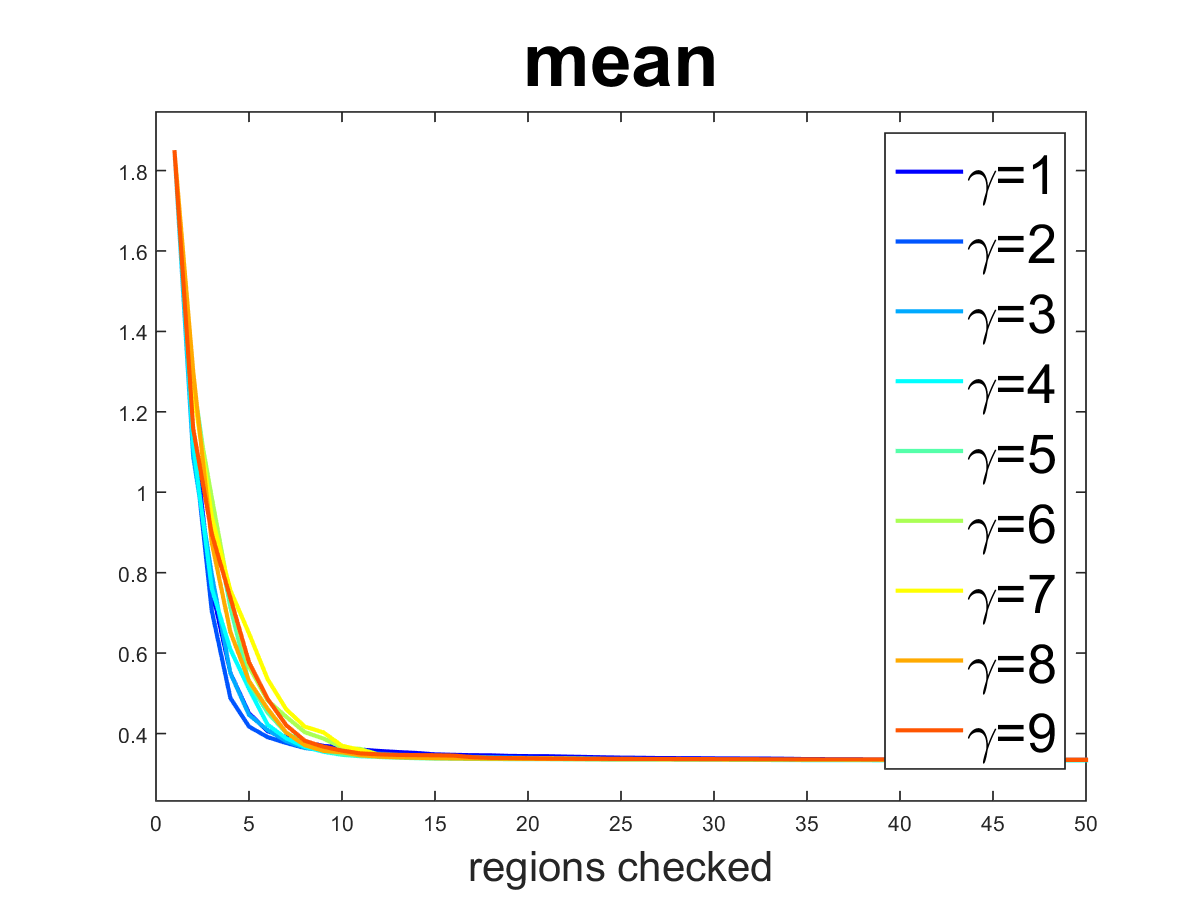}
	\caption{Progression of our attack for different values of the parameter $\gamma$ on GTS. We show median (left), maximum (center) and mean (right) of the norms of the adversarial perturbations found by our attack as a function of the explored linear regions. We repeat the experiments for $\gamma=1, \ldots, 9$ (see Equation \eqref{eq:sampl_2}) represented in different colors.}\label{fig:stats_gammas_gts}
\end{figure*}

\subsection{Attacking large models}\label{sec:madry_model}
In order to show the scalability of our approach to large models, we here attack the networks from "CIFAR-10 Adversarial Examples Challenge"\footnote{\path{https://github.com/MadryLab/cifar10_challenge} } trained on CIFAR-10 with either plain or $l_\infty$-adversarial training \cite{MadEtAl2018} (called \textit{naturally trained} and \textit{secret} in the original challenge). The architecture used is a residual convolutional network consisting of a convolutional layer, five residual blocks and a fully-connected layer, derived from the "w32-10 wide" variant of the TensorFlow model repository, with 2.883.593 units. In order to apply our algorithm we had to replace the per image normalization, which is not an affine operation on the input, with the following step: for each input image, we subtract the mean of its entries and divide it by a constant (0.21, which is an approximation of the average standard deviation across the images of the training set). Note that this small variation does not affect the performance of the classifier while allows the network to result in a piecewise affine function.\\
In Table \ref{tab:l2_madrymodel} we report the robust accuracy, on the first 100 test points, given by the three methods (in this case we use PGD with 1000 but not 10000 restarts as it would be computationally too expensive). We omit CW since with the default parameters it fails to provide meaningful results. For our method we use 5 starting points. While DeepFool is always worse than the others, PGD and our attack perform similarly, although the largest gap (3\%, achieved at $\epsilon=1.25$ for the $l_\infty$-\textit{at} model) is in favour of our method.

\subsection{Runtime comparison}
We analyze the runtime the different attacks take to return results on 500 test points on the \textit{plain} model on CIFAR-10 of Section \ref{sec:main_exp} using a single GPU. Note that CW, DeepFool and our method aim at finding the minimal adversarial perturbation within a limited budget of iterations while PGD takes as input a thresholds $\epsilon$ and looks for a manipulation with norm smaller than it, but does not try to minimize it. This means that, in order to build Table \ref{tab:l2_cifar10} one has to run PGD once for each value $\epsilon$. Conversely, for the other attacks a single run is sufficient to compute the robust accuracy at every threshold.\\
We compare the runtime of the attacks in the setting used for the experiment in Table \ref{tab:l2_cifar10}, and report the total time needed to run the attacks on 500 different points on a single GPU: PGD-10k takes around 18 hours for a single threshold. CW-100k needs 55 hours in total and our method takes 150 hours (using 3 starting points), while the fastest but also weakest attack is DeepFool with a runtime of less than one minute.

\subsection{Choosing parameters}\label{sec:parameters}
%We want to give some heuristics on how choosing the parameters of our attack, that is $p$ in \eqref{eq:psampl}, the number of linear regions $N$ to be checked and the value of $\gamma$ in Algorithm \ref{alg:algorithm-label}.\\
In order to choose a proper parameter $q$ for the sampling scheme in Equation \eqref{eq:psampl} we run our attack on the MNIST \textit{plain} model, already introduced in Section \ref{sec:main_exp}, with $q\in\{0.6, 0.7, 0.8, 0.9, 1.0\}$ and with the scheme proposed in \cite{CroHei18}, where the next linear region to check is chosen by sampling uniformly a direction from the current best solution (corresponding to $q=0.5$). In Figure \ref{fig:psampl} we show the development of median (left), maximum (center) and mean (right) of the $l_2$-norms of the adversarial perturbations found as a function of the explored linear regions. We can see that the final values of the statistics do not differ significantly. %However, since we are interested in getting quickly results close to optimal, $q=0.8$ is preferable as the maximum for this run is the fastest to converge to the final solution.\\
%However, we choose $q=0.8$
{%\color{blue}
Moreover, we repeat the previous experiment, this time varying the value of $\gamma$ in Equation \eqref{eq:sampl_2}. In particular, we test $\gamma = 1, \ldots, 9$ and report in Figure \ref{fig:stats_gammas} median (left), maximum (center) and mean (right) of the $l_2$-norms of the adversarial perturbations found as a function of the explored linear regions. We can see that, while for $\gamma=1$ the results are much worse and for $\gamma=2$ the convergence to the final solution is significantly slower, the algorithm appears to perform similarly with $\gamma$ between 3 and 9.\\
We also run the experiments on the GTS \textit{plain} model. In Figure \ref{fig:psampl_gts} one can see that higher values of $q$ lead to faster convergence to the final solutions (we fix $\gamma=9$). In Figure \ref{fig:stats_gammas_gts} we test different values of $\gamma$ between 1 and 9 keeping constant $q=0.8$. We notice that all the runs achieve similar performance, even for small values of $\gamma$ differently from what happens on MNIST.
This observation, together with the fact that about 15 regions are sufficient for the results to be almost indistinguishable from the final ones, suggests that this model is easier to attack than the one on MNIST.\\
Thus we choose to set $q=0.8$ as for smaller values the runs converge slightly more slowly, while for $q=1.0$ the maximum appears to be marginally suboptimal (anyway we want to highlight that the results are in the end almost identical).\\
Finally, from these ablation studies one can also appreciate the stability of the method with respect to the random part inherent to the algorithm. In fact, with the exception of the case $\gamma=1$ on MNIST, in all the runs obtained varying the parameters $q$ and $\gamma$, median, maximum and mean converge to the same or very similar values, meaning that sampling different points and then possibly checking different regions does not lead to inconsistent results.}

\begin{figure*}[p]
	\centering
	%\begin{tabularx}{\textwidth}%{cccccccc}
	%{C{0.25\columnwidth}C{0.25\columnwidth}C{0.25\columnwidth} C{0.25\columnwidth} C{0.25\columnwidth} C{0.25\columnwidth} C{0.25\columnwidth} C{0.25\columnwidth}}
	%starting image & image after linear search & \multicolumn{4}{c}{ \textbf{intermediate images}} & final output & target image \\
	%\end{tabularx}
	\hspace{-1mm}\includegraphics[width=2.01\columnwidth]{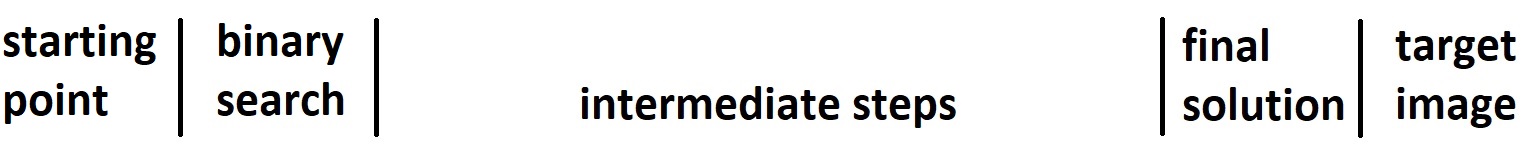}
	\\
	\vspace{4mm}
	%\begin{tabular}{c c c | c}
	\includegraphics[width=0.25\columnwidth]{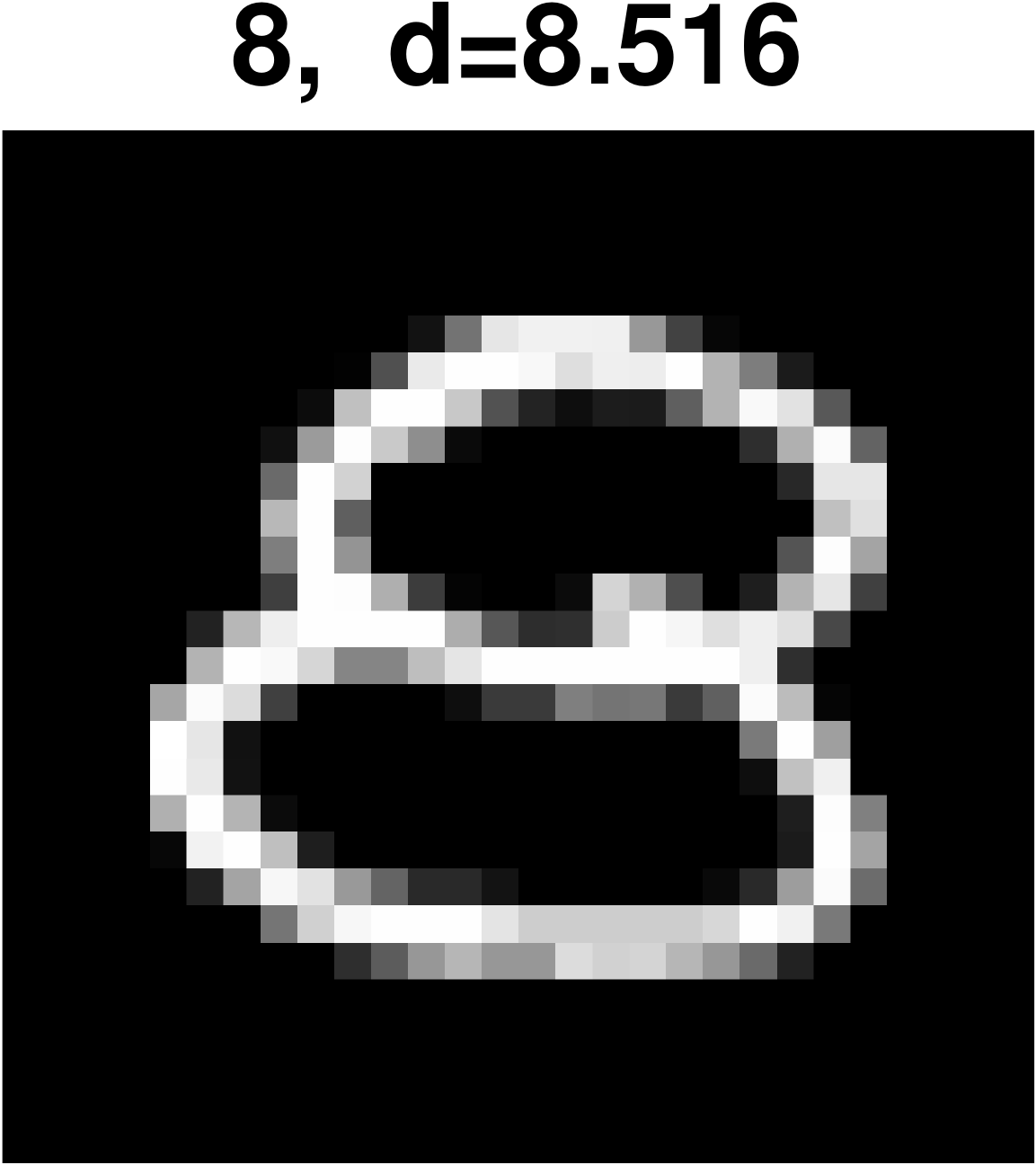}
	\includegraphics[width=0.25\columnwidth]{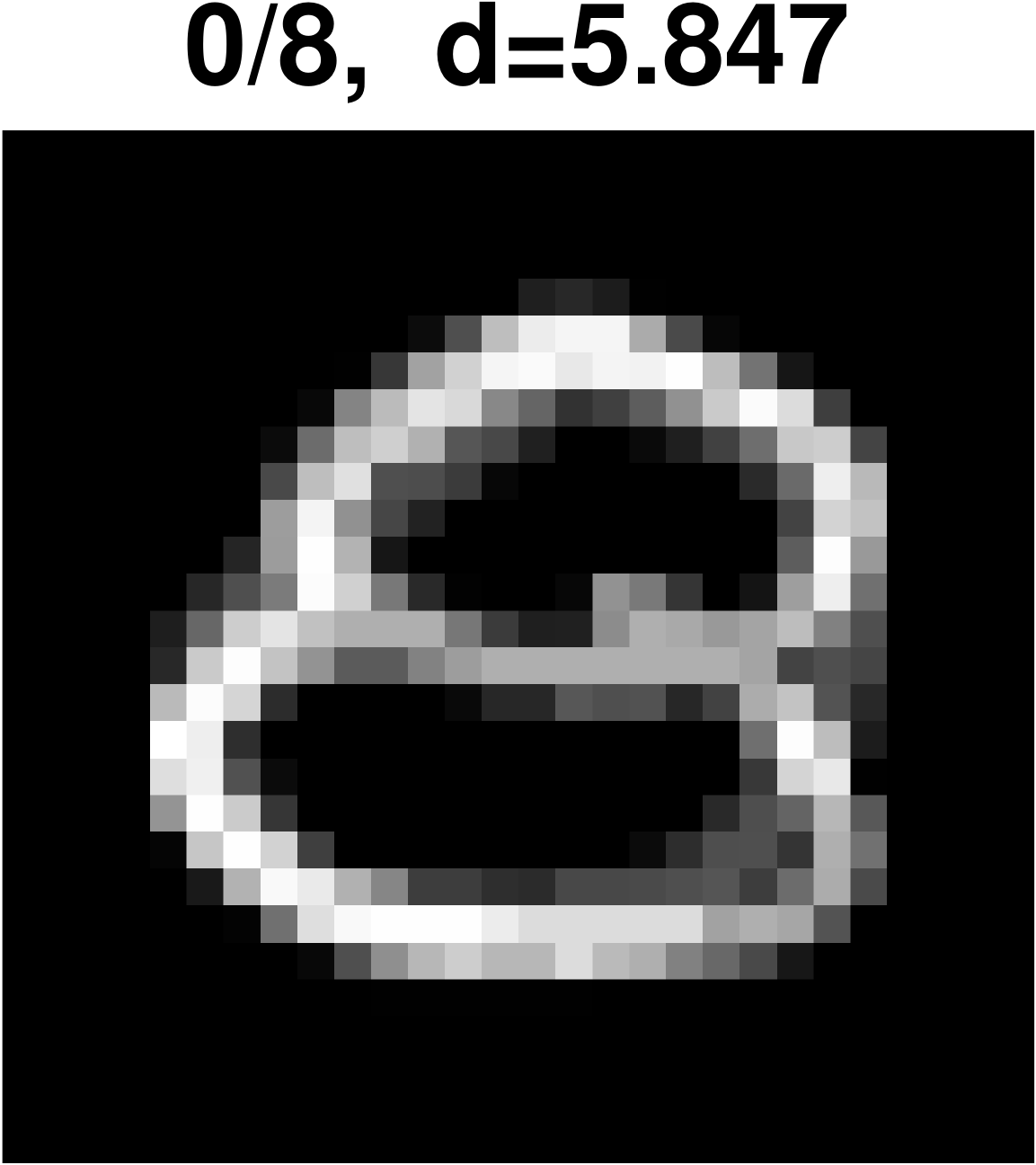}
	\includegraphics[width=0.25\columnwidth]{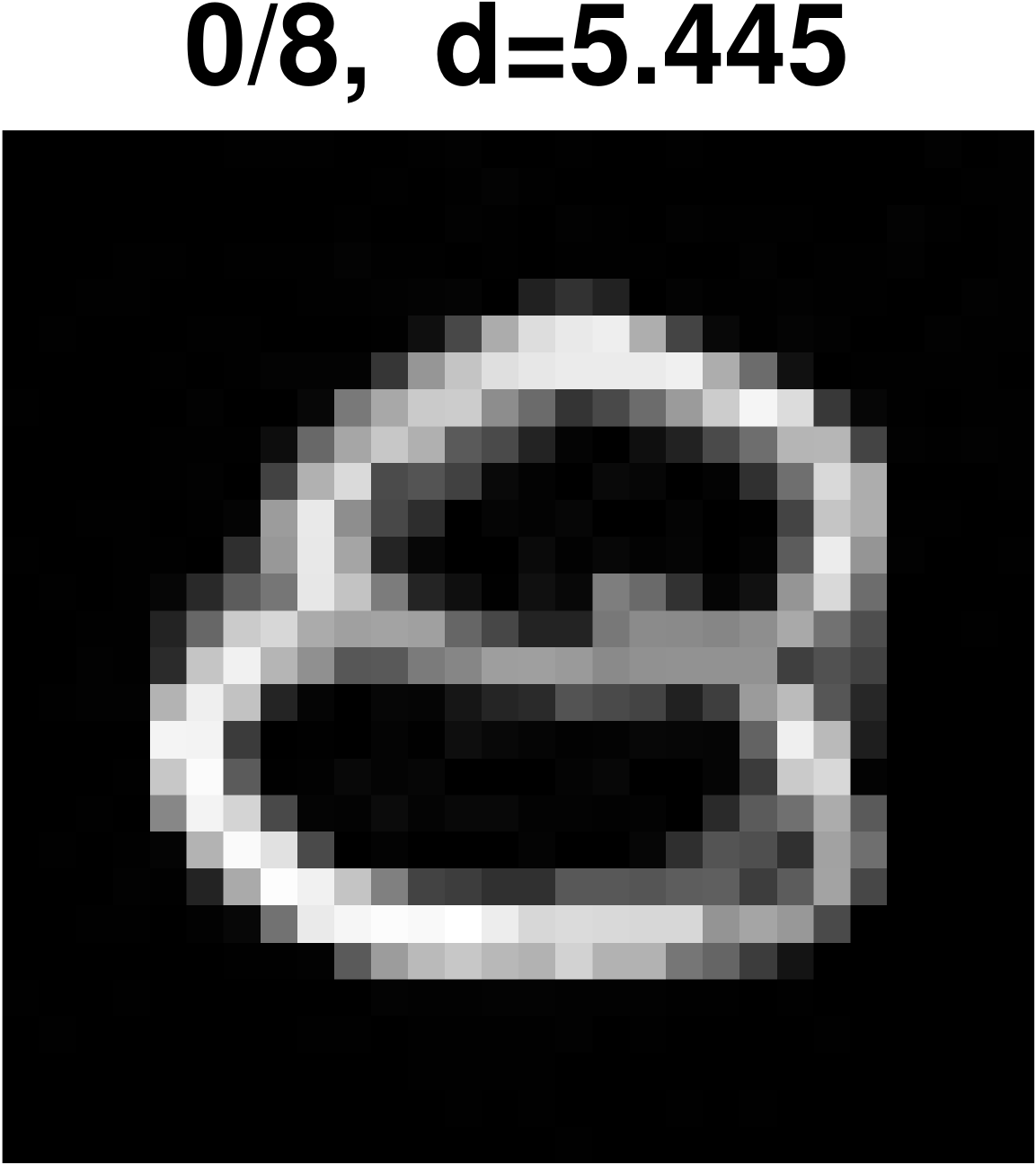}
	\includegraphics[width=0.25\columnwidth]{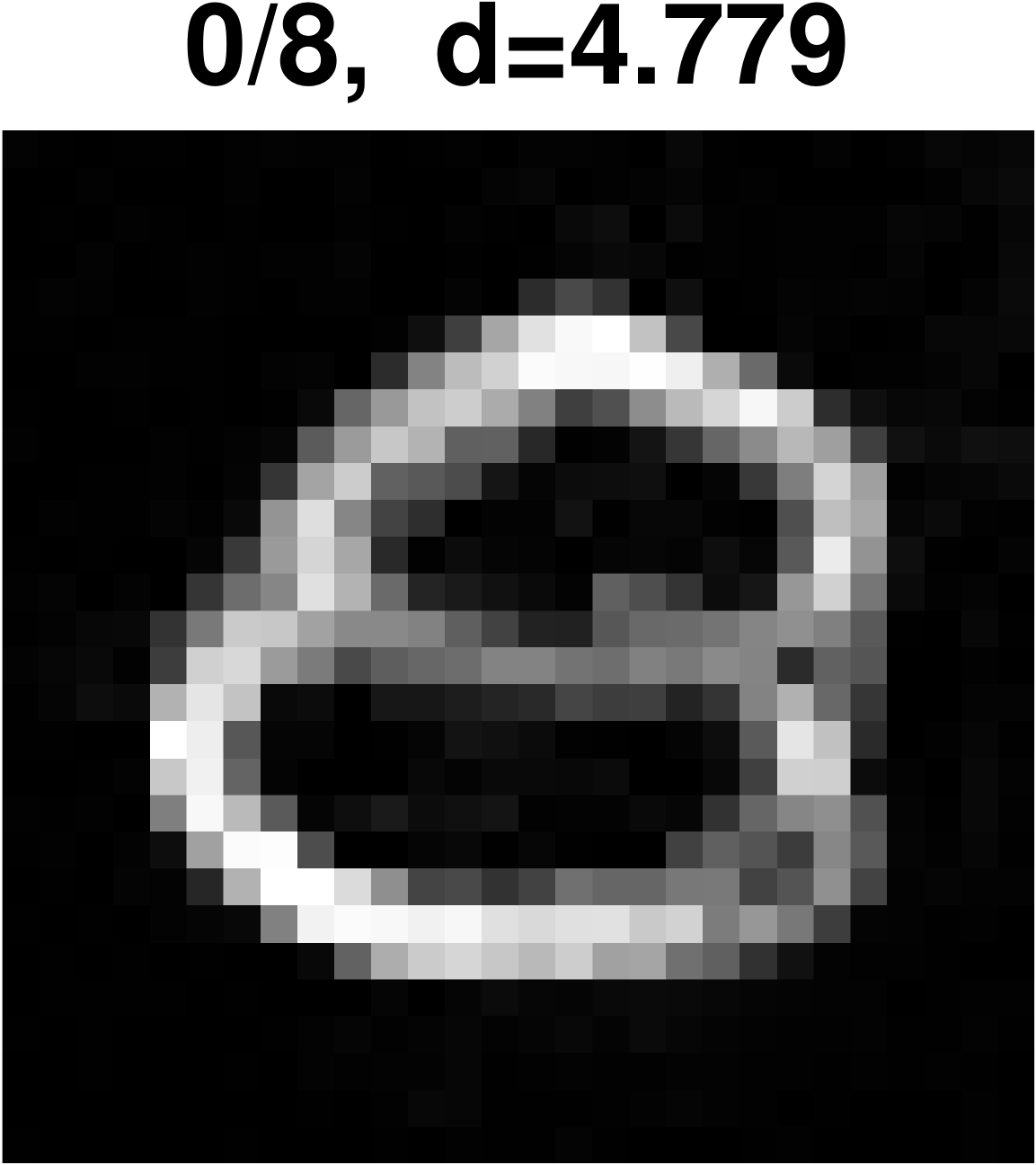}
	\includegraphics[width=0.25\columnwidth]{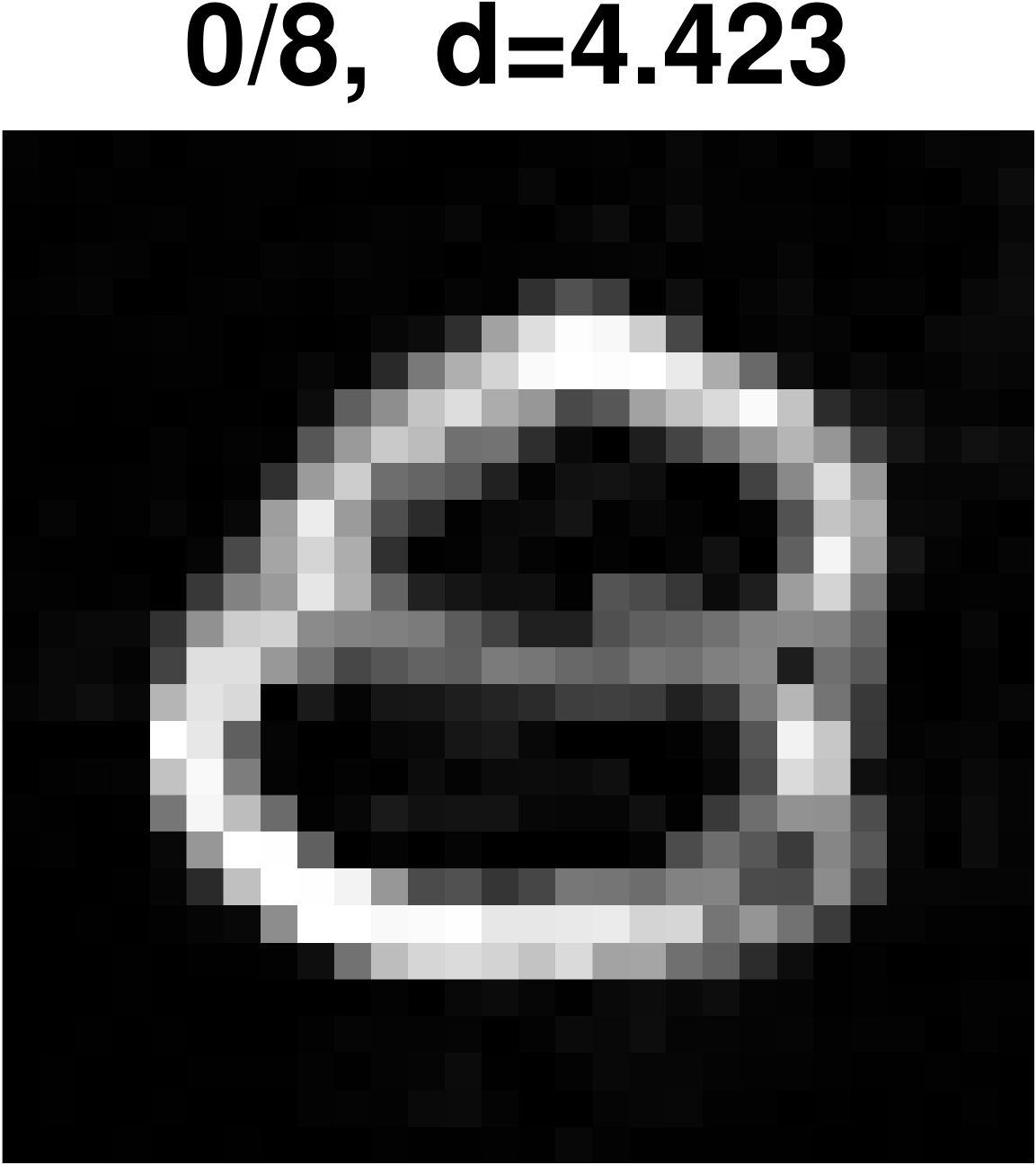}
	\includegraphics[width=0.25\columnwidth]{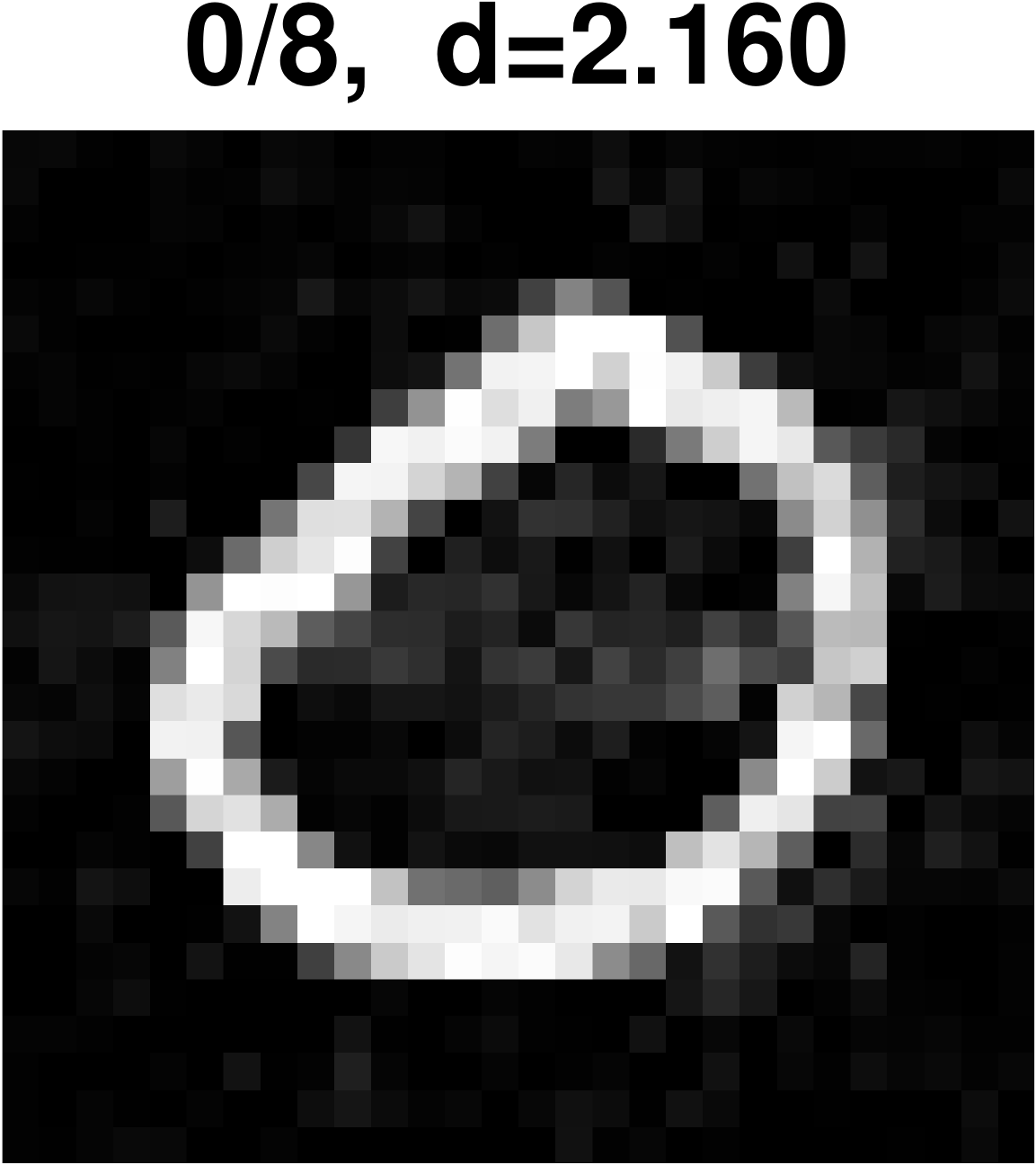}
	\includegraphics[width=0.25\columnwidth]{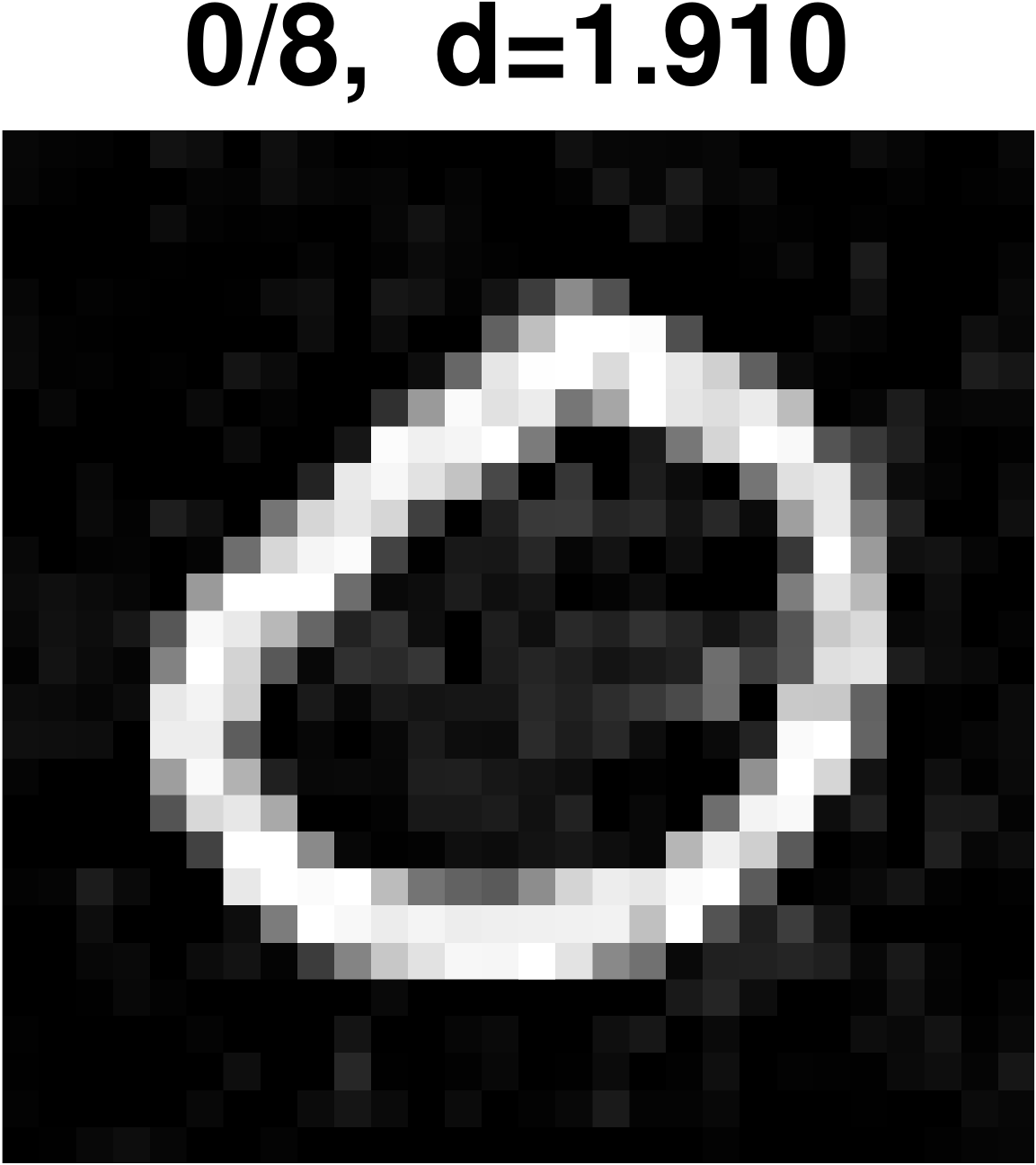}
	\includegraphics[width=0.25\columnwidth]{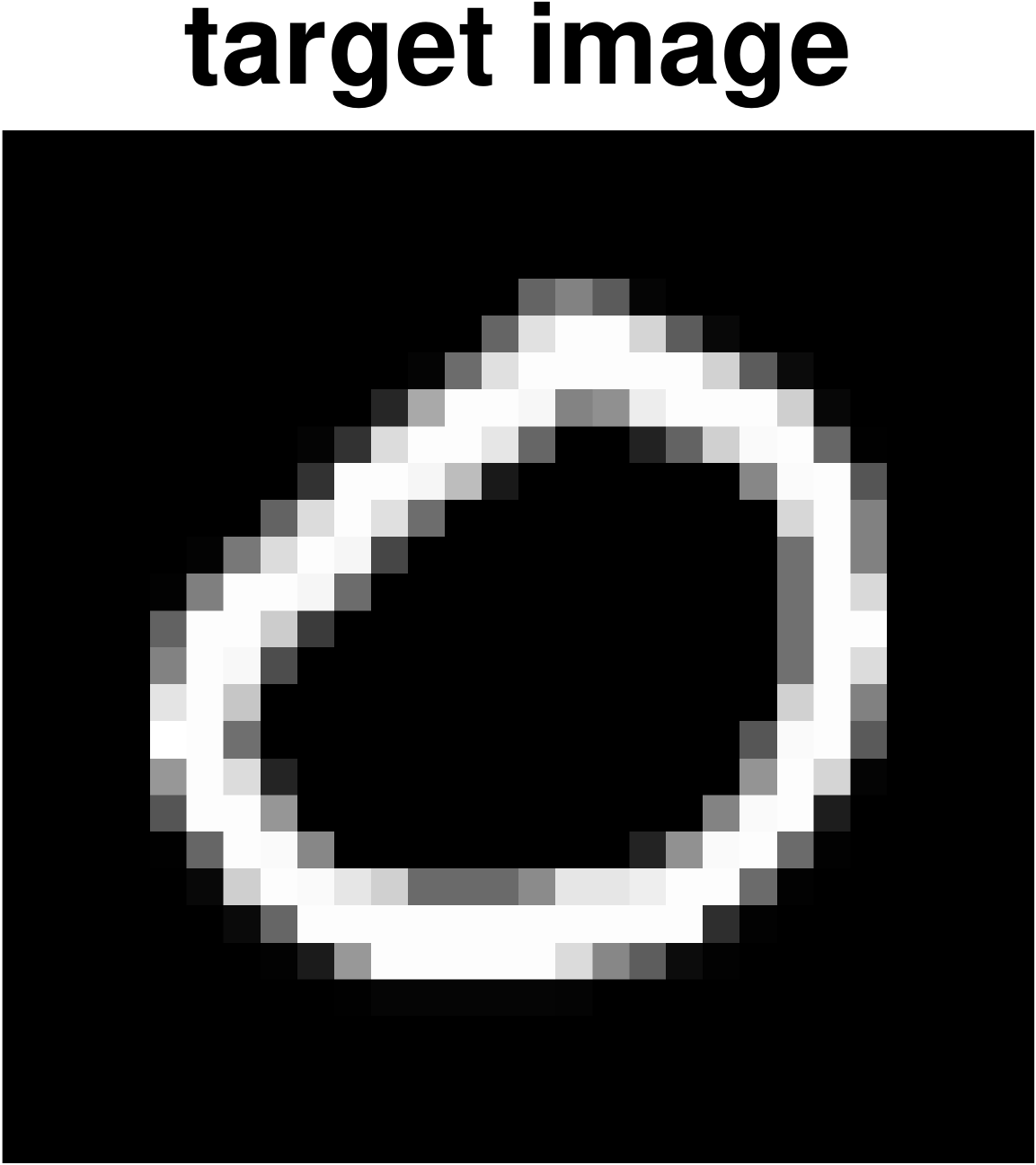}
	\\
	\vspace{4mm}
	\includegraphics[width=0.25\columnwidth]{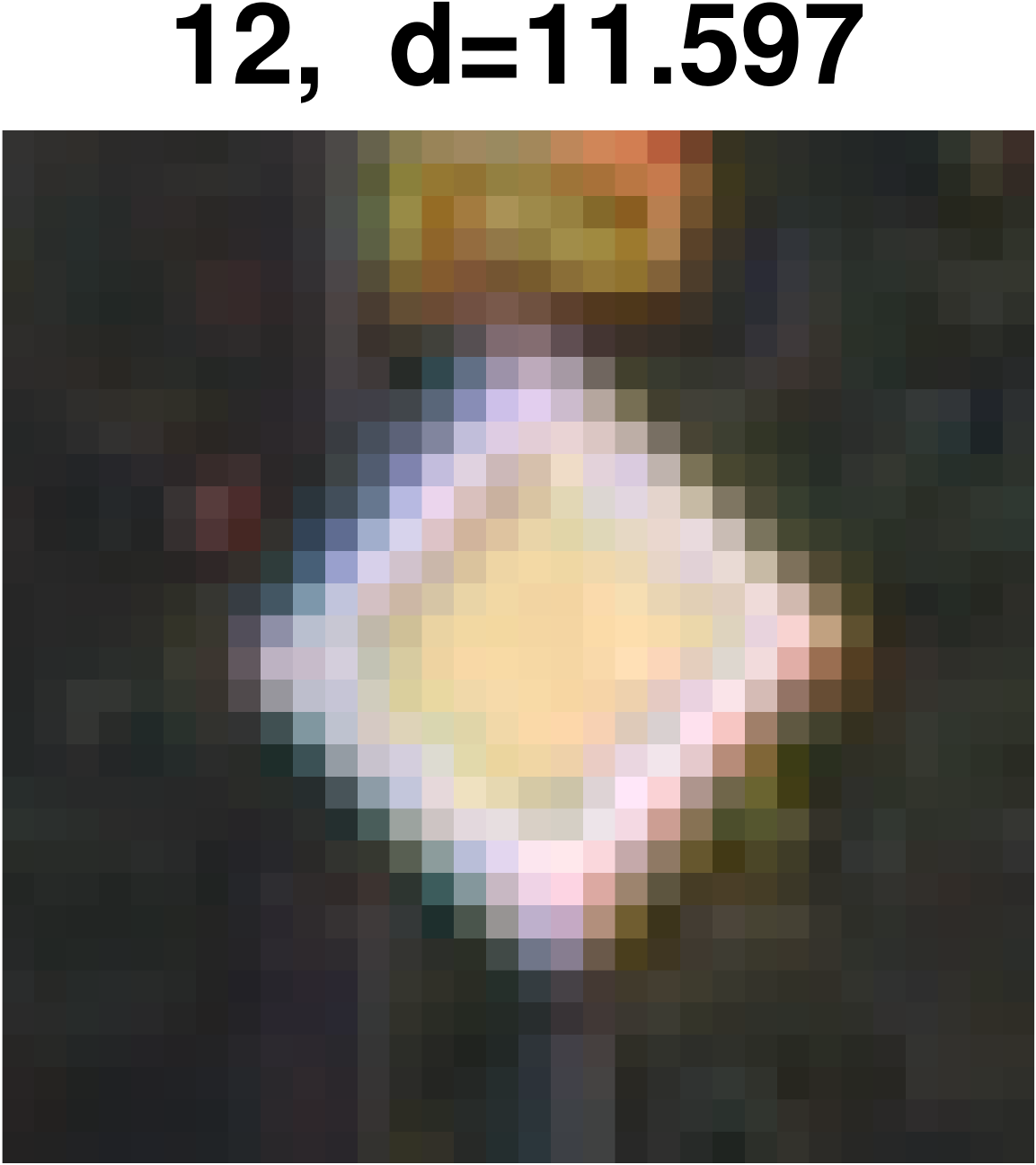}
	\includegraphics[width=0.25\columnwidth]{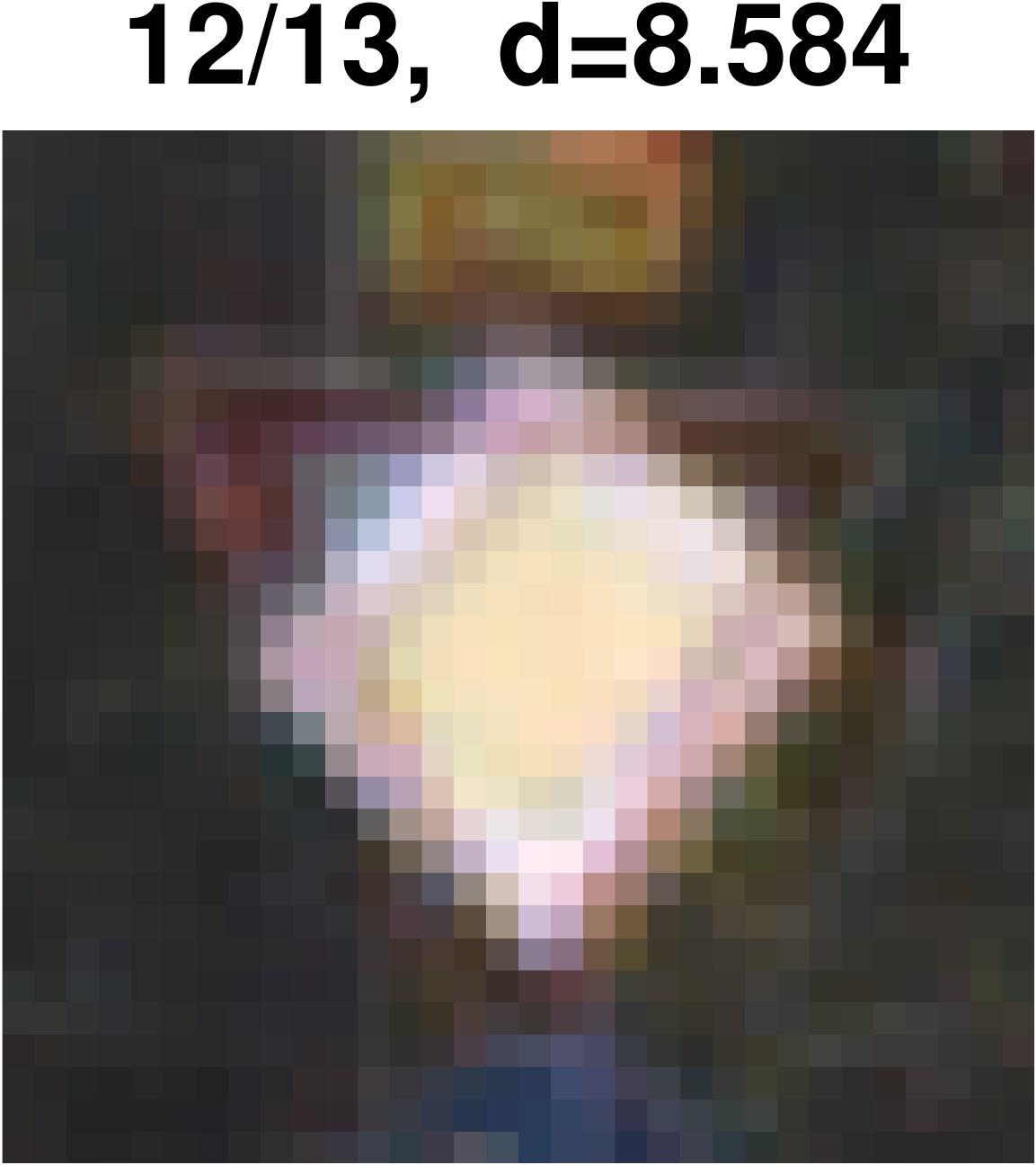}
	\includegraphics[width=0.25\columnwidth]{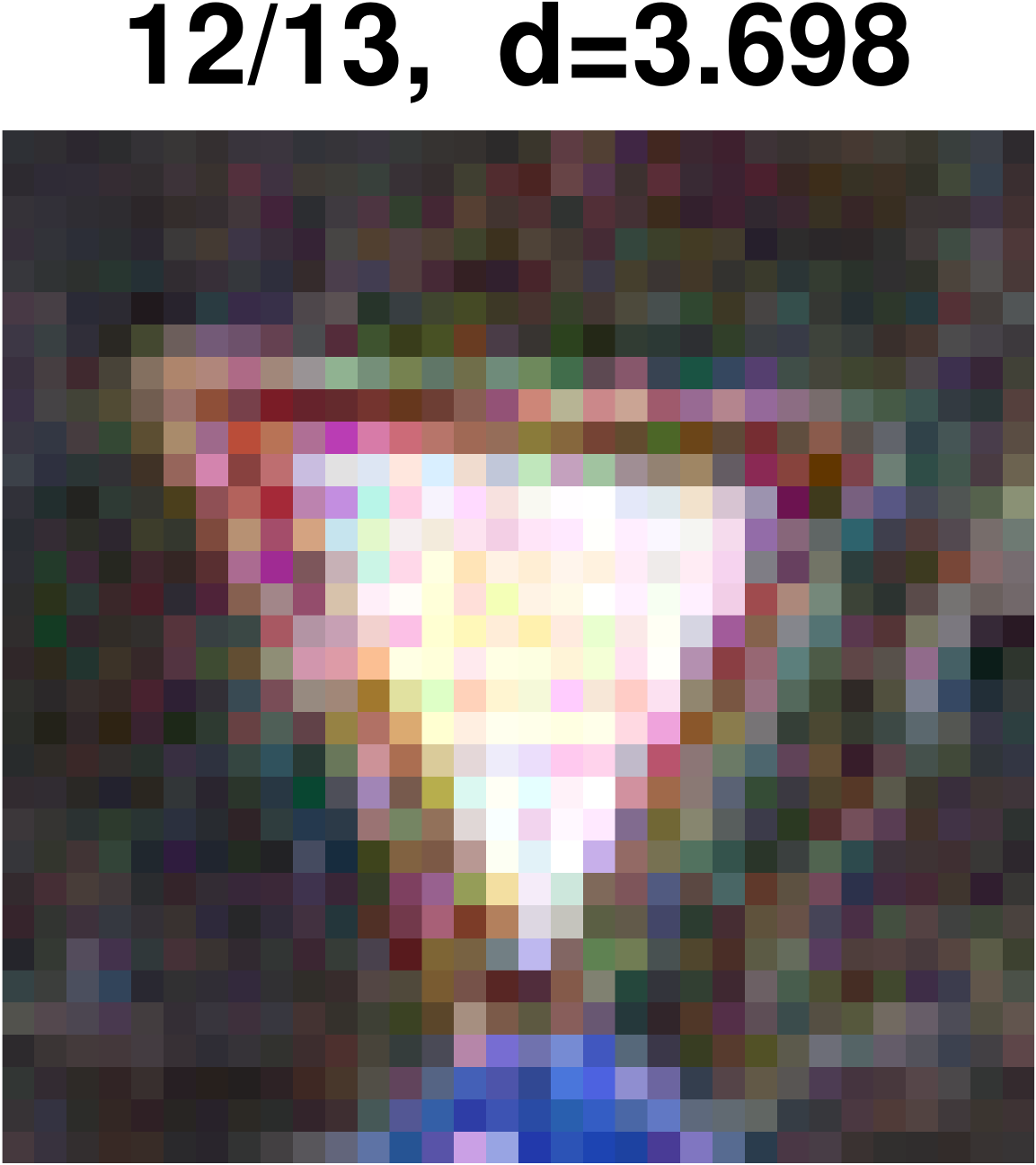}
	\includegraphics[width=0.25\columnwidth]{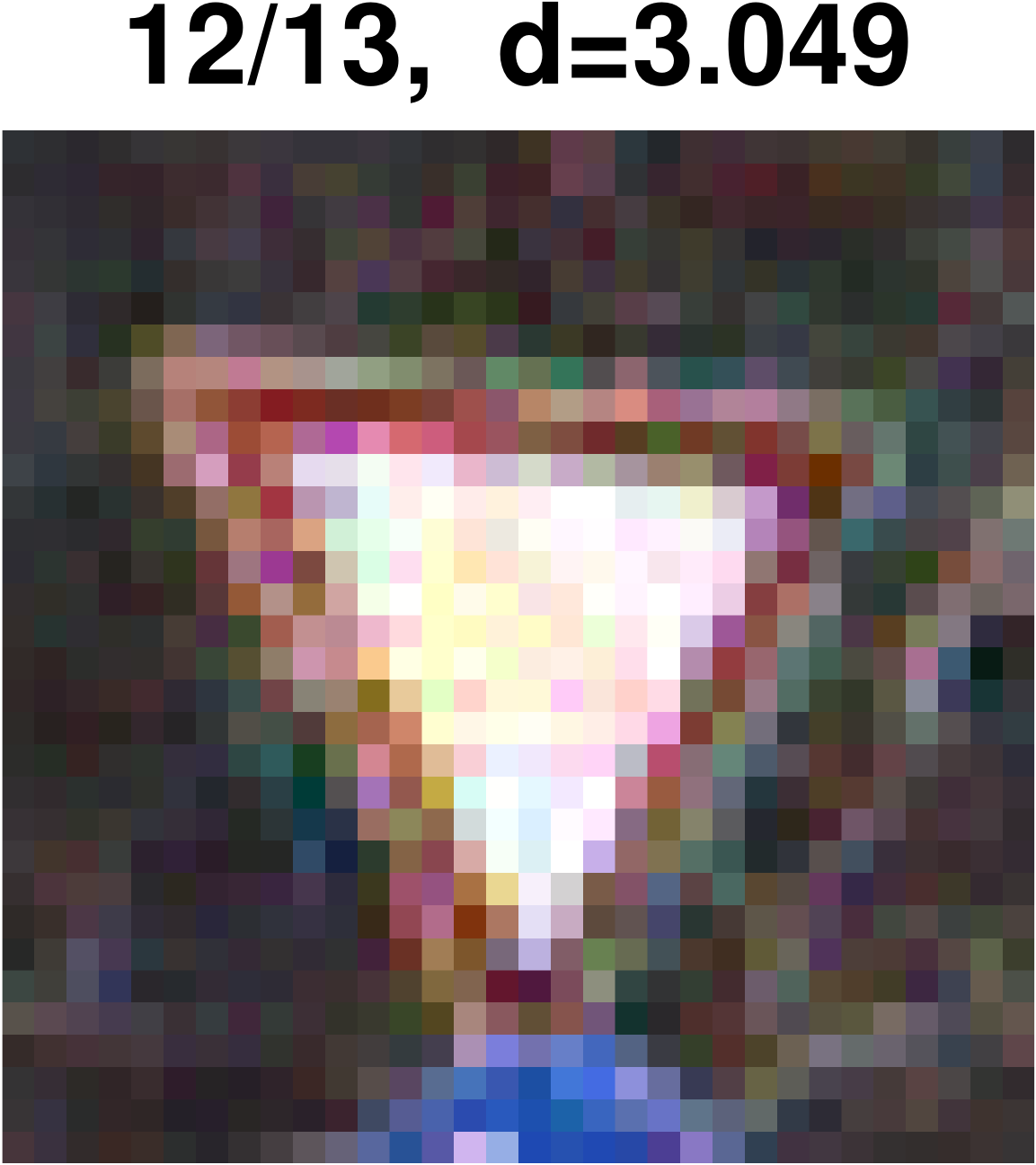}
	\includegraphics[width=0.25\columnwidth]{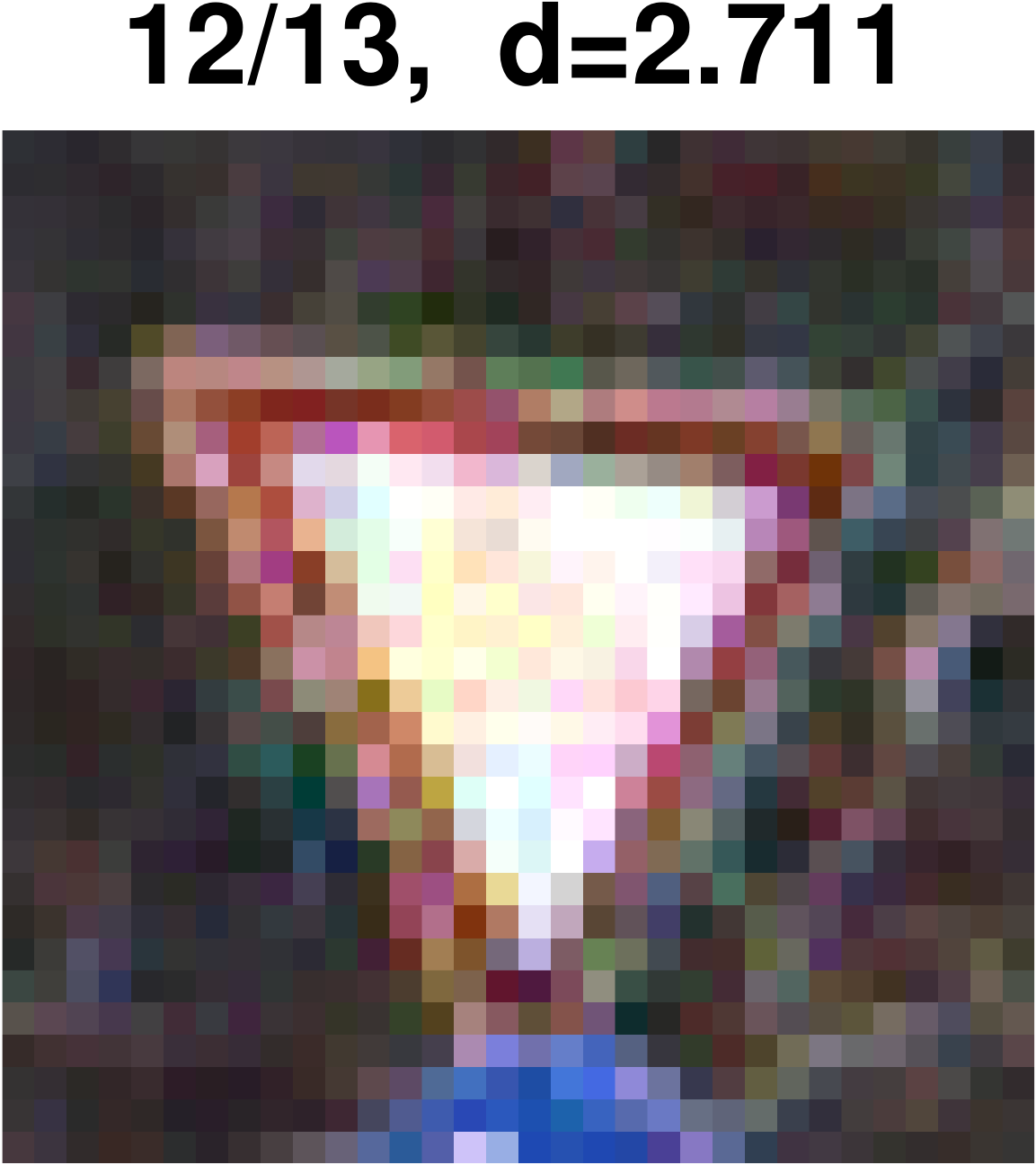}
	\includegraphics[width=0.25\columnwidth]{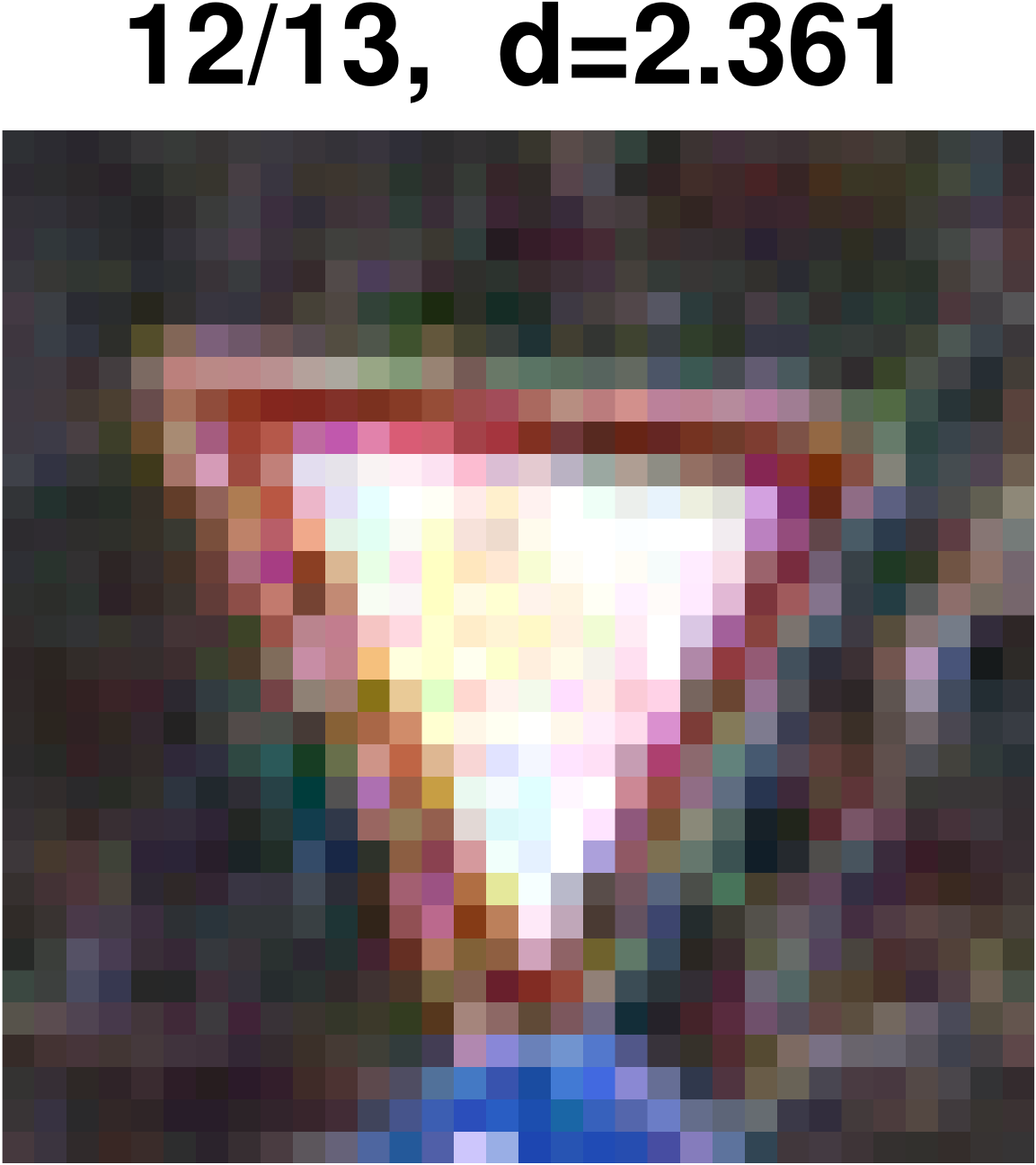}
	\includegraphics[width=0.25\columnwidth]{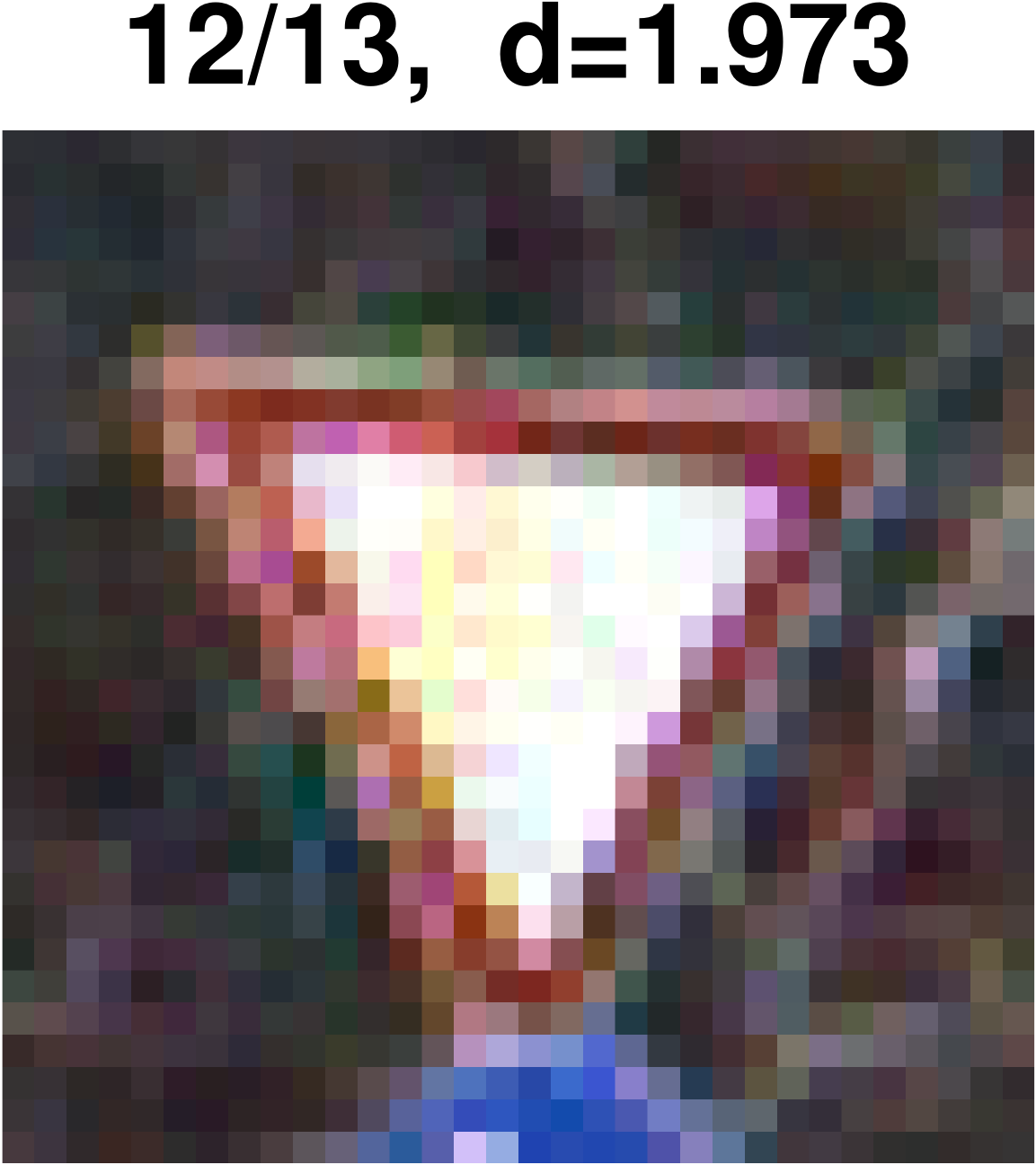}
	\includegraphics[width=0.25\columnwidth]{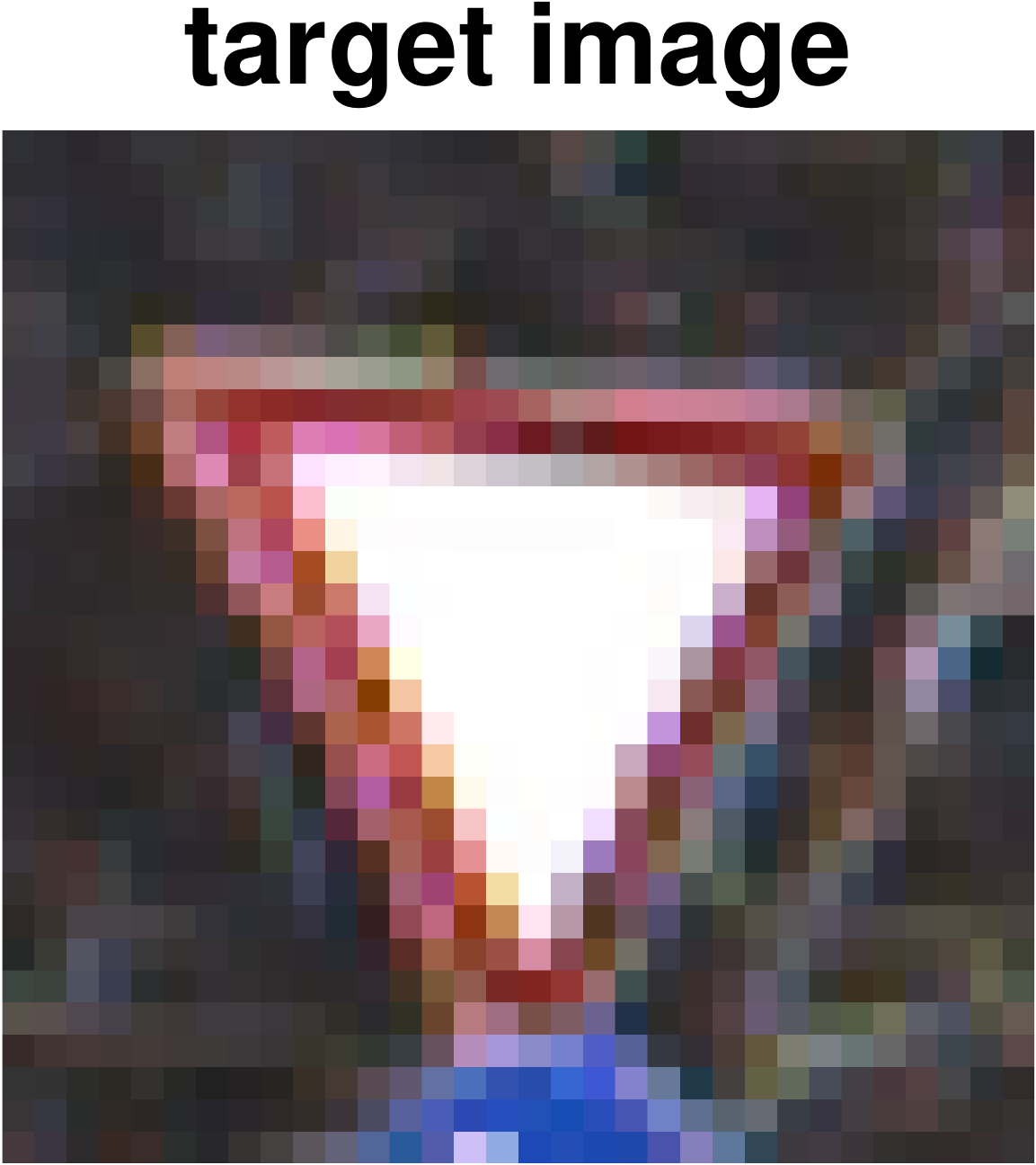}
	\\
	\vspace{4mm}
	\includegraphics[width=0.25\columnwidth]{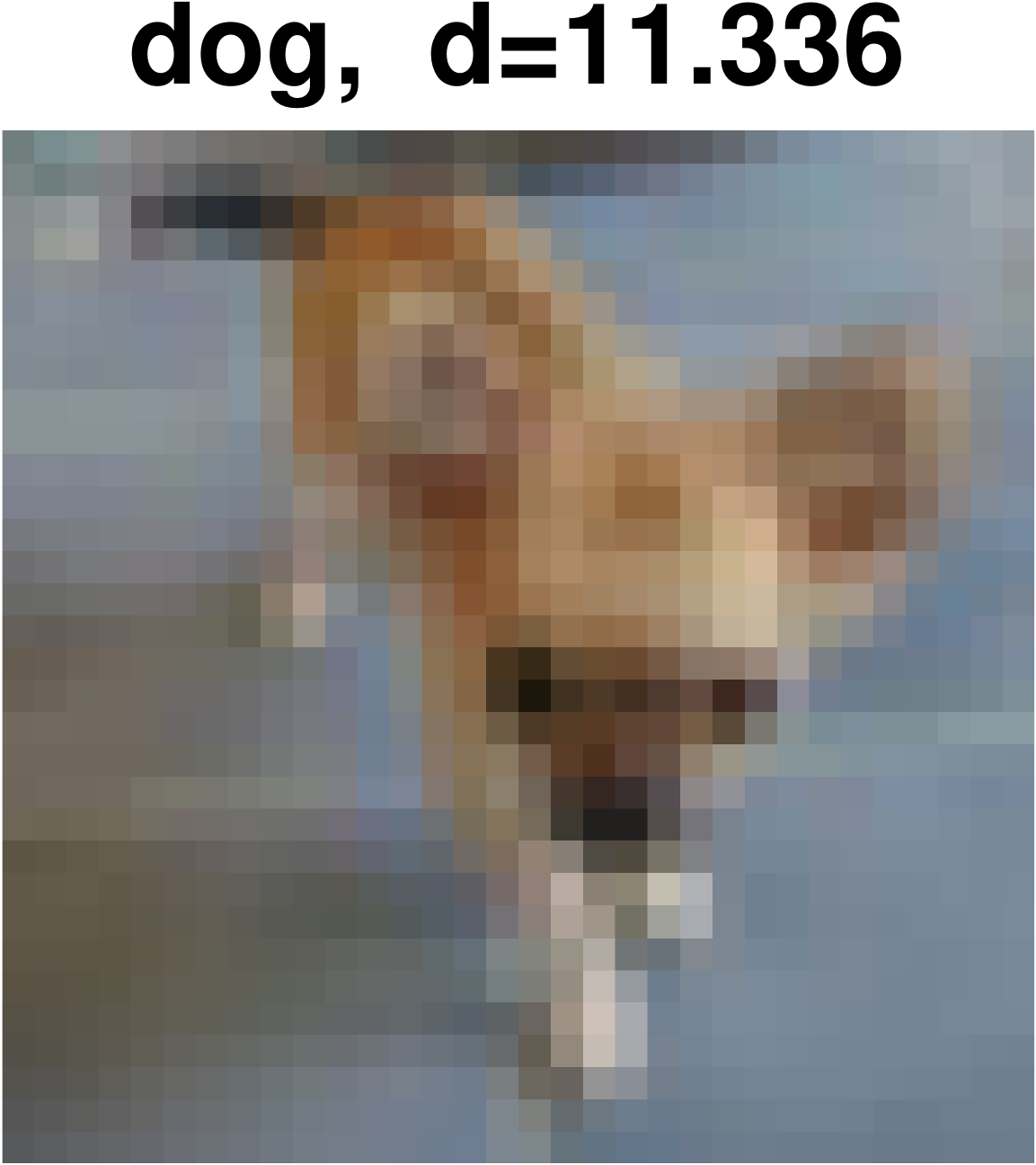}
	\includegraphics[width=0.25\columnwidth]{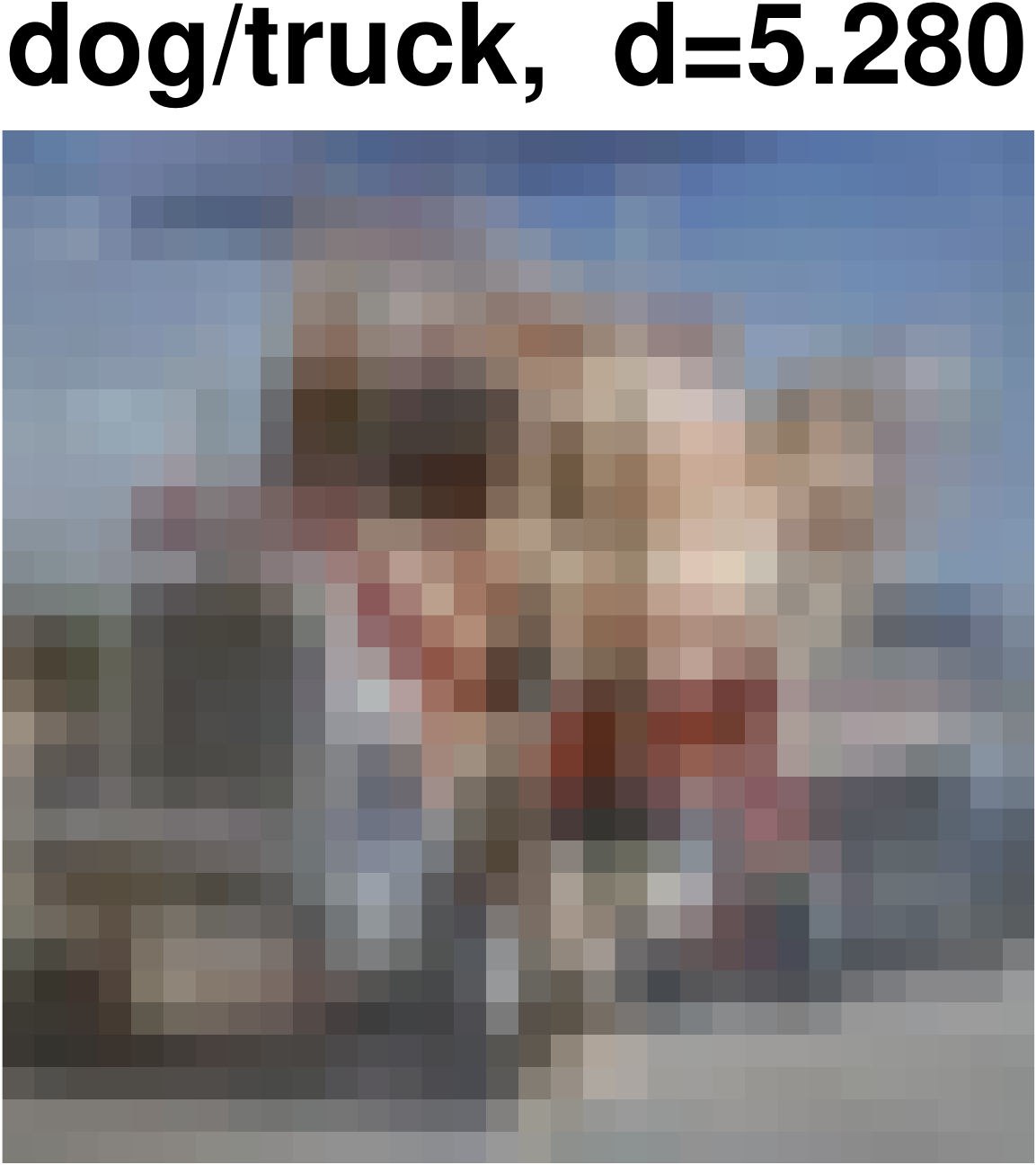}
	\includegraphics[width=0.25\columnwidth]{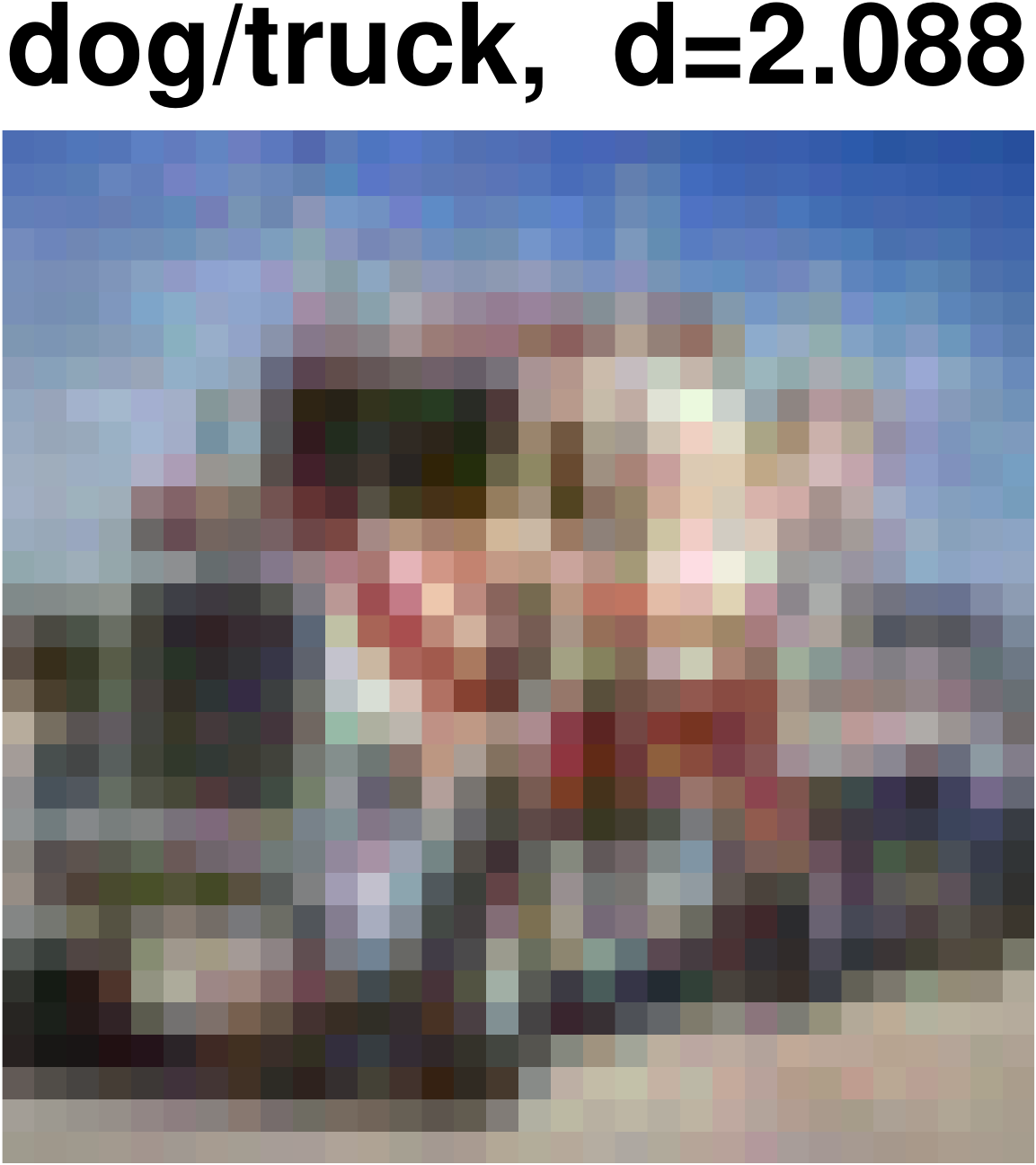}
	\includegraphics[width=0.25\columnwidth]{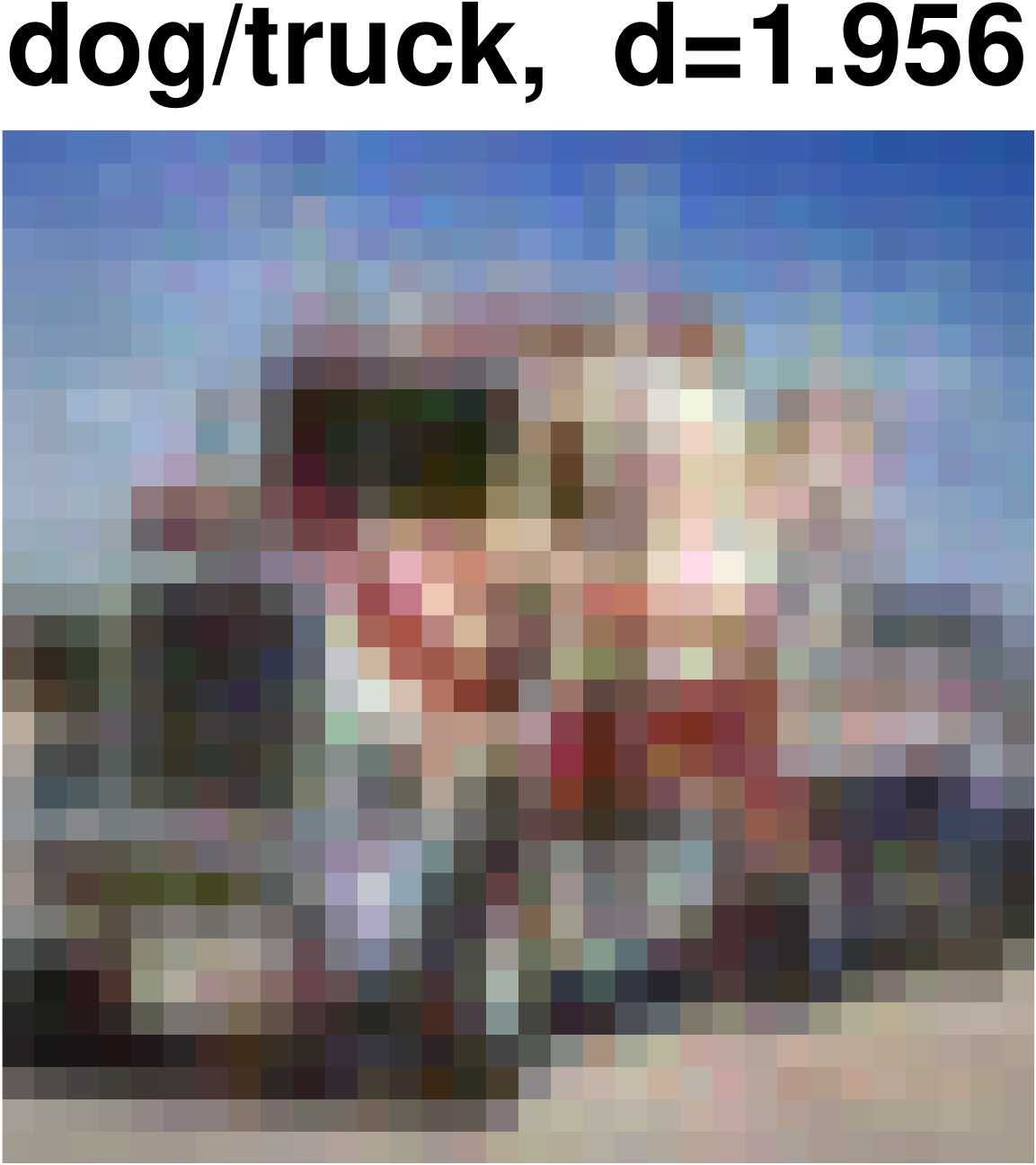}
	\includegraphics[width=0.25\columnwidth]{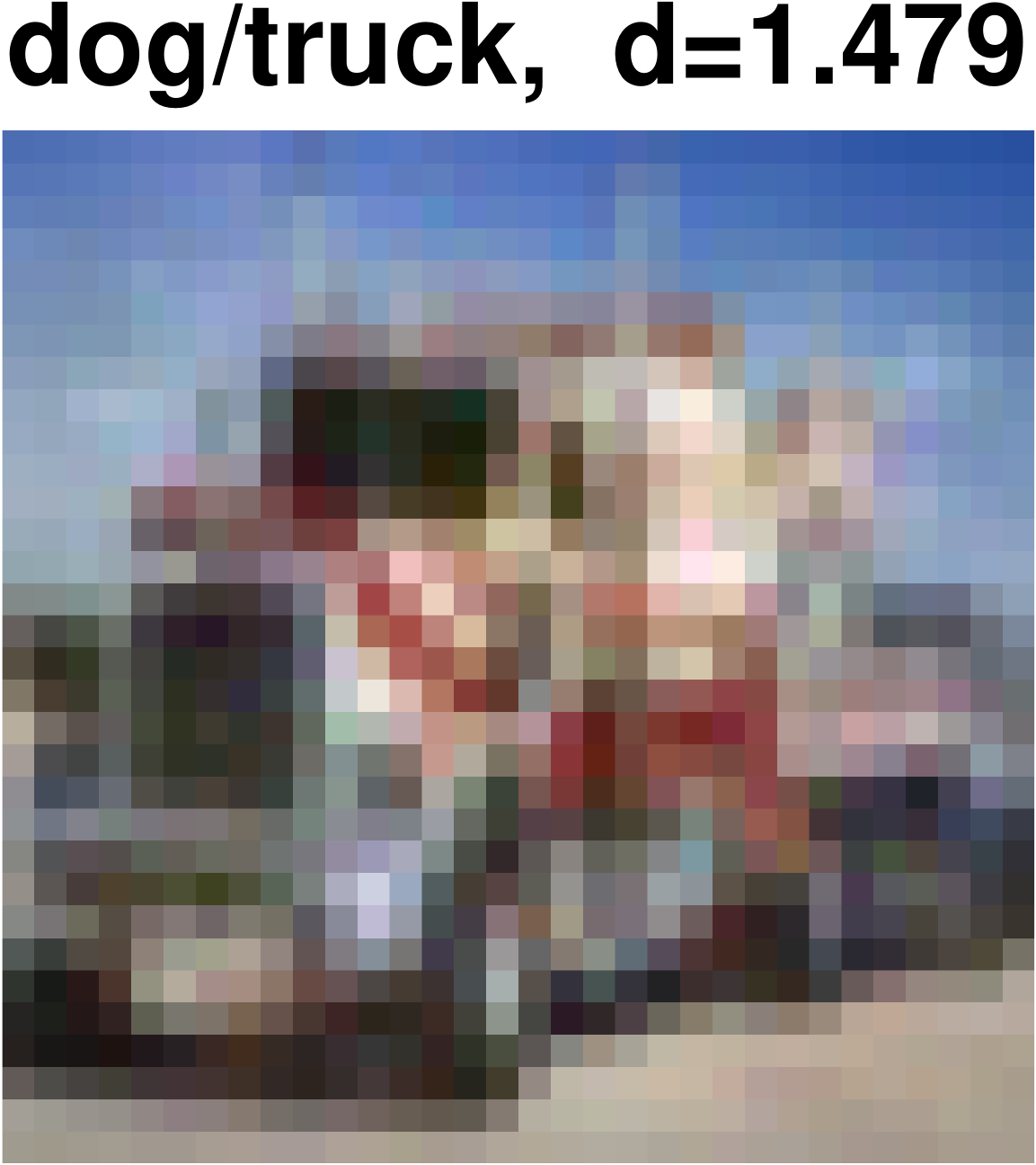}
	\includegraphics[width=0.25\columnwidth]{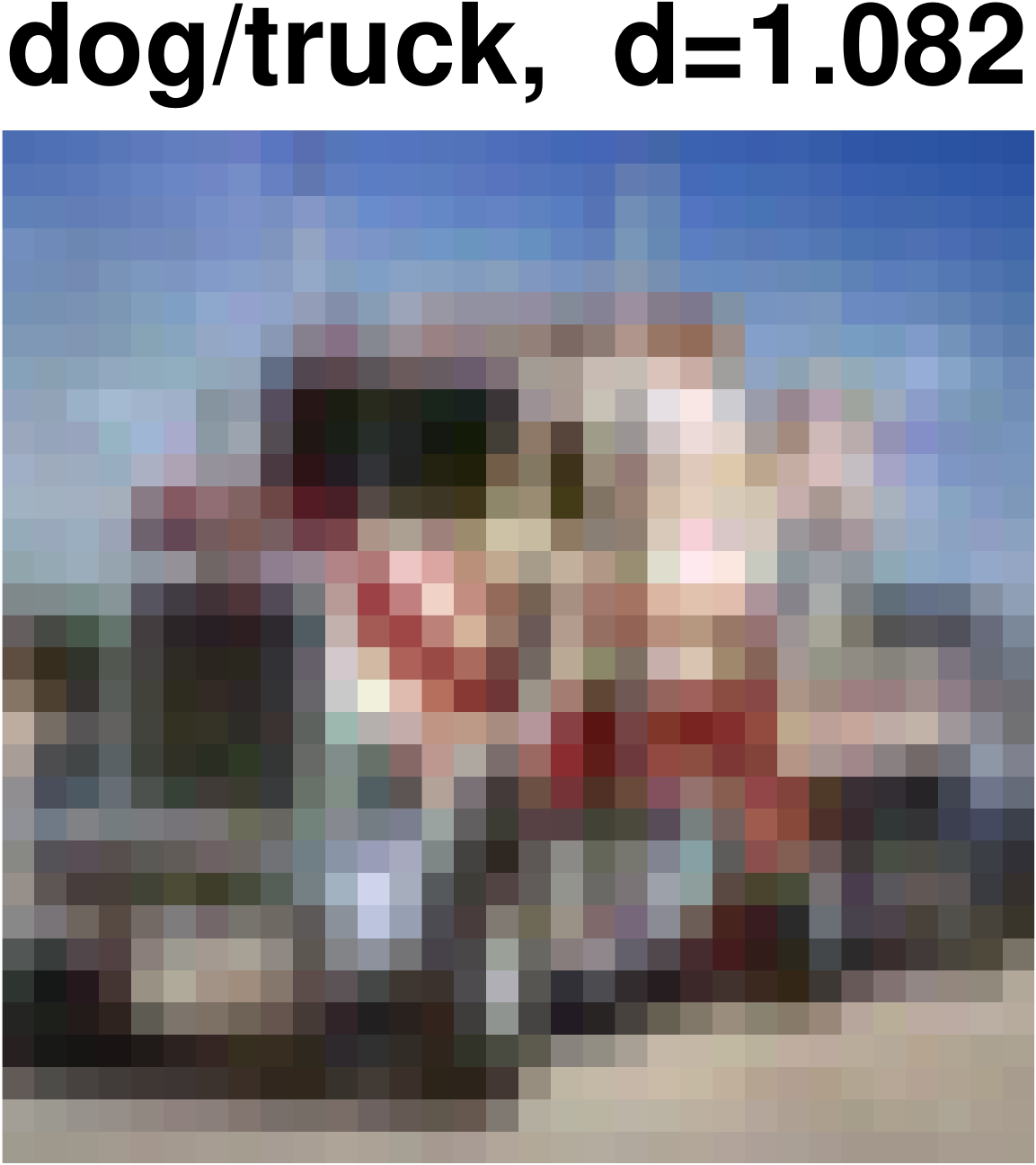}
	\includegraphics[width=0.25\columnwidth]{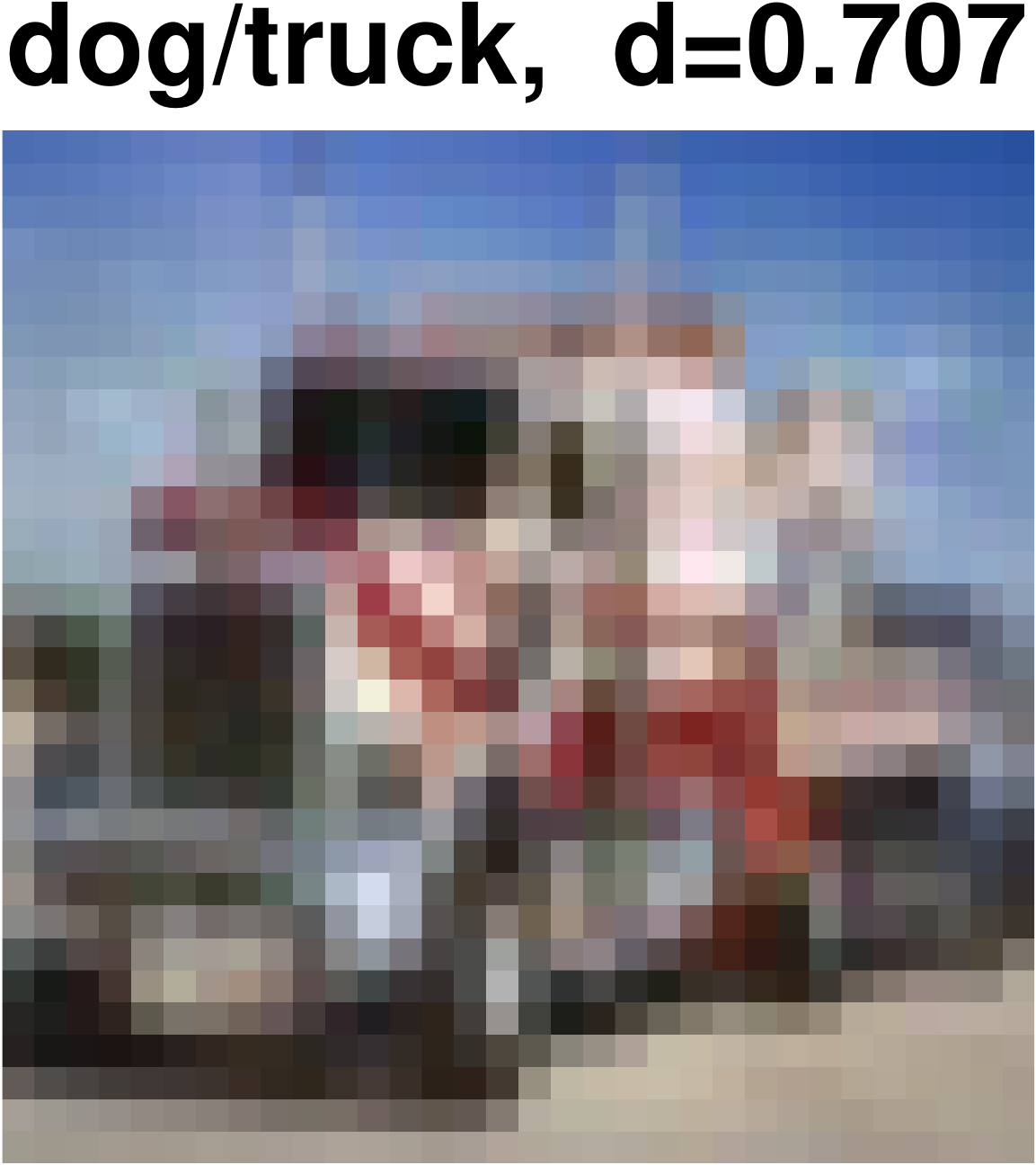}
	%\vline
	\includegraphics[width=0.25\columnwidth]{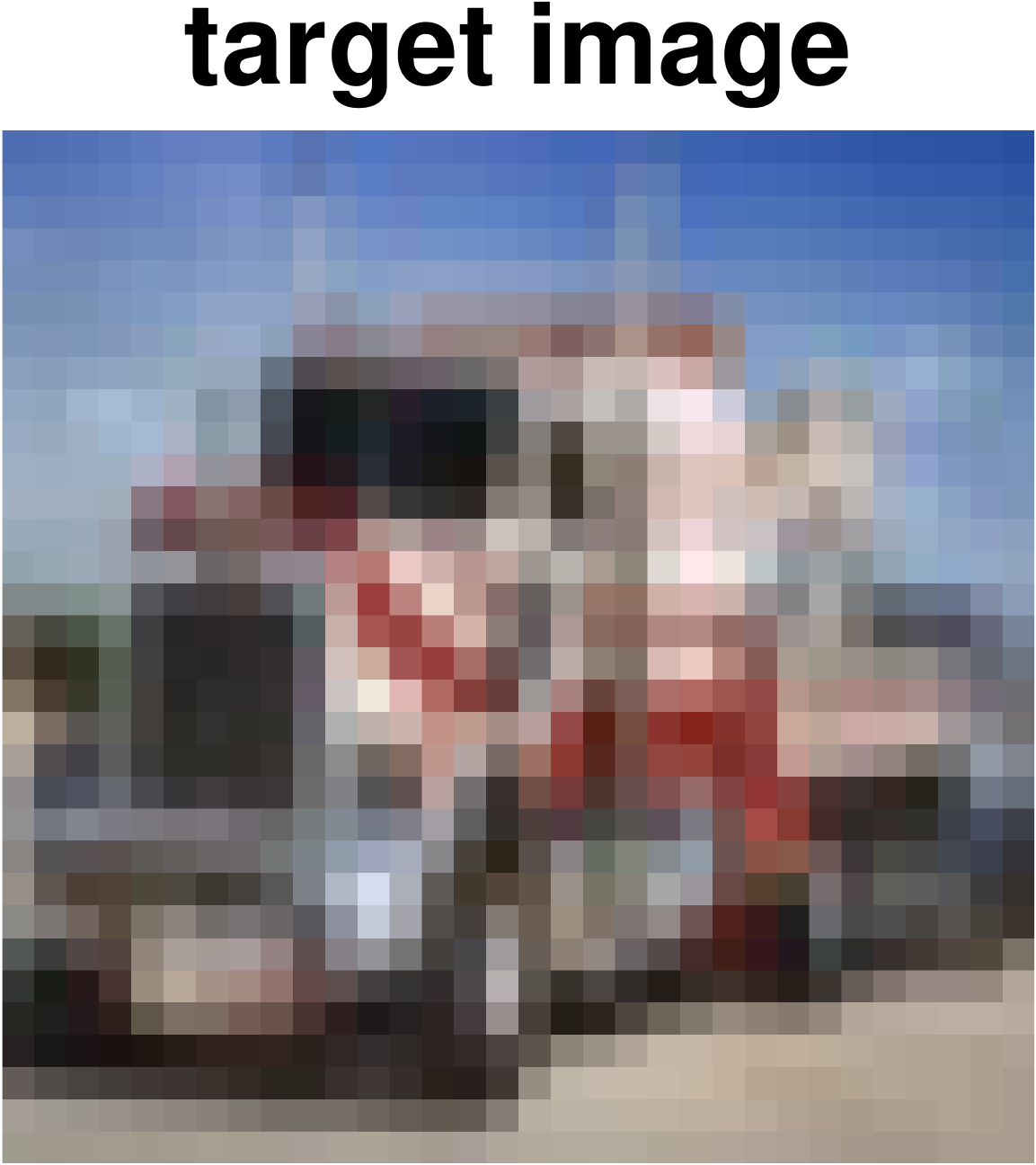}\\
	\vspace{4mm}
	\includegraphics[width=0.25\columnwidth]{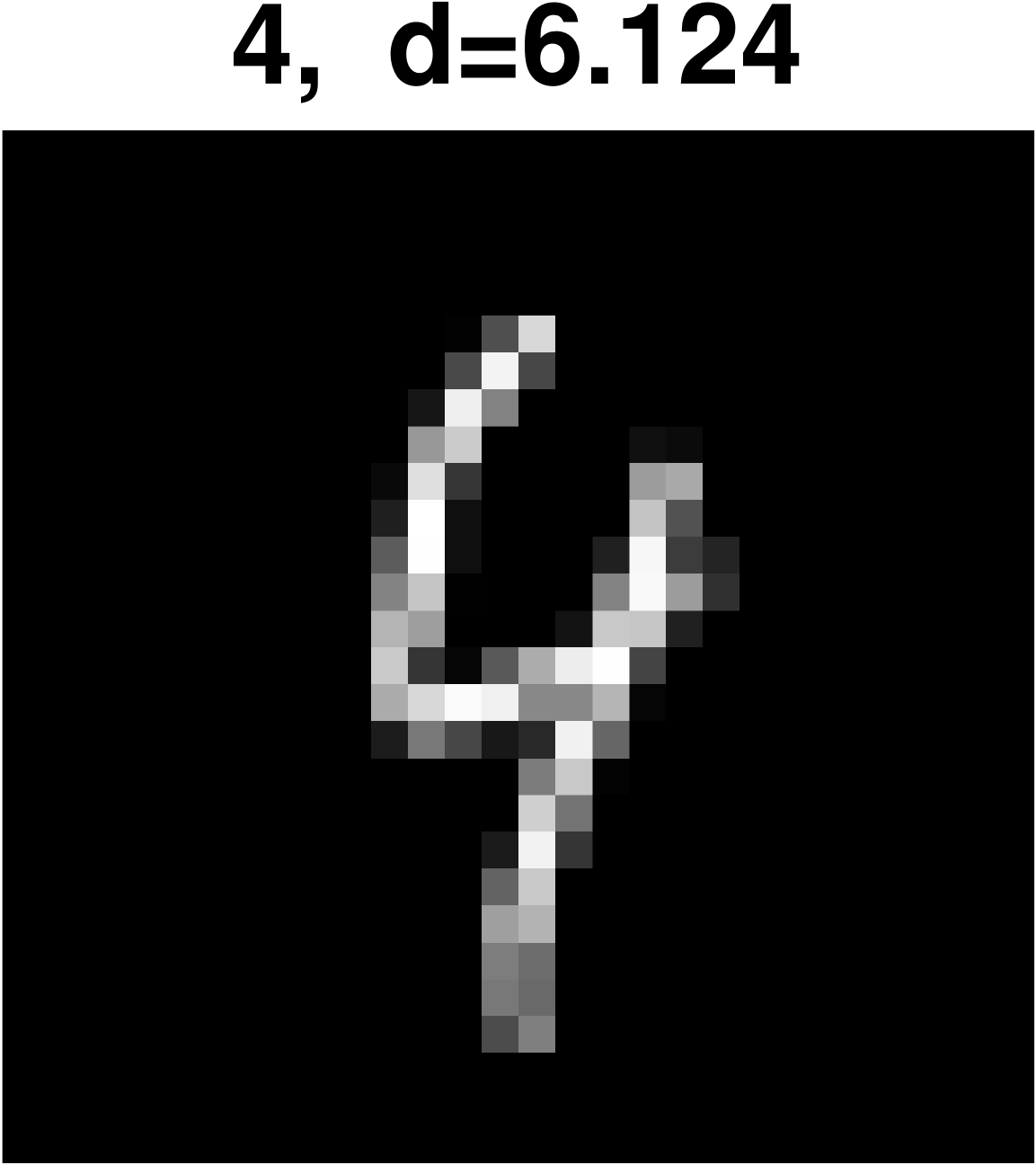}
	\includegraphics[width=0.25\columnwidth]{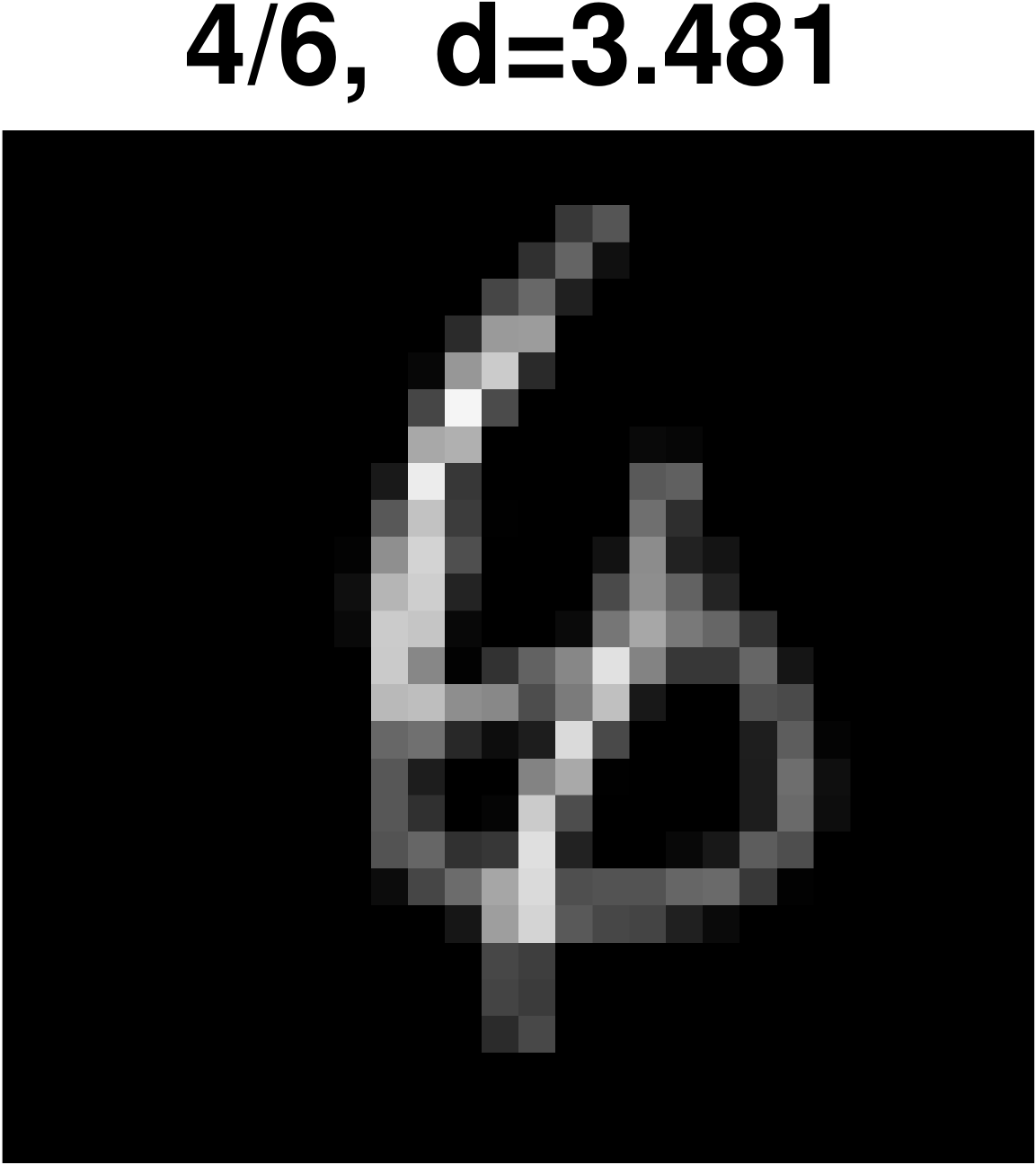}
	\includegraphics[width=0.25\columnwidth]{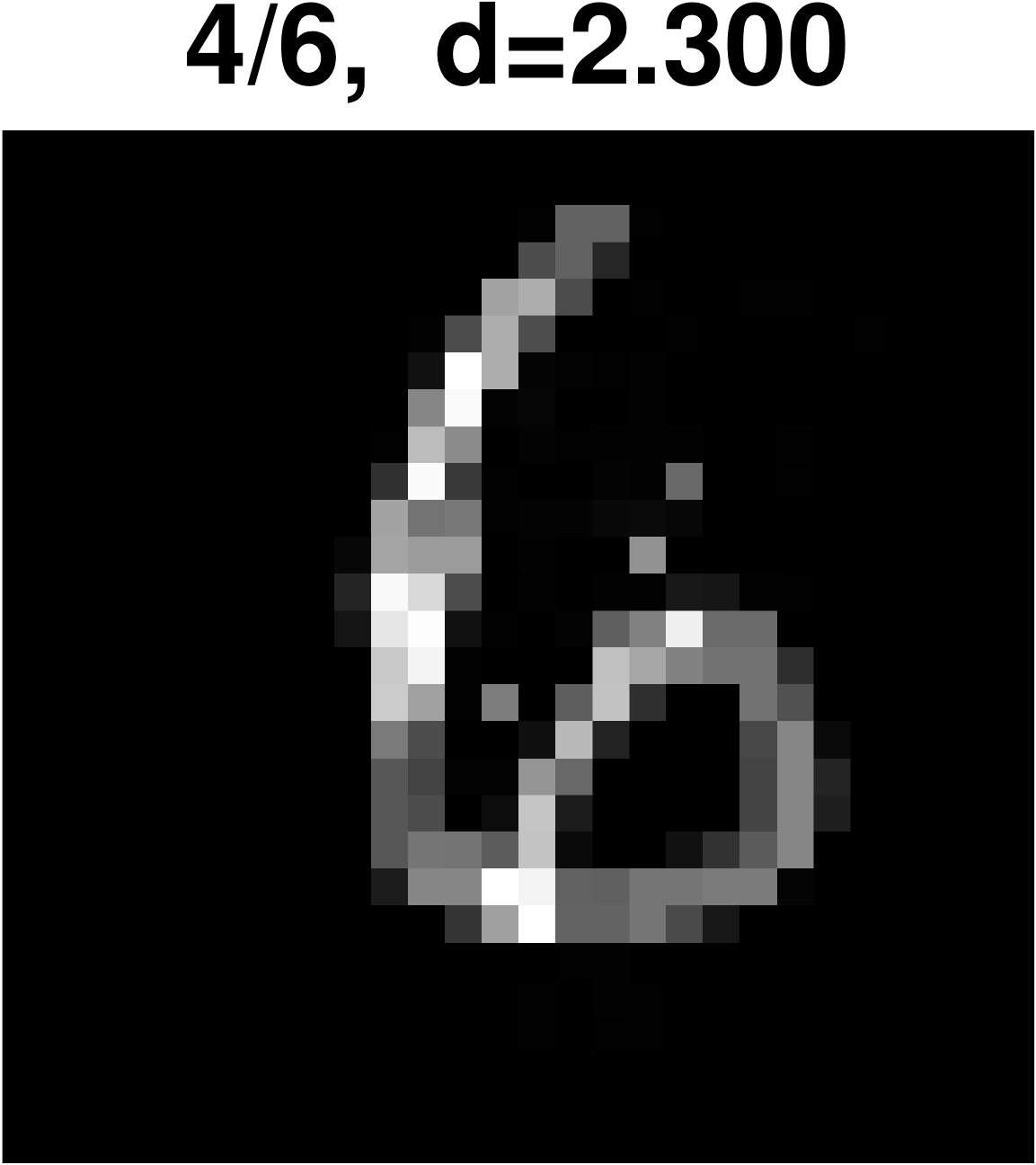}
	\includegraphics[width=0.25\columnwidth]{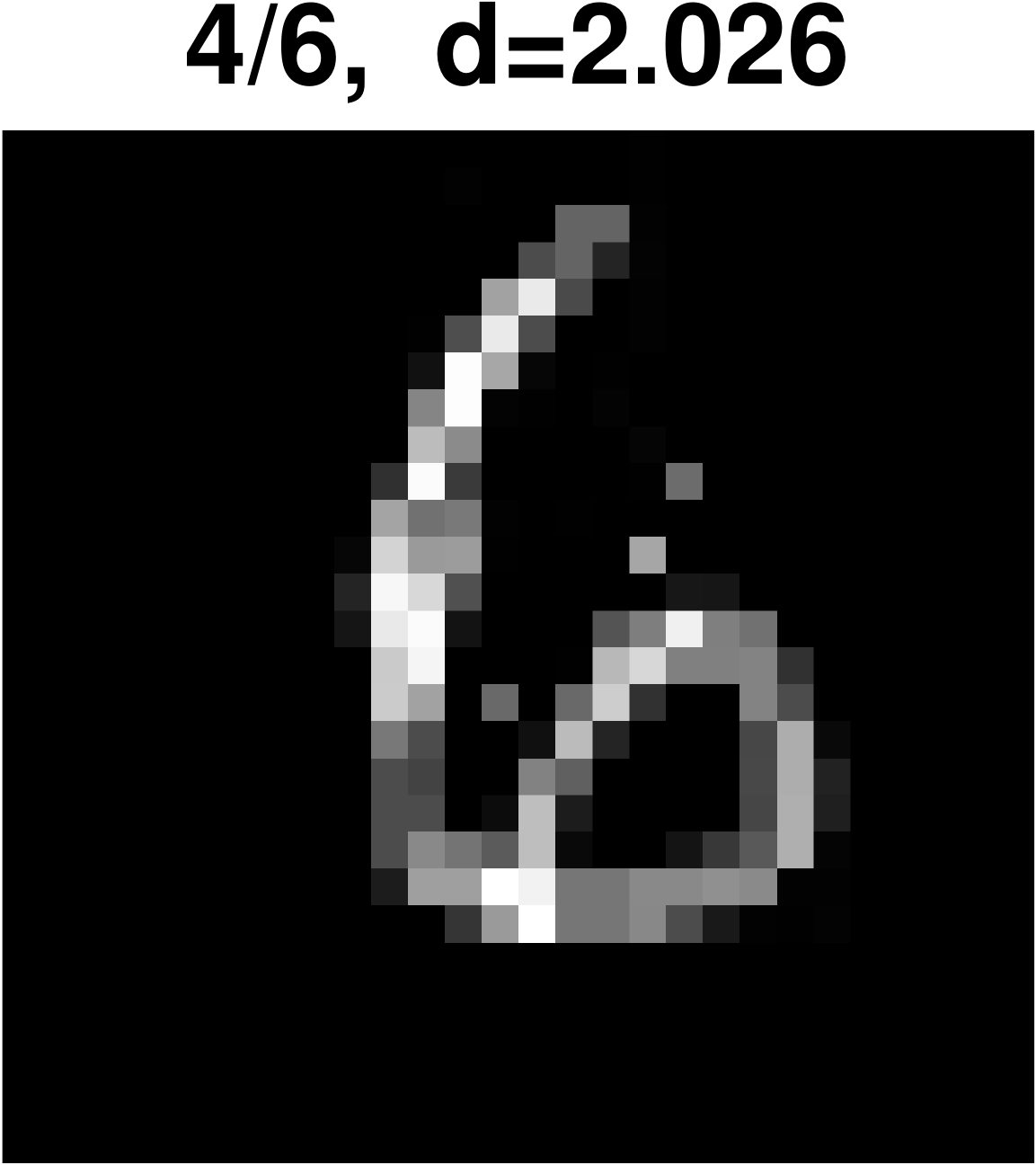}
	\includegraphics[width=0.25\columnwidth]{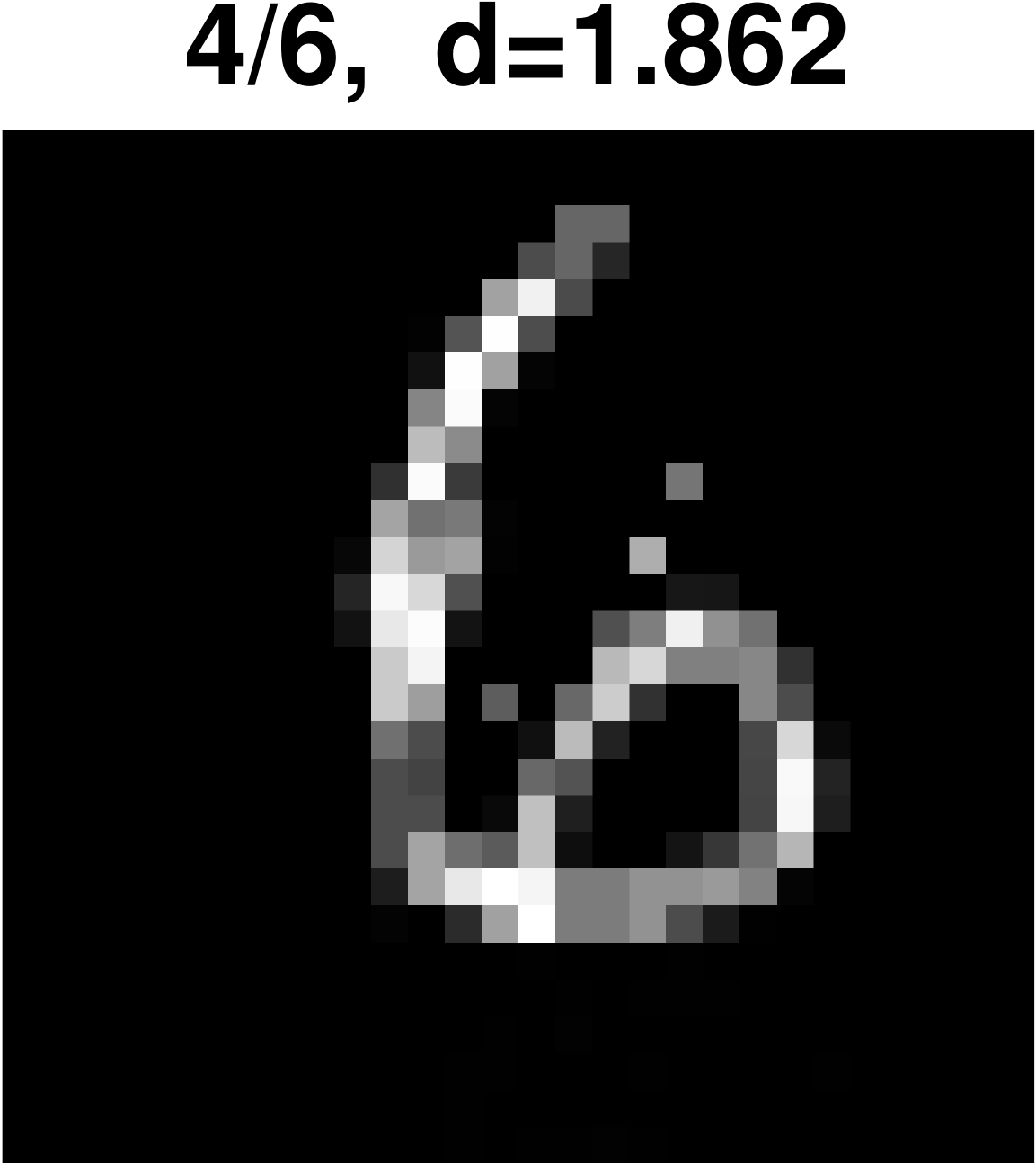}
	\includegraphics[width=0.25\columnwidth]{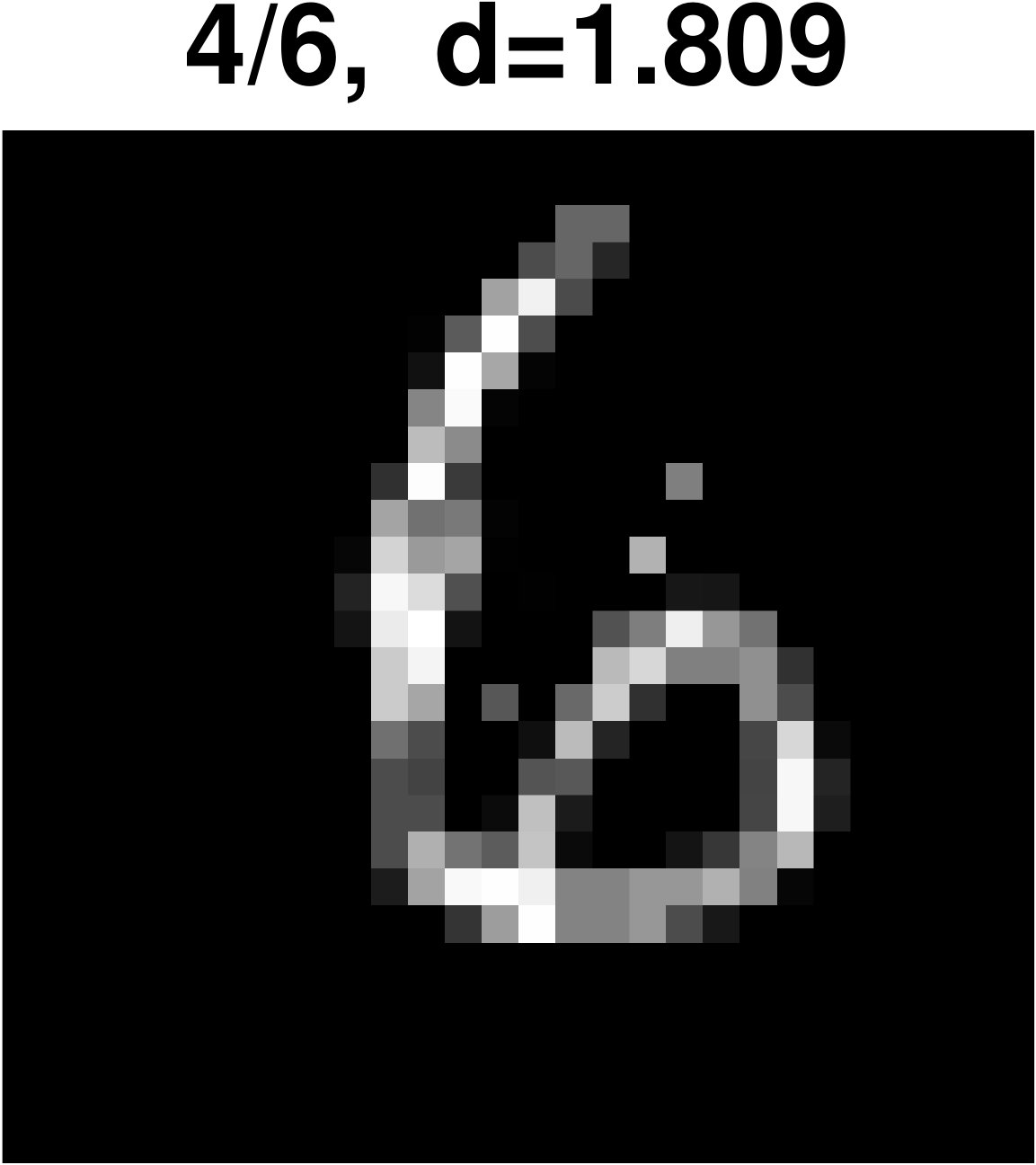}
	\includegraphics[width=0.25\columnwidth]{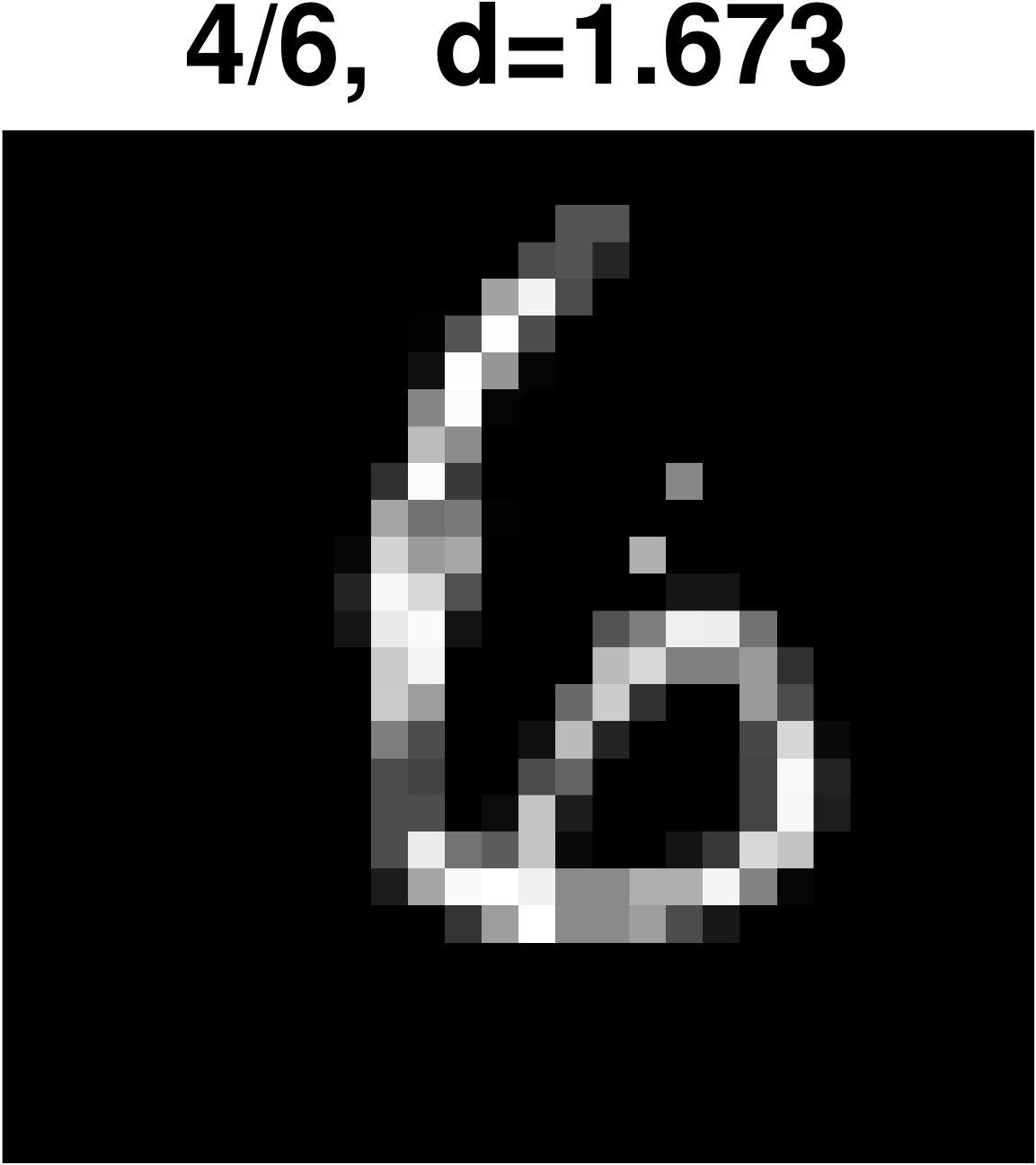}
	\includegraphics[width=0.25\columnwidth]{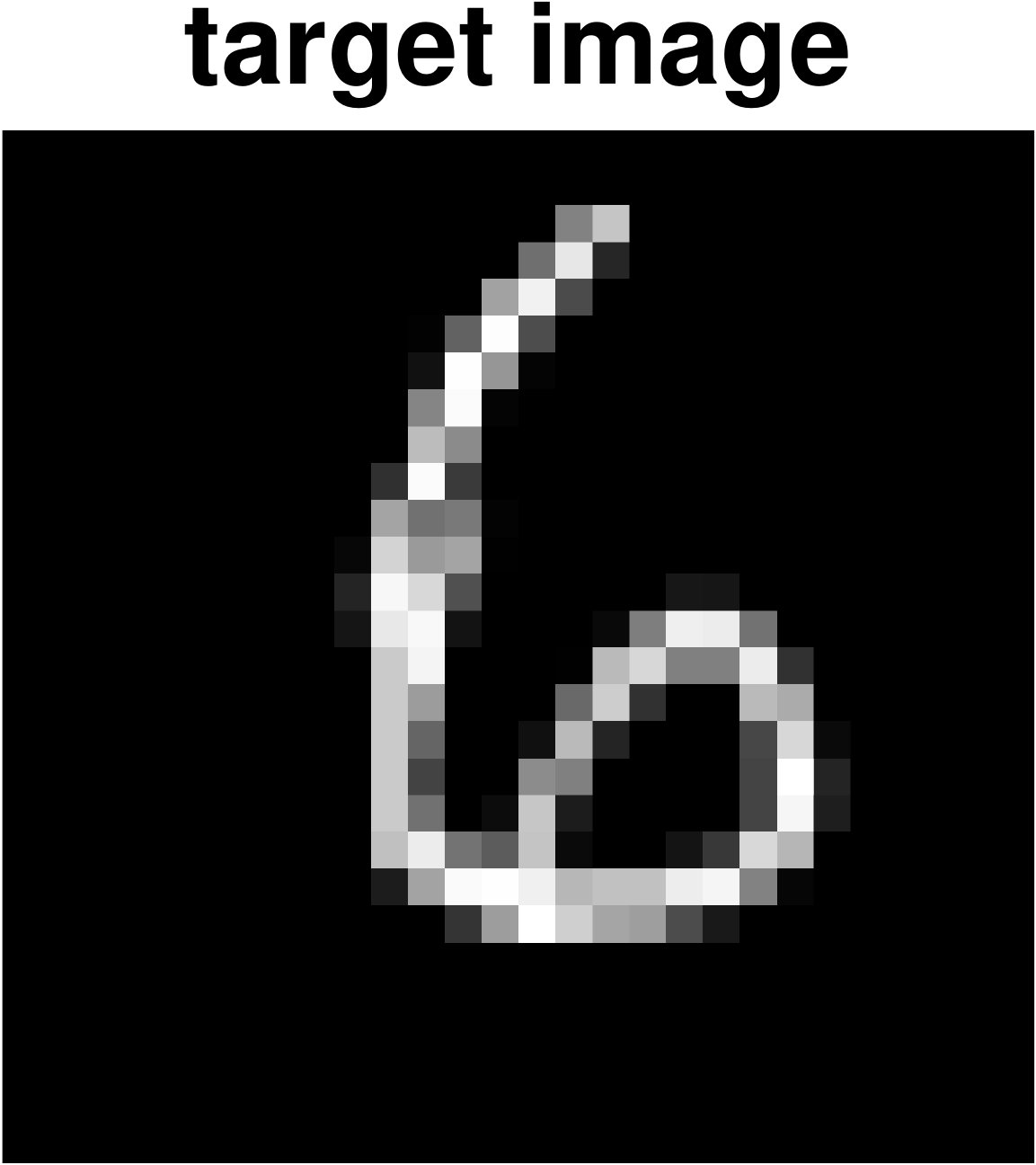}
	\\
	\vspace{4mm}
	\includegraphics[width=0.25\columnwidth]{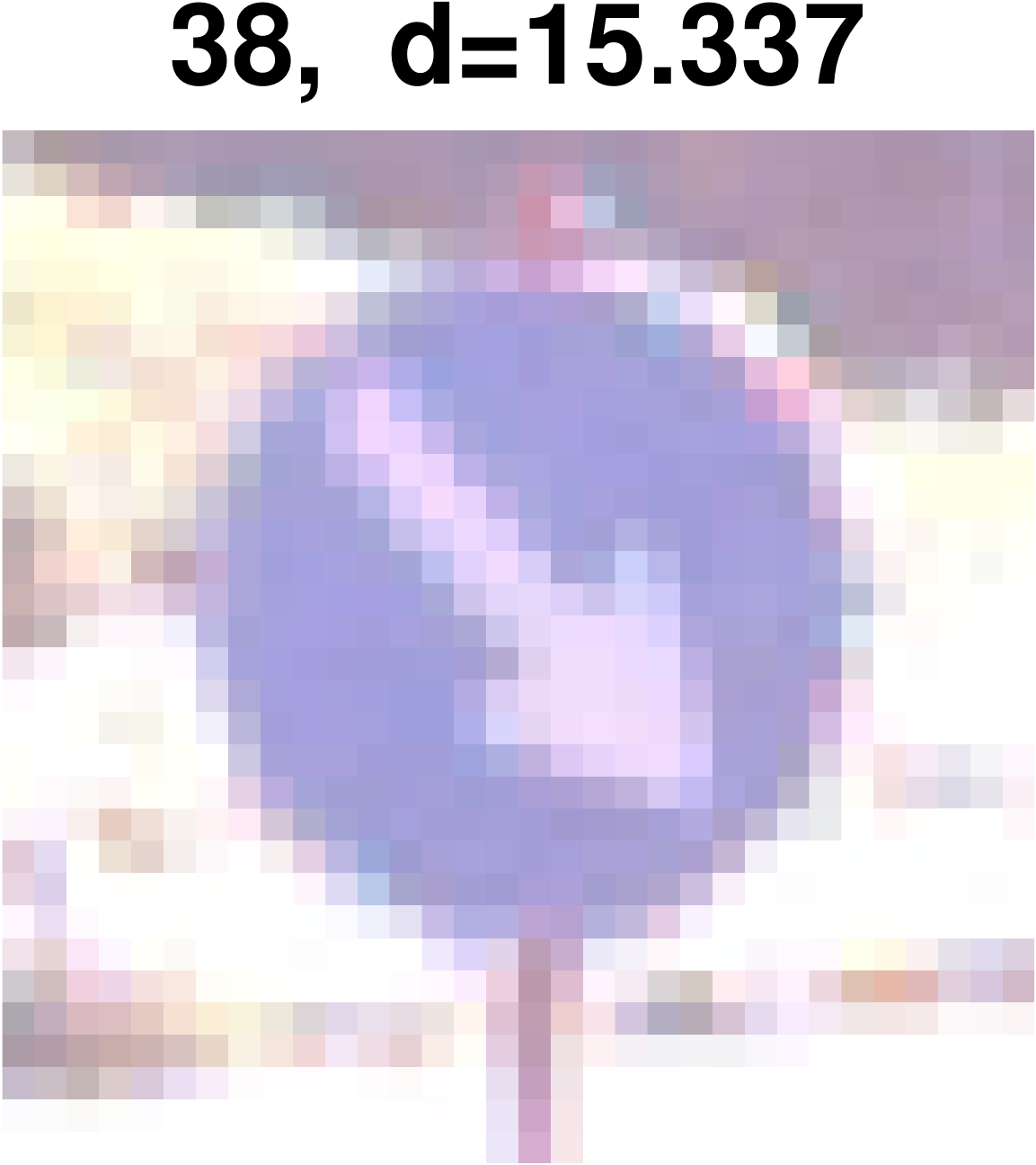}
	\includegraphics[width=0.25\columnwidth]{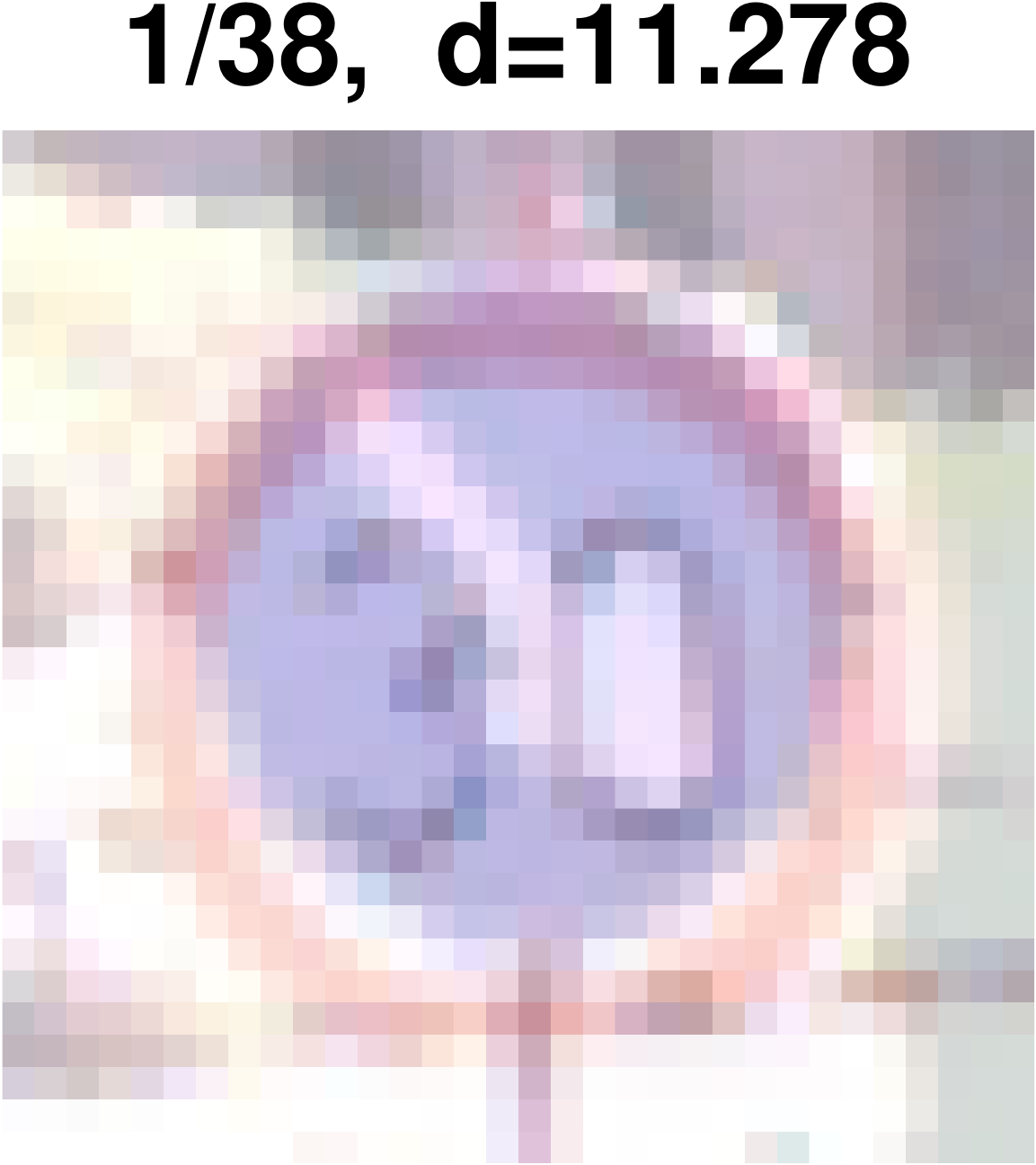}
	\includegraphics[width=0.25\columnwidth]{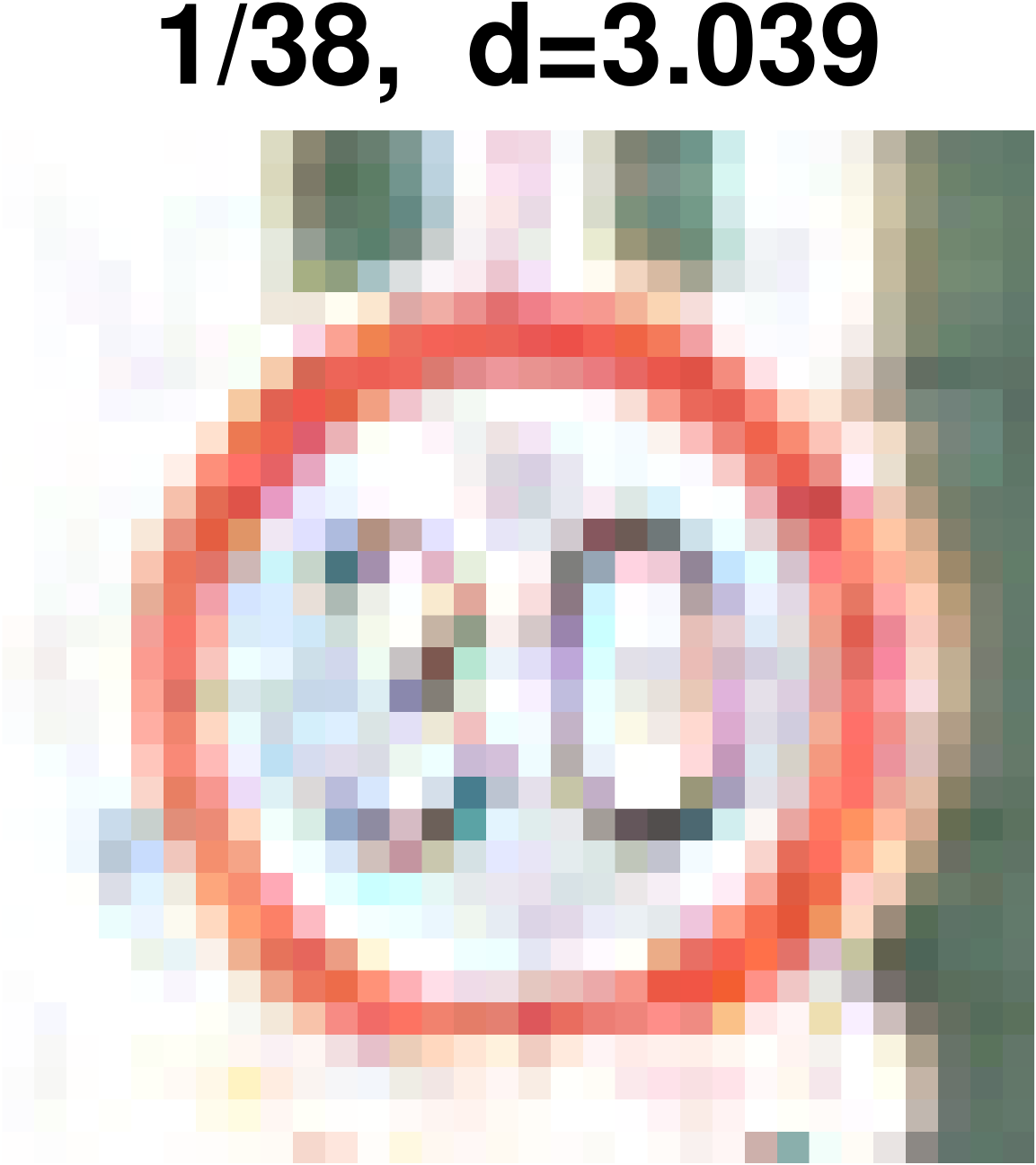}
	\includegraphics[width=0.25\columnwidth]{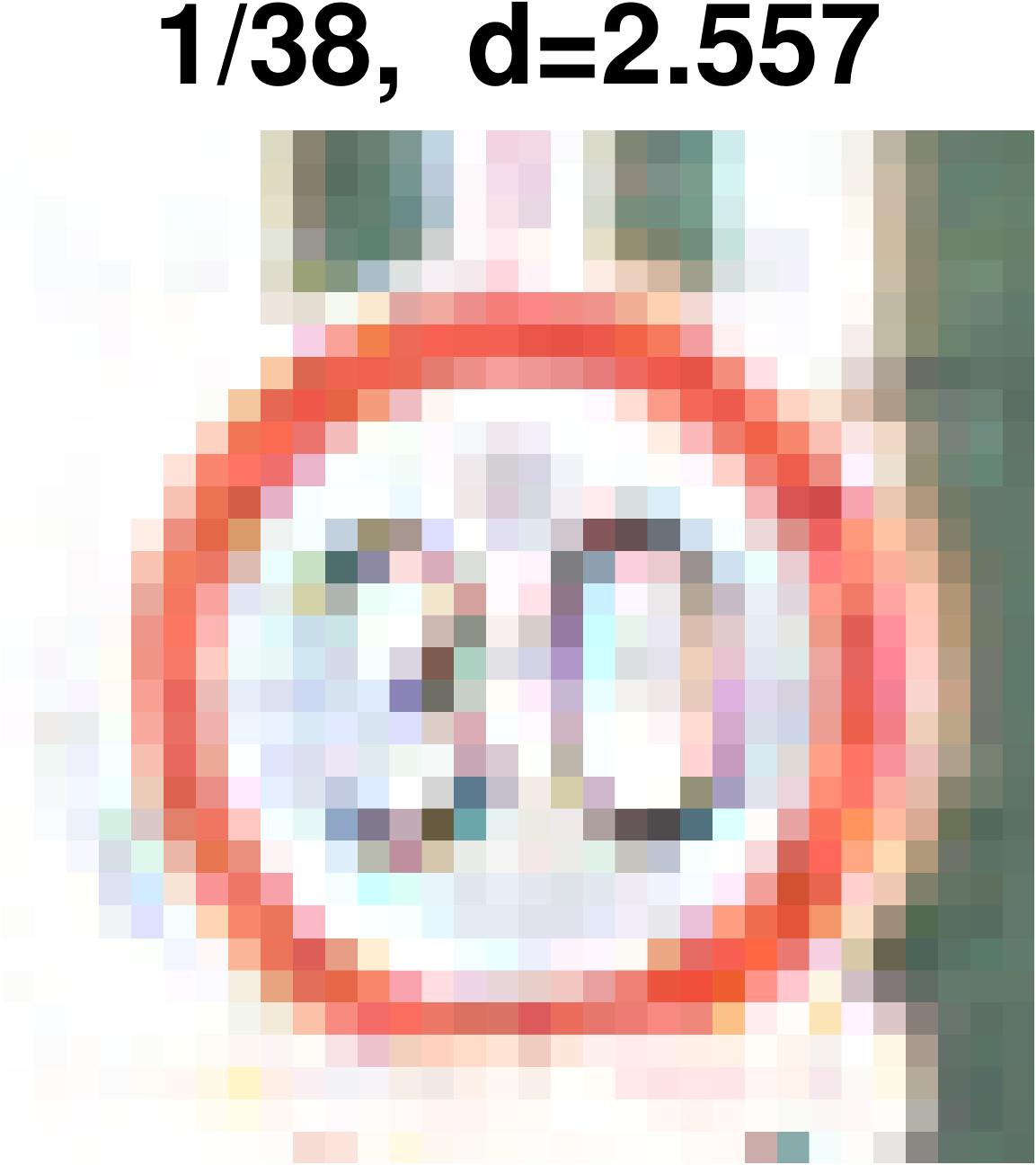}
	\includegraphics[width=0.25\columnwidth]{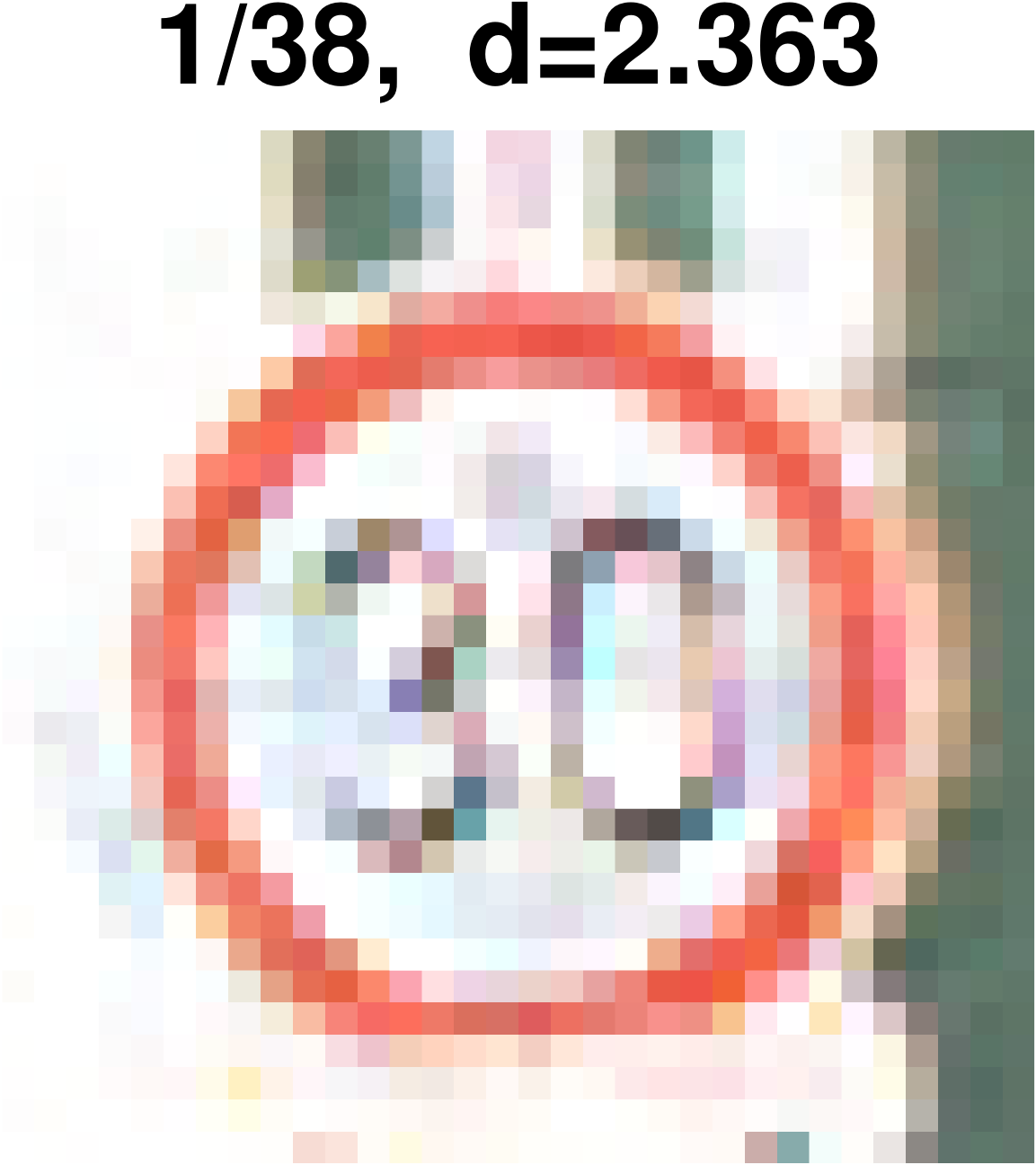}
	\includegraphics[width=0.25\columnwidth]{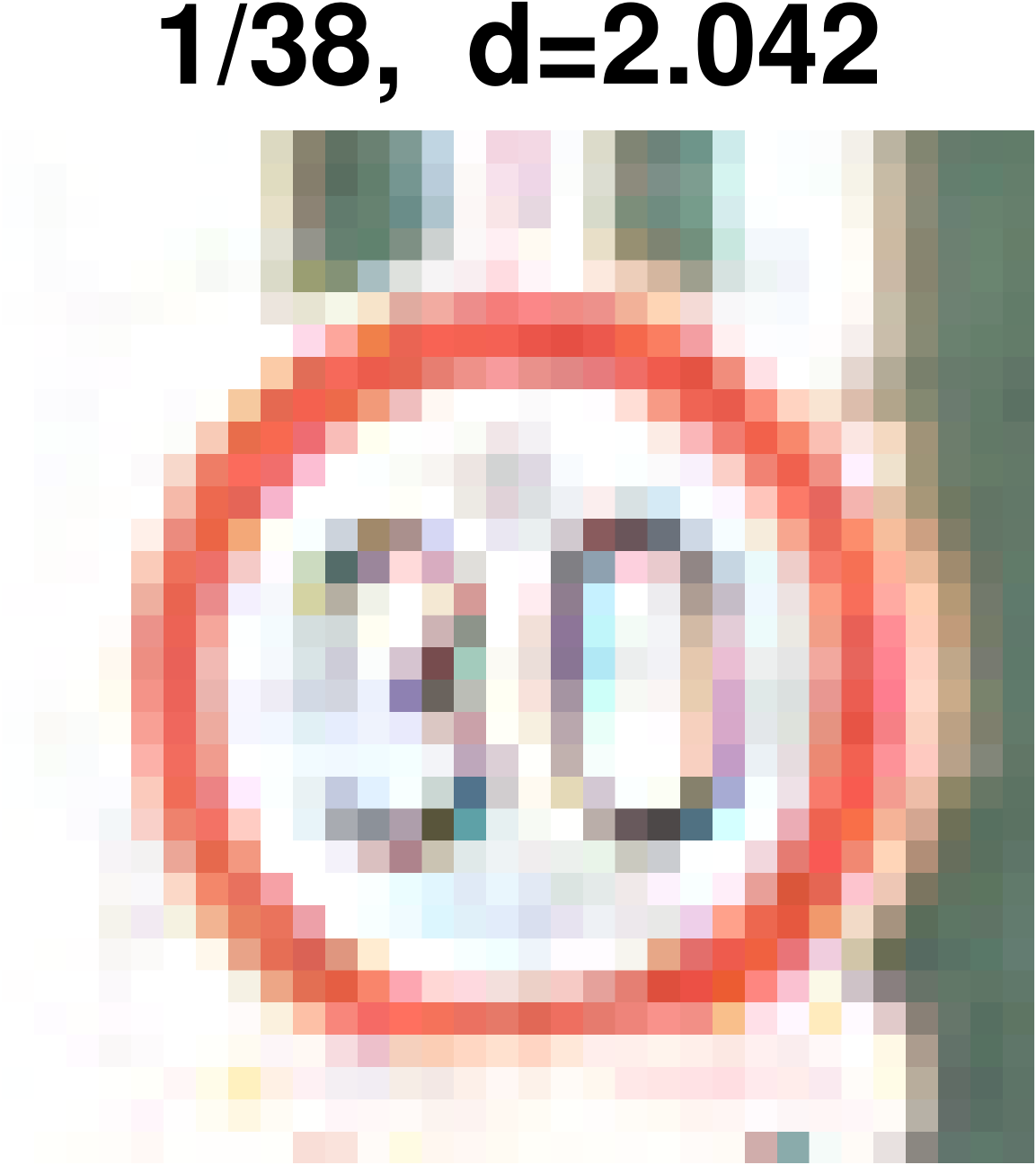}
	\includegraphics[width=0.25\columnwidth]{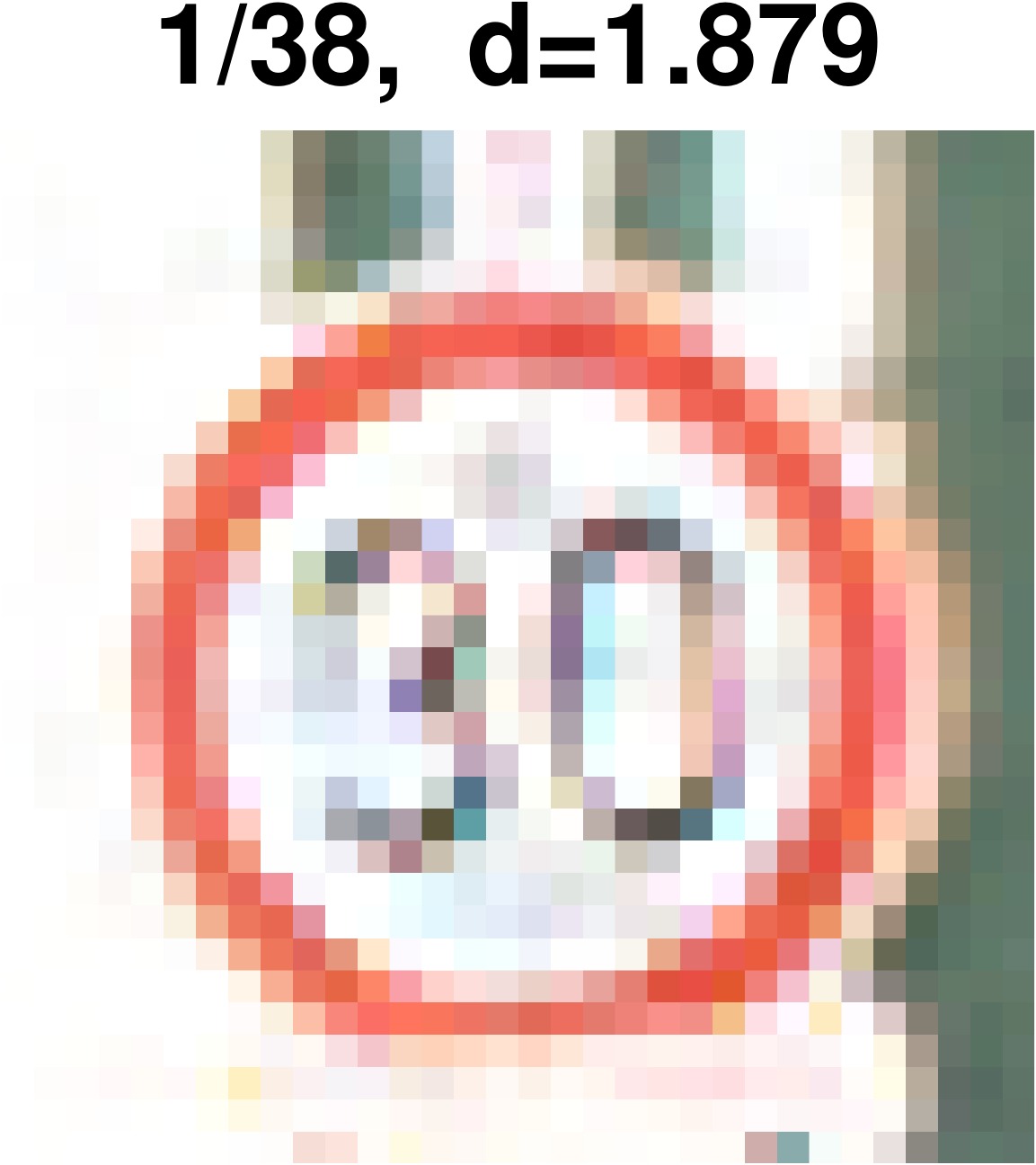}
	\includegraphics[width=0.25\columnwidth]{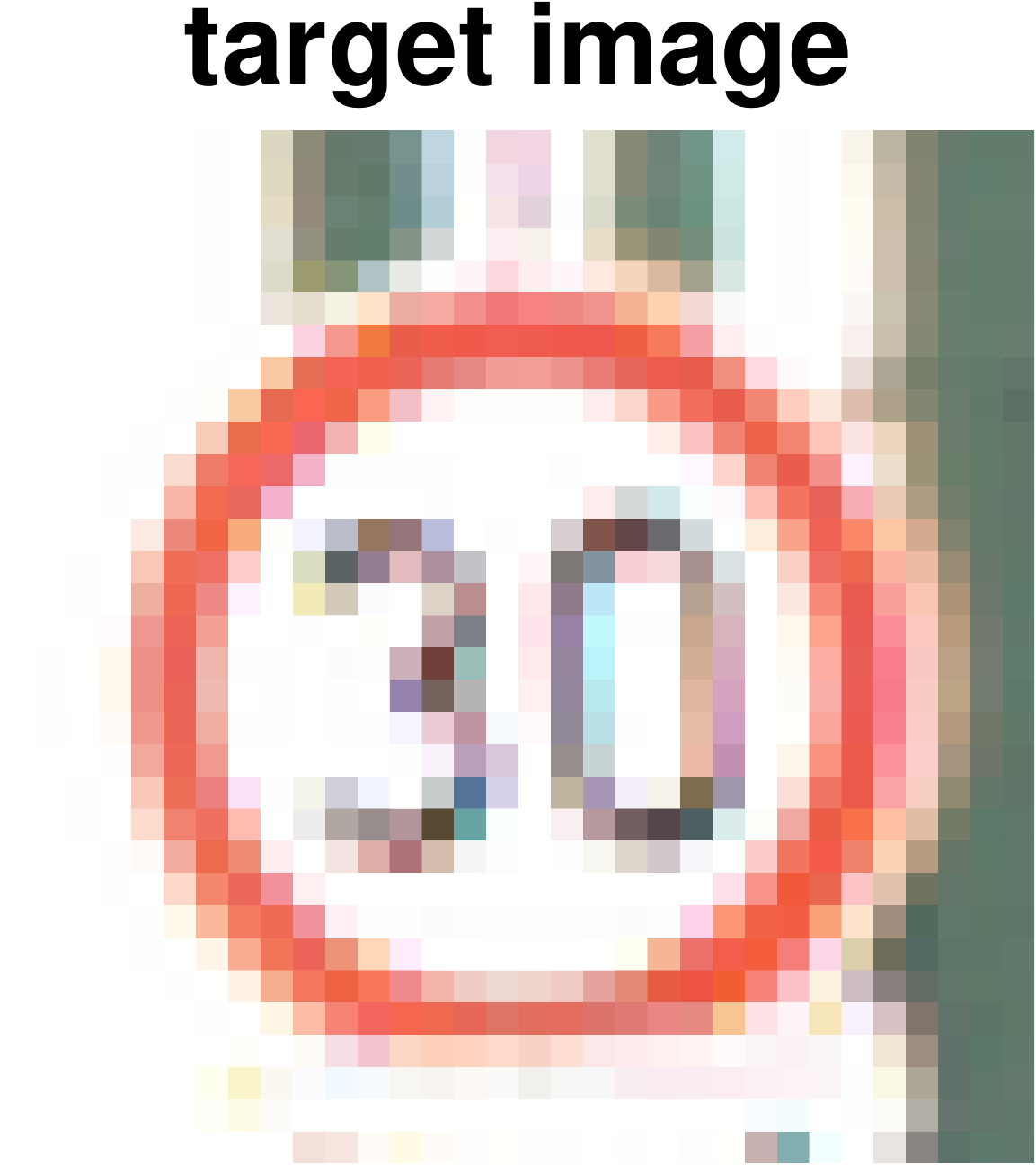}
	\\
	\vspace{4mm}
	\includegraphics[width=0.25\columnwidth]{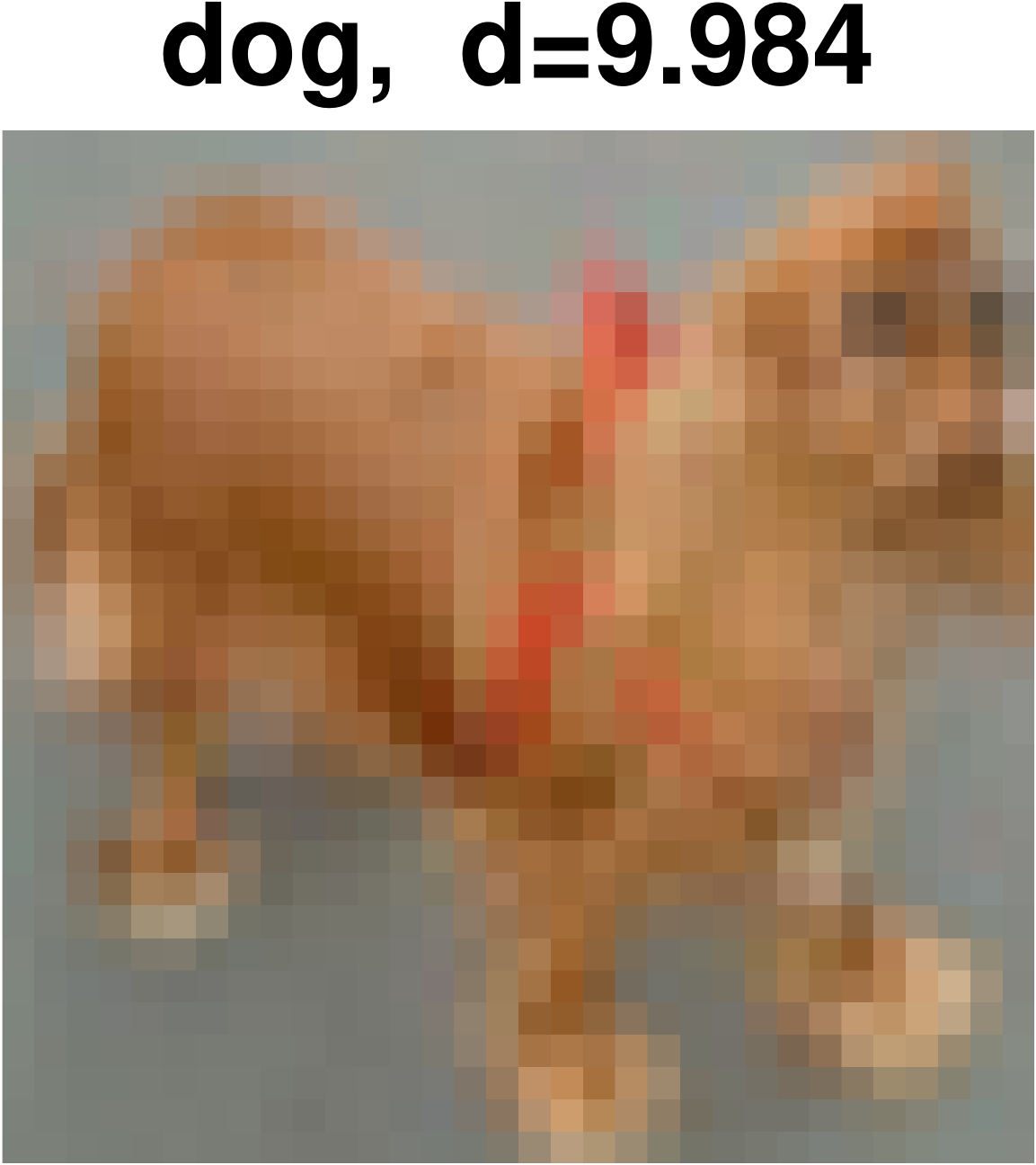}
	\includegraphics[width=0.25\columnwidth]{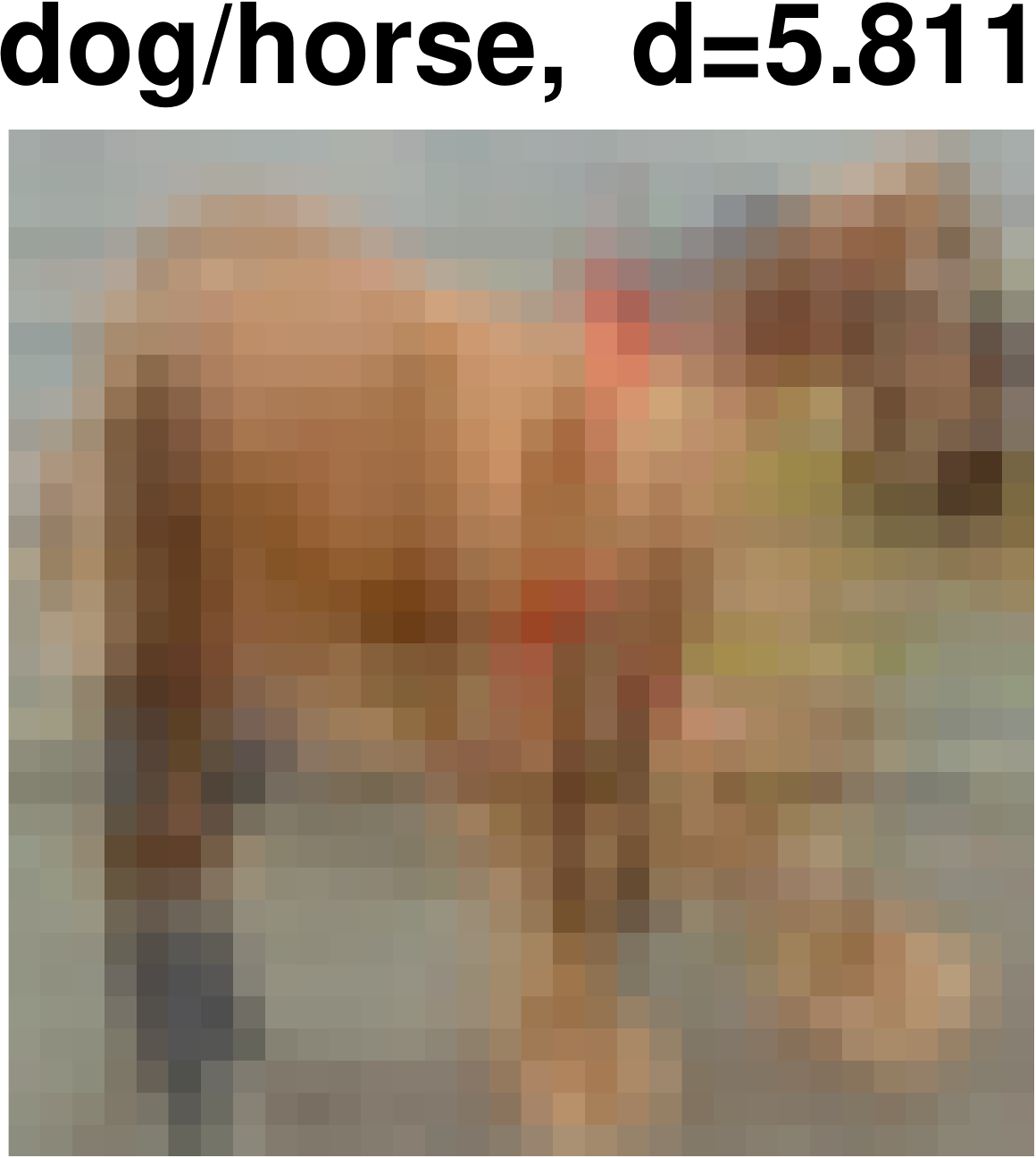}
	\includegraphics[width=0.25\columnwidth]{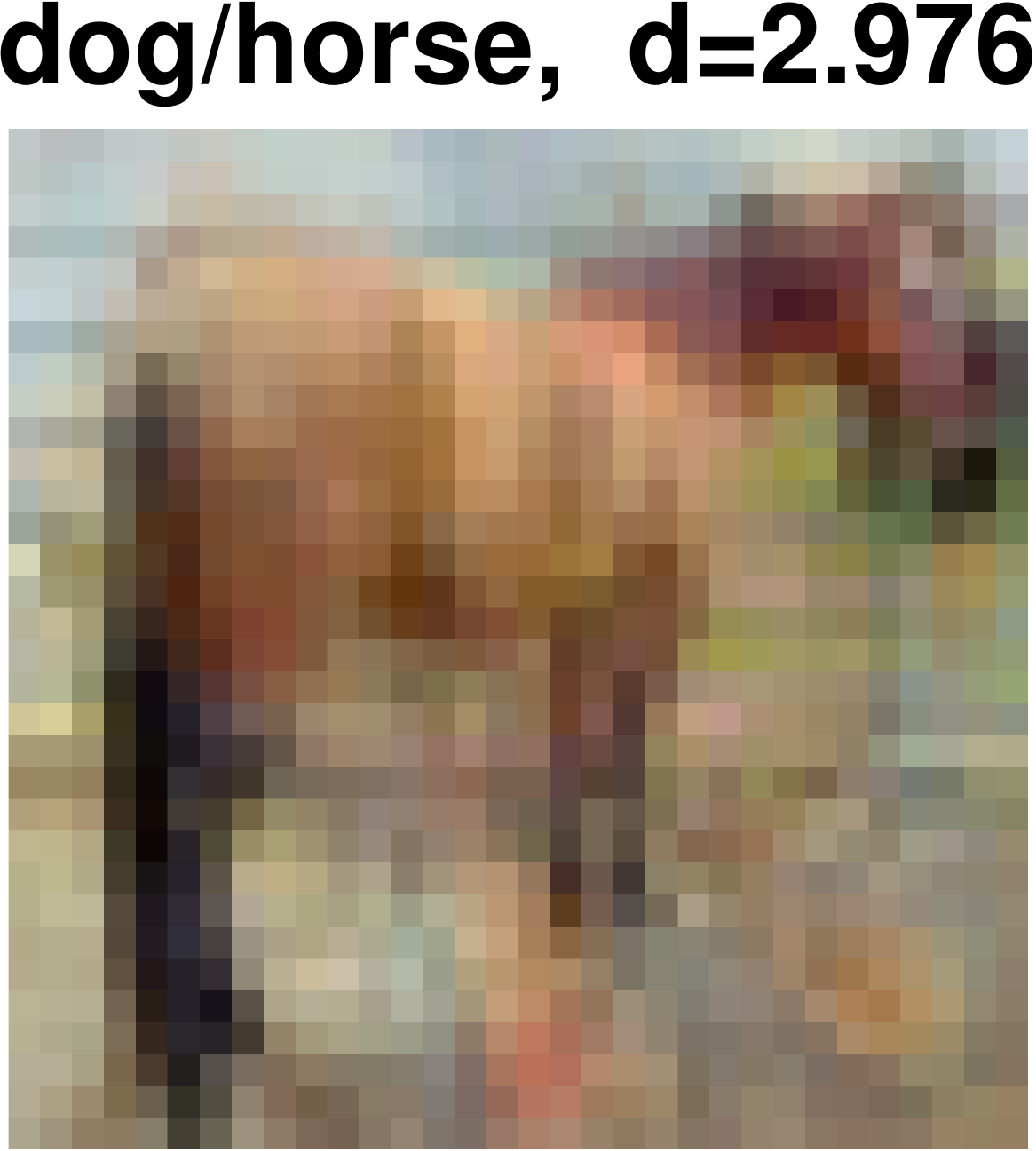}
	\includegraphics[width=0.25\columnwidth]{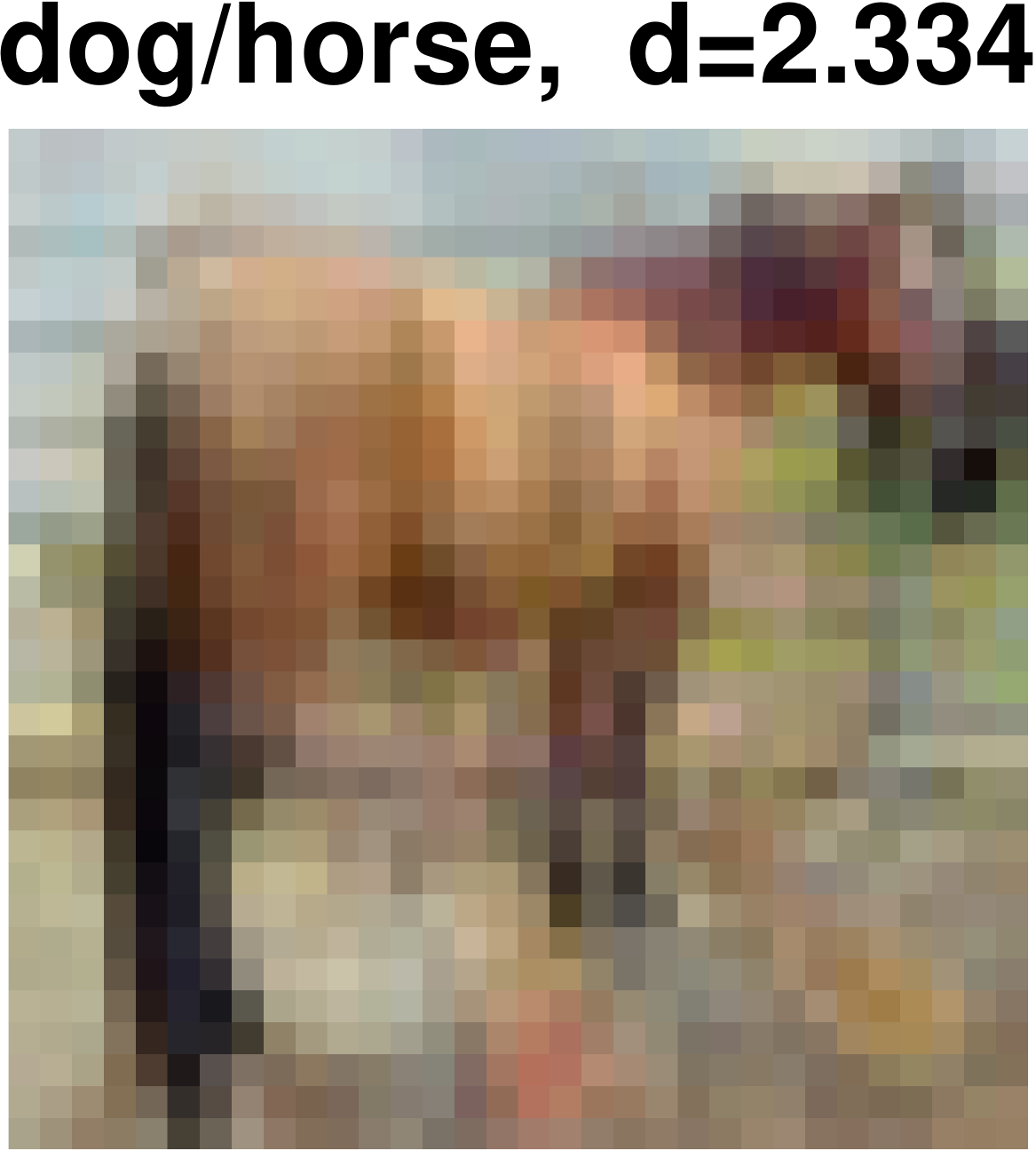}
	\includegraphics[width=0.25\columnwidth]{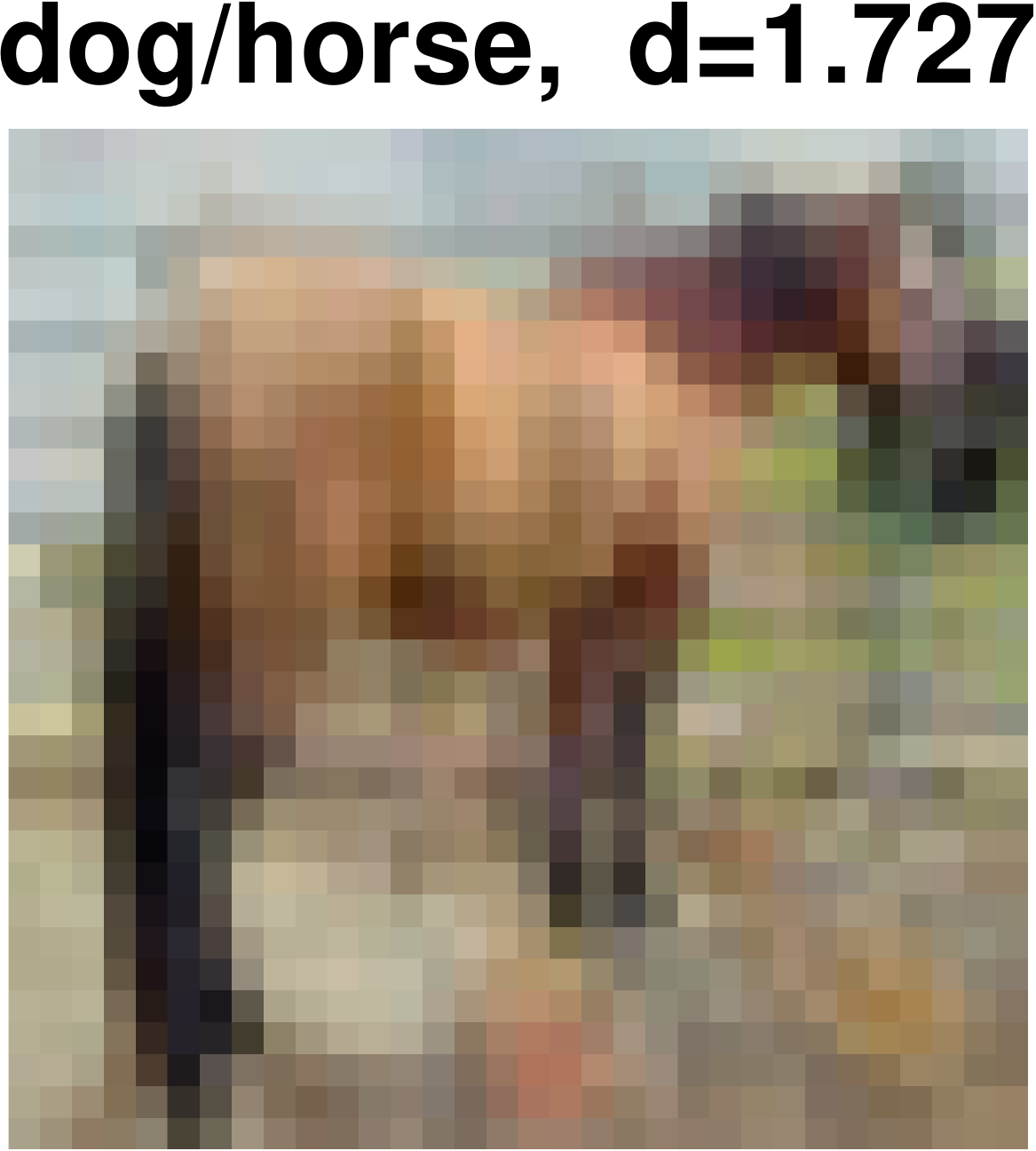}
	\includegraphics[width=0.25\columnwidth]{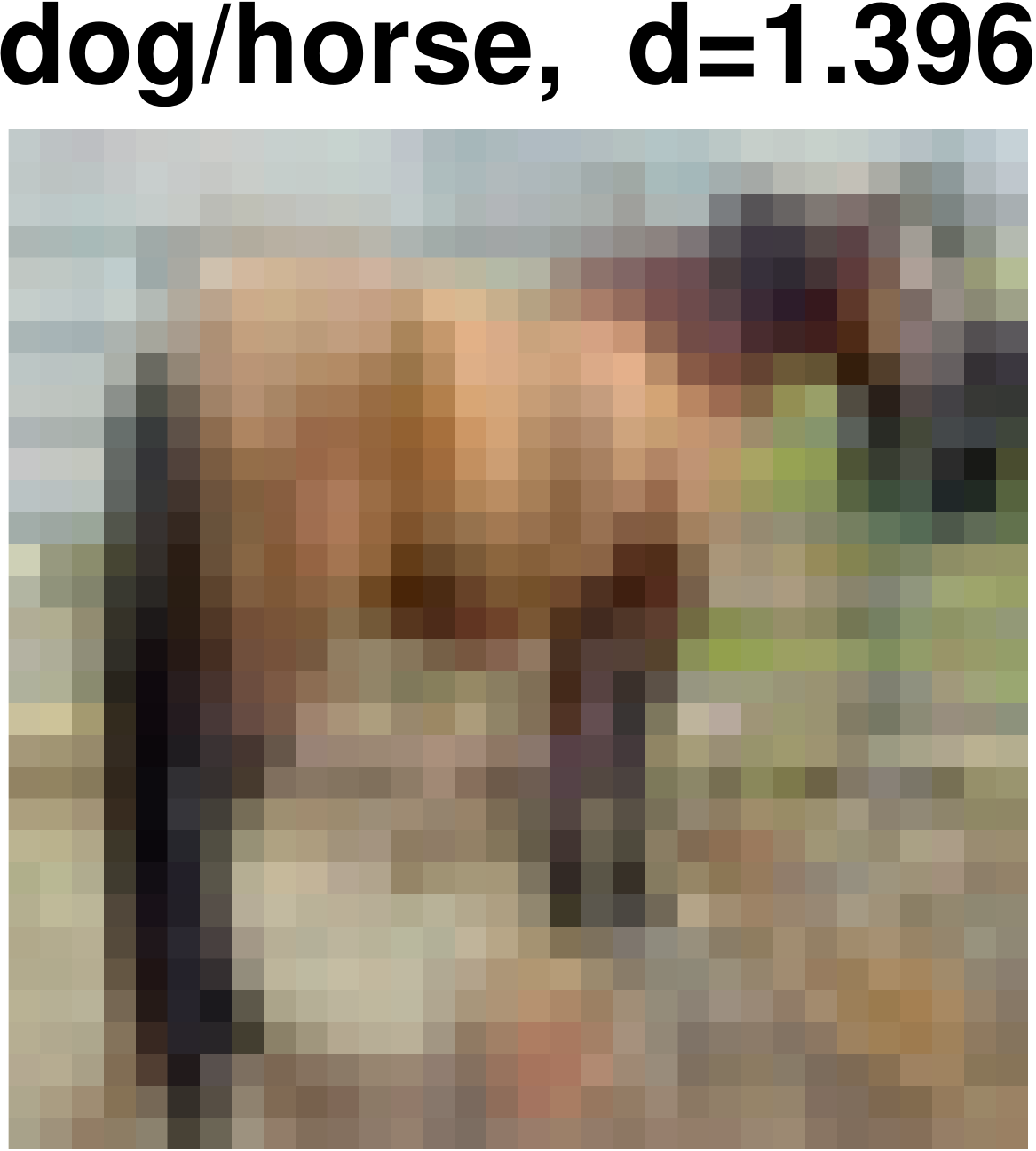}
	\includegraphics[width=0.25\columnwidth]{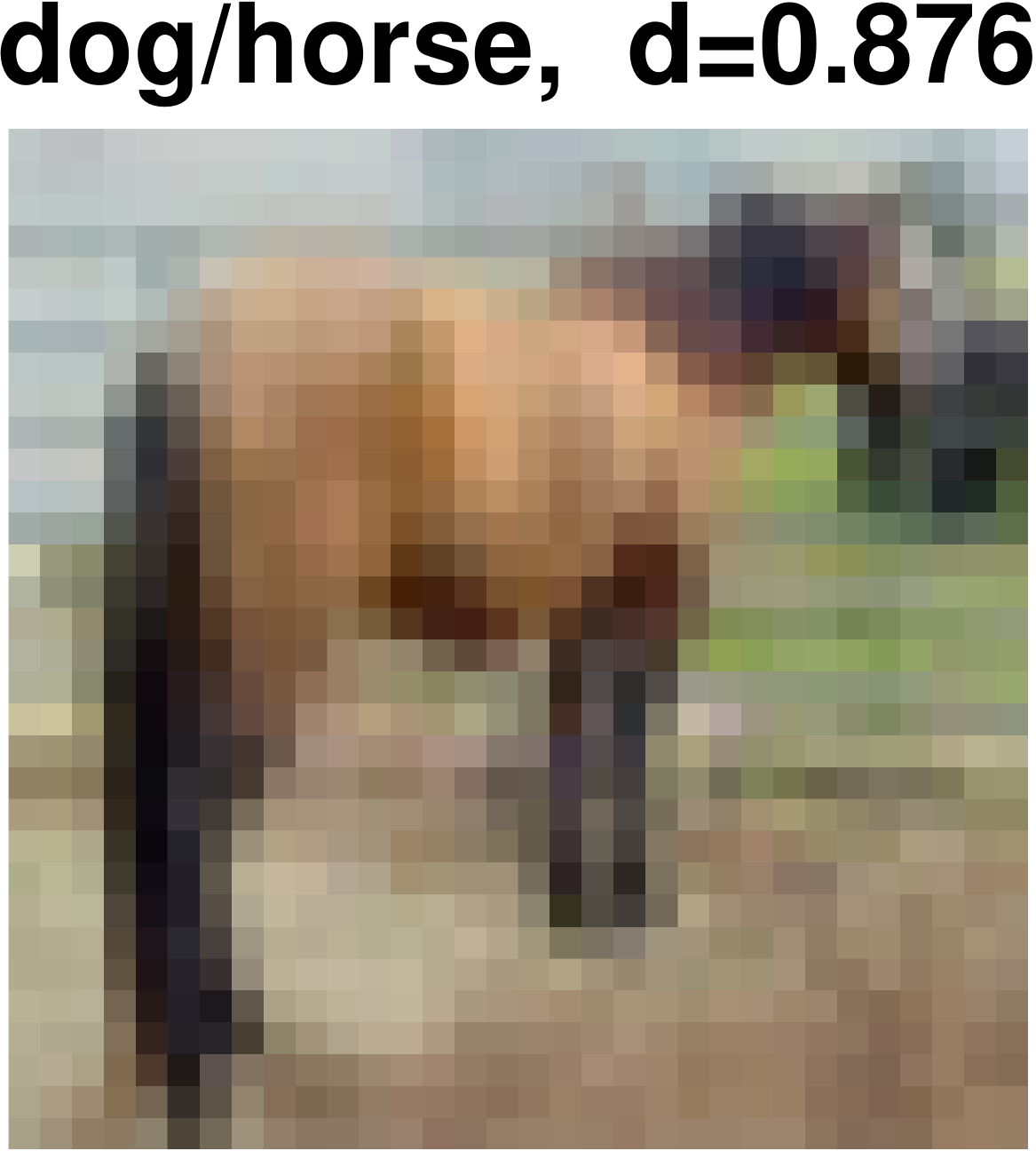}
	\includegraphics[width=0.25\columnwidth]{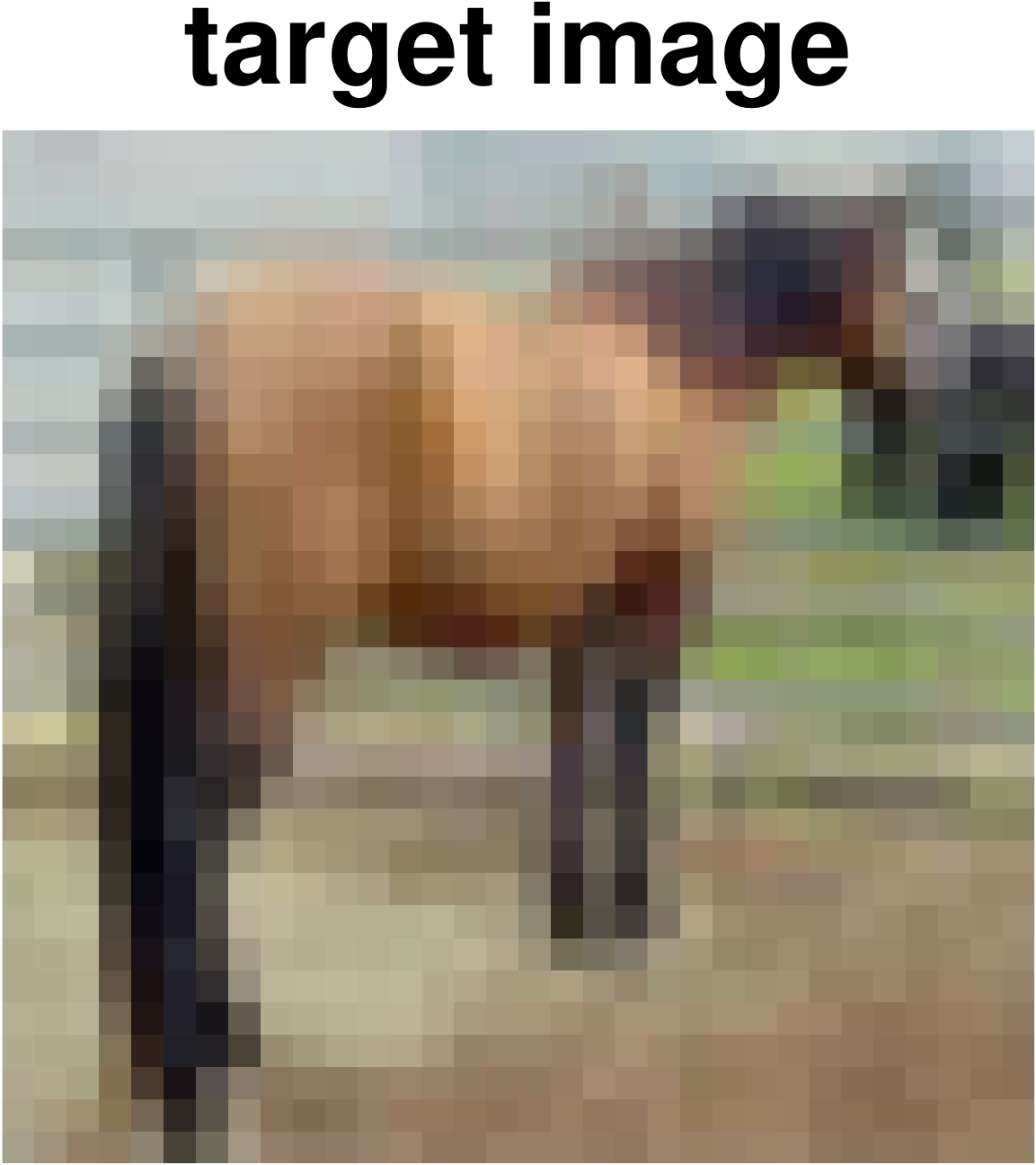}
	\caption{Progression of our attack. In each row, the first image is from the training set, the second is obtained with the linear search towards the target image (last image) for which we create an adversarial example. The other images are intermediate adversarial images found by our attack (the seventh is the final output). Apart from the starting image and the target image, all are on the decision boundary, that is between the classes indicated on top of each picture ($0/8$ means it is on the decision boundary between class $0$ and $8$). We also report the $l_2$-distance between each image and the target image (d). First three rows: non-robust plain model, last three rows: $l_\infty$ (first) and $l_2$ (second, third) adversarially trained models on MNIST, GTS and CIFAR-10.}
	\label{fig:vis}
\end{figure*}

\section{Visualizing the decision boundary}\label{sec:vis}
While our attack runs, almost at each iteration an image lying on the decision boundary, that is the classifier outputs assigns the same (up to a tolerance) probability for the input to belong to different classes, is available. In fact, unless the linear region to which the current solution belongs does not intersect the decision boundary, %(where the original and the target class are the same probability),
the solution of problem \eqref{eq:advopt_lin} is attained when the first constraint holds as an equality.\\
In Figure \ref{fig:vis} we show some of these intermediate solutions found while crafting an adversarial example. The first three rows are obtained attacking the plain models reported in the Section \ref{sec:exp}, while for the fourth to sixth row we used respectively the $l_\infty$-\textit{at} network on MNIST and the $l_2$-\textit{at} classifiers on GTS and CIFAR-10. For every row, the first image is the starting point of our method and belongs to the training set of the respective dataset, while the second image is the point we get through the initial binary search on the segment joining the starting point and the target image for which we want to provide an adversarial perturbation (represented in the last image of each row). We also report the $l_2$-distance between each image and the target image, which is equivalent to the $l_2$-norm of the adversarial manipulation found at that iteration of the algorithm.\\
We can see that, apart from the starting image and the target image, all the images lie on the decision boundary. Furthermore, in many cases, although the distance from the target image is notable, they are clearly assignable to a specific class, meaning that the decision boundary is still wrong showing that there is still quite some way to go
if we want to achieve robustness with respect to human perception of these images.\\

\begin{figure}[h]
	\centering
	\includegraphics[width=0.4\columnwidth]{gts_lenet_small_plaind_11566_sp_16141_ls_2_1_2.pdf}
	\includegraphics[width=0.4\columnwidth]{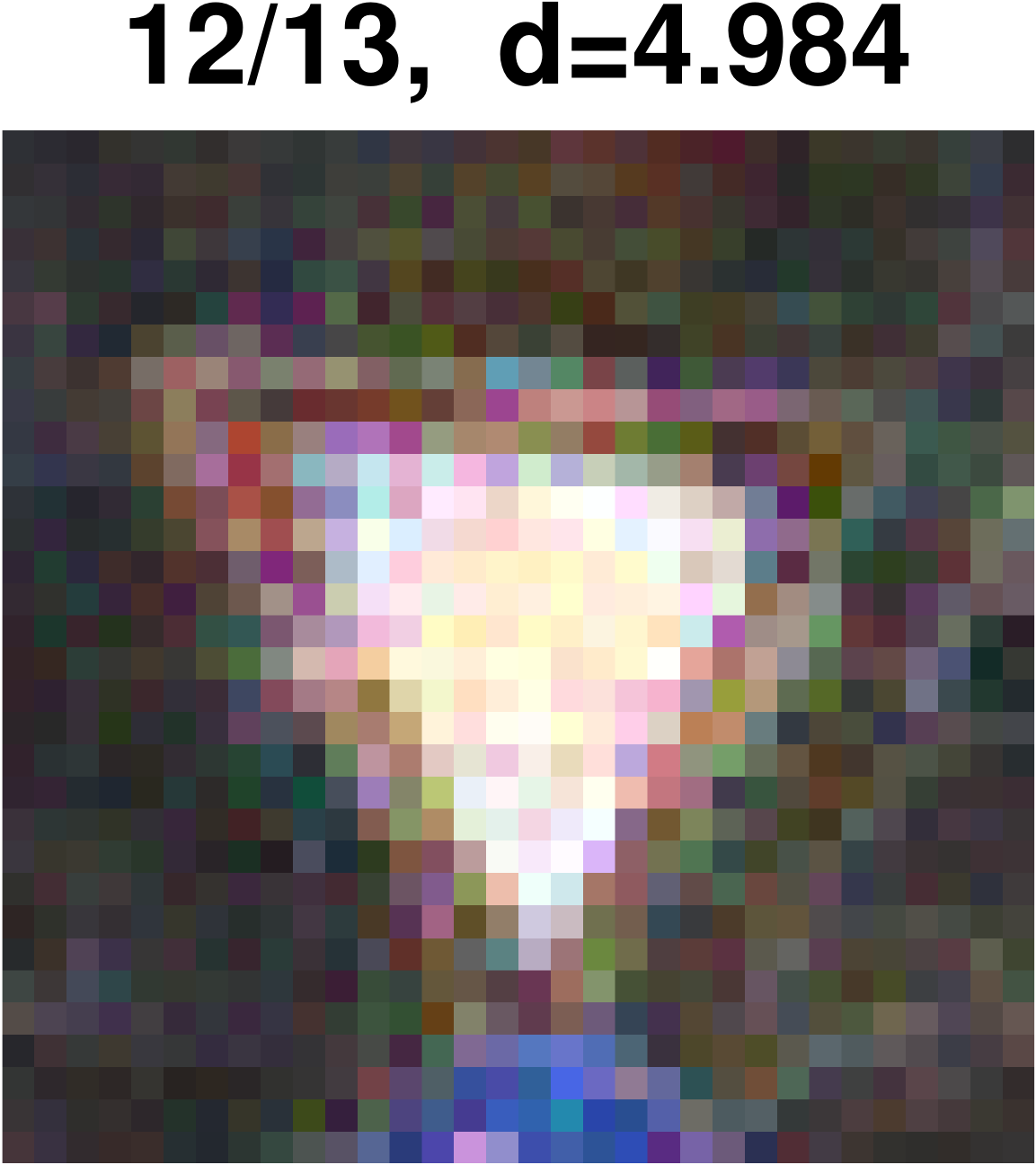}
	\\
	\vspace{4mm}
	\includegraphics[width=0.4\columnwidth]{gts_lenet_small_at_5966_sp_32682_ls_1_2.pdf}
	\includegraphics[width=0.4\columnwidth]{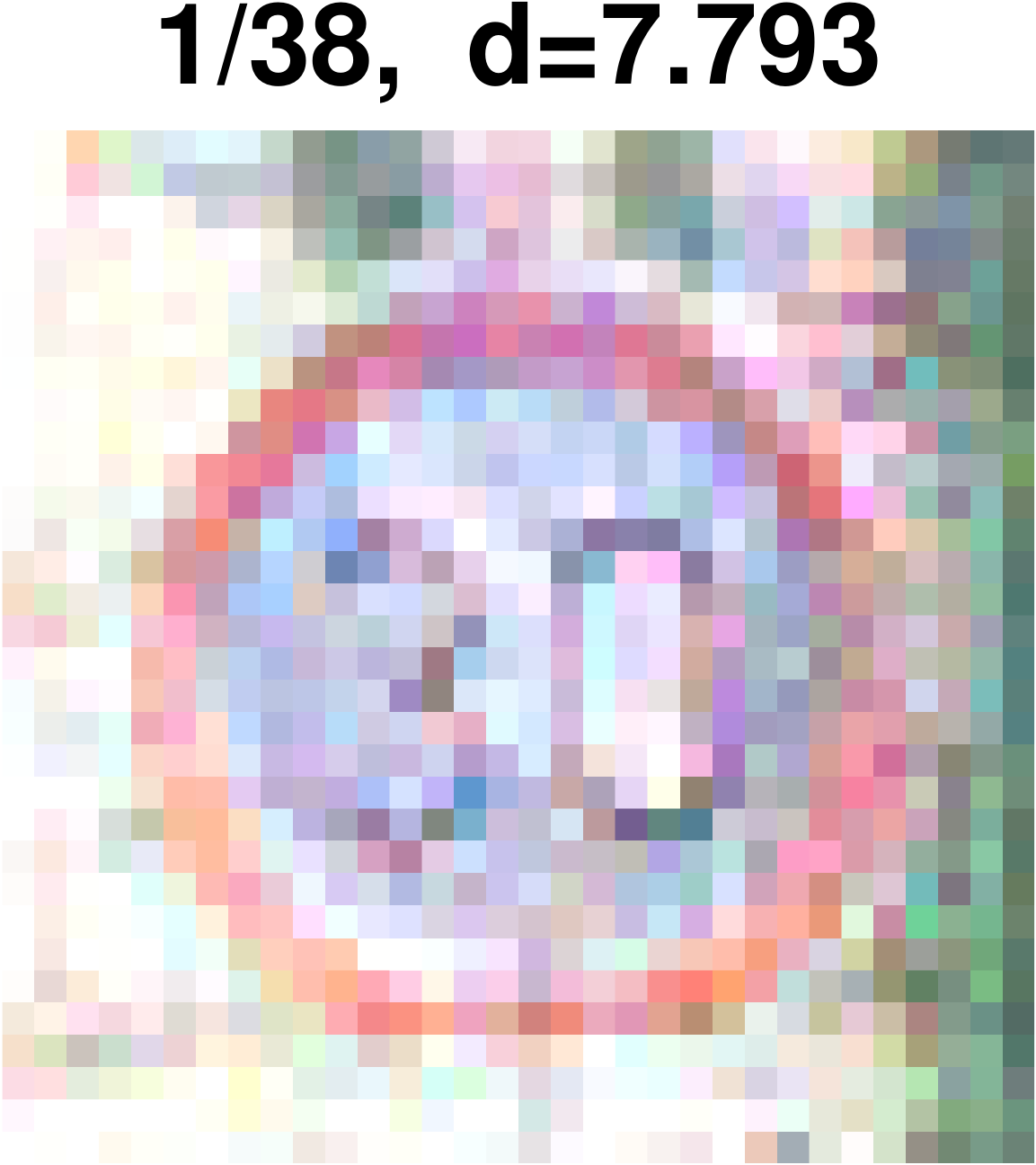}
	\caption{The linear regions can be large. For the same cases reported in Figure \ref{fig:vis} for GTS we show here the image got by the initial linear search, say $y$, and that obtained by solving \eqref{eq:advopt_lin} on the first region $Q(y)$. This means that the two images of each row belong to the same linear region even though their appearance is
	quite different. This shows that some of the linear regions cover quite large parts of the input space.}
	\label{fig:starting_points}
\end{figure}

We can also check how large the linear regions are. The first polytope $Q(y)$ our attack checks is the one containing the point $y$ of the linear search performed as initial step of the attack between the image from the training set and the target image. We show the image $y$ and the solution of \eqref{eq:advopt_lin} on $Q(y)$. Both images are contained in $Q(y)$ and both lie on the decision boundary. In Figure \ref{fig:starting_points} we show these two images for some cases for the GTS models. It is interesting that, although the number of polytopes is extremely large, they are still wide enough to contain images of such different appearance and with significant $l_2$-distance. 
%In particular, already with a single step is sometimes possible to obtain what one can consider an adversarial example, without even leaving the first linear region selected.

\section{Conclusion}
{%\color{blue}
We extended the white-box gradient-free adversarial attack of \cite{CroHei18} by i) deriving a new, scalable QP solver, ii) solving the QP problem efficiently on GPU without computing the constraint matrix explicitly, iii) adding support for more layer types, and iv) introducing a new attack scheme to select regions. Taken together, these improvements allowed us to attack larger and more complex neural networks in less time and finding better adversarial examples. We demonstrated the importance of evaluating robustness with our attack by showing that all the established methods for producing adversarial examples have at least one case where they estimate a robust accuracy at least 50\% higher (in absolute value) than that given by the best attack, while our attack is never farther than 5\%. This means that, while most of the attacks perform well on average, for all of them except ours there exist situations where they heavily overestimate the adversarial robustness.}

\begin{acknowledgements}
F.C. and M.H. acknowledge support from the BMBF through the T\"ubingen AI Center (FKZ: 01IS18039A) and by the DFG via grant 389792660 as part of TRR 248
and the Excellence Cluster ``Machine Learning - New Perspectives for Science''. J.R. acknowledges support from the Bosch Research Foundation (Stifterverband, T113/30057/17) and the International Max Planck Research School for Intelligent Systems (IMPRS-IS).
\end{acknowledgements}

\bibliographystyle{spbasic}      % basic style, author-year citations
%\bibliographystyle{plain}
%\bibliography{Literatur}   % name your BibTeX data base

\end{document}